%% file: main.tex
\documentclass[10pt,twocolumn,letterpaper]{article}

\usepackage[pagenumbers]{cvpr}

\usepackage[utf8]{inputenc} 
\usepackage[T1]{fontenc}    
\usepackage{url}            
\usepackage{booktabs}       
\usepackage{amsfonts}       
\usepackage{nicefrac}       
\usepackage{microtype}      
\usepackage{xcolor}         

\usepackage{graphicx}
\usepackage{amsmath}
\usepackage{amssymb}
\usepackage{algorithm}
\usepackage{algorithmic}
\usepackage{color, colortbl}
\usepackage{dsfont}
\usepackage{caption}
\usepackage{enumitem}
\usepackage{multirow}  
\usepackage[accsupp]{axessibility} 

\definecolor{Gray}{gray}{0.9}
\input{macros}
\newcommand{\z}{\ensuremath{\mathbf{z}}}
\newcommand{\x}{\ensuremath{\mathbf{x}}}
\newcommand{\Z}{\ensuremath{\mathbf{Z}}}
\newcommand{\X}{\ensuremath{\mathbf{X}}}

\newcommand{\Programs}{\ensuremath{P}}
\newcommand{\RealShapes}{\ensuremath{S^*}}

\usepackage[pagebackref,breaklinks,colorlinks]{hyperref}

\usepackage[capitalize]{cleveref}
\crefname{section}{Sec.}{Secs.}
\Crefname{section}{Section}{Sections}
\Crefname{table}{Table}{Tables}
\crefname{table}{Tab.}{Tabs.}

\def\cvprPaperID{7785}
\def\confName{CVPR}
\def\confYear{2022}

\begin{document}

\title{PLAD: Learning to Infer Shape Programs with\\Pseudo-Labels and Approximate Distributions}

\author{R. Kenny Jones\\
Brown University\\
\and
Homer Walke \\
UC Berkeley\\
\and
Daniel Ritchie \\
Brown University\\
}
\maketitle

\begin{abstract}

\input{00-abstract.tex}

\end{abstract}

\input{01-intro.tex}

\input{02-related.tex}

\input{03-method.tex}
\input{04-results.tex}
\input{05-conclusion.tex}
\input{06-acks}

{\small
\bibliographystyle{ieee_fullname}
\bibliography{main}
}

\appendix
\input{supplemental.tex}

\end{document}

%% file: macros.tex
\newcommand{\Reals}{\ensuremath{\mathds{R}}}

\newenvironment{packed_itemize}
{\begin{itemize}
    \vspace{-\topsep}
    \setlength{\itemsep}{1pt}
    \setlength{\parskip}{0pt}
    \setlength{\parsep}{0pt}
}{\end{itemize}}

\newenvironment{packed_enumerate}
{\begin{enumerate}
    \vspace{-\topsep}
    \setlength{\itemsep}{1pt}
    \setlength{\parskip}{0pt}
    \setlength{\parsep}{0pt}
}{\end{enumerate}}

%% file: 00-abstract.tex
Inferring programs which generate 2D and 3D shapes is important for reverse engineering, editing, and more.
Training models to perform this task is complicated because paired (shape, program) data is not readily available for many domains, making exact supervised learning infeasible. 
However, it is possible to get paired data by compromising the accuracy of either the assigned program labels or the shape distribution.
Wake-sleep methods use samples from a generative model of shape programs to approximate the distribution of real shapes.
In self-training, shapes are passed through a recognition model, which predicts programs that are treated as ‘pseudo-labels’ for those shapes.
Related to these approaches, we introduce a novel self-training variant unique to program inference, where program pseudo-labels are paired with their executed output shapes, avoiding label mismatch at the cost of an approximate shape distribution.
We propose to group these regimes under a single conceptual framework, where training is performed with maximum likelihood updates sourced from either Pseudo-Labels or an Approximate Distribution (PLAD).
We evaluate these techniques on multiple 2D and 3D shape program inference domains.
Compared with policy gradient reinforcement learning, we show that PLAD techniques infer more accurate shape programs and converge significantly faster.
Finally, we propose to combine updates from different PLAD methods within the training of a single model, and find that this approach outperforms any individual technique.



%% file: 01-intro.tex
\section{Introduction}
\label{sec:intro}


Having access to a procedure which generates a visual datum reveals its underlying structure, facilitating high-level manipulation and editing by a person or autonomous agent.
Thus, inferring such programs from visual data is an important problem.
In $\Reals^2$, inferring shape programs has applications in the design of diagrams, icons, and other 2D graphics.
In $\Reals^3$, it has applications in reverse engineering of CAD models, procedural modeling for 3D games, and 3D structure understanding for autonomous agents.


We formally define shape program inference as obtaining a latent program $\z$ which generates a given observed shape $\x$. We model $p(\z | \x)$ with deep neural networks that train over a distribution of real shapes in order to amortize the cost of shape program inference on unseen shapes (e.g. a test set). 
This is a challenging problem: it is a structured prediction problem whose output is high-dimensional and can feature both discrete and continuous components (i.e. program control flow vs. program parameters).
Nevertheless, learning $p(\z | \x)$ becomes tractable provided that one has access to paired $(\X,\Z)$ data (i.e. a dataset of shapes and the programs which generate them) \cite{willis2020fusion}.
Unfortunately, while shape data is increasingly available in large quantities~\cite{chang2015shapenet}, these shapes do not typically come with their generating program.

To circumvent this data problem, researchers have typically \emph{synthesized} paired data by generating synthetic programs and pairing them with the shapes they output \cite{sharma2018csgnet, tian2019learning}.
However, as there is typically significant distributional mismatch between these synthetic shapes and ``real'' shapes from the distribution of interest, $\RealShapes$, various techniques must be employed to fine-tune $p(\z | \x)$ models towards $\RealShapes$.

A number of these fine-tuning strategies attempt to directly propagate gradients from geometric similarity measures back to $p(\z | \x)$. When the program executor is not a black-box, it may be possible to do this by implementing a differentiable relaxation of its behavior \cite{kania2020ucsgnet}, but this requires knowledge of its functional form. One can also try learning a differentiable proxy of the executor's behavior ~\cite{tian2019learning}, but this approximation introduces errors. Moreover, as shape programs typically involve many discrete structural decisions, training such a model end-to-end is usually infeasible in many domains.
Thus, many prior works often resort to general-purpose policy gradient reinforcement learning~\cite{sharma2018csgnet,ellis2019write}, which treats the program executor as a (non-differentiable) black-box. The downside of this strategy is that RL is notoriously unstable and slow to converge.

In this paper, we study a collection of methods that create (shape, program) data pairs used to train $p(\z | \x)$ models with maximum likelihood estimation (MLE) updates while treating the program executor as a black-box.
As discussed, ground-truth (shape, program) pairs are often unavailable, so these techniques must make compromises in how they formulate paired data. 
In wake-sleep, a generative model $p(\z)$ is trained to convergence on alternating cycles with respect to $p(\z | \x)$. 
When training $p(\z | \x)$, paired data can be created by sampling from $p(\z)$.
Each program label $\z$ is valid with respect to its associated $\x$ shape, but there is often a distributional mismatch between the generated set of shapes, $\X$, and shapes from the target distribution, $\RealShapes$.
In self-training, one uses $p(\z | \x)$ to infer latent $\z$'s for unlabeled input $\x$'s; these $\z$'s then become ``pseudo-labels'' which are treated as ground truth for another round of supervised training.
In this paradigm, there is no distributional shift between $\X$ and $\RealShapes$, but each $\z$ is only an approximately correctly label with respect to its paired $\x$.

We observe that shape program inference has a unique property that makes it especially well-suited for self-training: the distribution $p(\x | \z)$ is known \emph{a priori}---this is a delta distribution defined by the program executor.
When using a model $p(\z | \x)$ to infer a program $\z$ from some shape $\x^*$  of interest, one can use this executor to produce a shape $\x$ that is consistent with the program $\z$: in the terminology of self-training, $\z$ is guaranteed to be the ``correct label'' for $\x$.
However, similar to wake-sleep, formulating $\X$ as shape executions produced by model inferred programs can cause a distributional shift between $\X$ and $\RealShapes$.
Since this variant of self-training involves executing the inferred latent program $\z$, we call this procedure latent execution self-training (LEST).

As all of the aforementioned fine-tuning regimes use either Pseudo-Labels or Approximate Distributions to formulate (shape, program) pairs, we group them under a single conceptual framework: PLAD. We evaluate PLAD methods experimentally, using them to fine-tune shape program inference models in multiple shape domains: 2D and 3D constructive solid geometry (CSG), and assembly-based modeling with ShapeAssembly, a domain-specific language for structures of manufactured 3D objects~\cite{ShapeAssembly}.
We find that PLAD training regimes offer substantial advantages over the de-facto approach of policy gradient reinforcement learning, achieving better shape reconstruction performance while requiring significantly less computation time.
Further, we explore combining training updates from a mixture of PLAD methods, and find that this approach leads to better performance compared with any individual method.
Code for our method and experiments can be found at found at https://github.com/rkjones4/PLAD .

In summary, our contributions are:
\begin{packed_enumerate}
    \item Proposing the PLAD conceptual framework to group a family of related self-supervised learning techniques for shape program inference. 
    \item Introducing latent execution self-training, a PLAD method, to take advantage of the unique properties of the shape program inference problem.
    \item Experiments across multiple 2D and 3D shape program inference domains, demonstrating that (i) fine-tuning under PLAD regimes outperforms policy gradient reinforcement learning and (ii) combining PLAD methods is better than any individual technique.
\end{packed_enumerate}

%% file: 02-related.tex
\section{Related Work}
\label{sec:relwork}

\begin{table*}
    \centering
    \begin{tabular}{lcccc}
        \toprule
        \textbf{Method} 
        & \textbf{Models} 
        & \textbf{ Black-Box $p(\x | \z)$?} 
        & \textbf{$X = \RealShapes$ }
        & \parbox{8.0em}{\centering \textbf{Low variance, \\ unbiased gradients}}
        \\
        \midrule
        Policy gradient RL & $p(\z | \x)$ & $\checkmark$ &  $\checkmark$ & X
        \\
        Differentiable executor & $p(\z | \x)$ & X & $\checkmark$ & $\checkmark$
        \\
        Variational Bayes & $p(\z | \x)$, $p(\z)$ & X & $\checkmark$ & $\checkmark$ \\ 
        \hline
        Wake-sleep, EM & $p(\z | \x)$, $p(\z)$ & $\checkmark$ & X & $\checkmark$
        \\
        Self-training & $p(\z | \x)$ & $\checkmark$ & $\checkmark$ & X
        \\
        LEST & $p(\z | \x)$ & $\checkmark$ & X & $\checkmark$ 
        \\
        \bottomrule
    \end{tabular}
    \caption{
    Comparison of different methods for fine-tuning $p(\z | \x)$, in terms of the models that must be trained, if they treat the program executor as black-box, if their distribution of training shapes matches the distribution real shapes ($X = \RealShapes$), and if their loss gradients are unbiased with low-variance. The last three rows describe methods that fall under the PLAD framework.
    }
    \label{tab:method_comparison}
\end{table*}

Program synthesis is a broad field that has employed many techniques throughout its history. A program synthesizer takes as input a domain-specific language (DSL) and a specification; it outputs a program in the DSL that meets the specification.
Machine learning has been used to improve performance on program synthesis tasks by e.g. performing a neurally-guided search over all possible programs or letting a recurrent network predict program text directly \cite{robustFill,nsd, nsps, sun2018neural, Bunel2018, chen2018executionguided}.

In this work, we are interested in the sub-problem of \emph{shape program inference}, which is a type of \emph{visual program induction} problem~\cite{EG2020STAR}.
In our case, the specification is a visual representation of an object which the inferred program's output must geometrically match.
Here, we discuss prior work that has attacked this problem, organized by methodology used to learn $p(\z | \x)$;
see Table~\ref{tab:method_comparison} for an overview.
As discussed, a common practice in this prior work is to start with a model that has been pretrained on synthetically generated (shape, program) pairs with supervised learning, and then perform fine-tuning towards a distribution of interest \cite{ellis2018learning, liu2019learning, sharma2018csgnet, tian2019learning,ellis2019write}.

{\bf Policy Gradient Reinforcement Learning }
The most general method for fine-tuning a pretrained $p(\z | \x)$ is reinforcement learning: treating $p(\z | \x)$ as a policy network and using policy gradient methods~\cite{REINFORCE}.
The geometric similarity of the inferred program's output to its input is the reward function; the program executor $p(\x | \z)$ can be treated as a (non-differentiable) black-box.
CSG-Net uses RL for fine-tuning~\cite{sharma2018csgnet, sharma2020neural}, as does other recent work on inferring CSG programs from input geometry~\cite{ellis2019write}.
While CSG-Net has been improved to allow it to converge without supervised pretraining~\cite{zhou2020unsupervised}, not starting from the supervised model results in worse performance.
The main problem with policy gradient RL is its instability due to high variance gradients, leading to slow convergence.
Like RL, PLAD methods treat the program executor as a black-box, but as we show experimentally, they converge faster and achieve better reconstruction performance.

{\bf Differentiable Executor }
If the functional form of the program executor $p(\x | \z)$ is known and differentiable, then the gradient of the reward with respect to the parameters of $p(\z | \x)$ can be computed, making policy gradient unnecessary.
Shape programs are typically not fully differentiable, as they often involve discrete choices (e.g. which type of primitives to create).
UCSGNet uses a differentiable relaxation to circumvent this issue~\cite{kania2020ucsgnet}.
Other work trains a differentiable network to approximate the behavior of the program executor~\cite{tian2019learning}, which introduces errors.
PLAD regimes do not require the program executor to be differentiable, yet they perform better than other approaches (e.g. policy gradient RL) that share this desirable property.

{\bf Generative Model Learning }
Shape program inference has also been explored in the context of learning a generative model $p(\x,\z)$ of programs and the shapes they produce.
The most popular approach for training such models is variational Bayes, in particular the variational autoencoder~\cite{kingma2014auto}.
This method simultaneously trains a generative model $p(\x,\z)$ and a recognition model $p(\z | \x)$ by optimizing a lower bound on the marginal likelihood $p(\x)$.
When the $\z$'s are shape programs, the program executor is $p(\x | \z)$, so learning the generative model reduces to learning a prior over programs $p(\z)$.
Training such models with gradient descent requires that the executor $p(\x | \z)$ be differentiable.
When this is not possible, the wake-sleep algorithm is a viable alternative~\cite{WakeSleep}.
This approach alternates training steps of the generative and recognition models, training one on samples produced by the other.
Recent work has used wake-sleep for visual program induction~\cite{MemoizedWakeSleep,DreamCoder}.
If one trains the generative model and the inference model to convergence before switching to training the other, this is equivalent to expectation maximization (viewed as alternating maximization~\cite{EMmaxmax}).

{\bf Self-Training }
Traditionally, self-training has been employed in weakly-supervised learning paradigms to increase the predictive accuracy of simple classification models \cite{STOrig, yarowsky-1995-unsupervised, mcclosky-etal-2006-effective}.
Recently, renewed interest in self-training-inspired data augmentation approaches have demonstrated empirical performance improvements for neural models in domains such as large-scale image classification, machine translation, and speech recognition ~\cite{RethinkingSelfTraining, He2020Revisiting, kahn2020self}. 
But while self-training has been shown to yield practical gains for some domains, for others it can actually lead to worse performance, as training on too many incorrect pseudo-labels can cause learning to degrade \cite{EugBadST, steedman-etal-2003-bootstrapping}.
For self-training within the PLAD framework, the assigned pseudo-label for each example changes during fine-tuning whenever the inference model discovers a program that better explains the input shape;
similar techniques have been proposed for learning programs that perform semantic parsing under the view of iterative maximum likelihood \cite{liang2017neural}. 
To our knowledge, self-training has not been applied for fine-tuning visual program inference models, likely because it is somewhat unintuitive to view a program as a ``label" for a visual datum.

%% file: 03-method.tex
\section{Method}
\label{sec:method}

\newcommand{\MYCOMMENT}[1]{\STATE{\color{gray}{// #1}}}
\definecolor{ppt_purple}{rgb}{0.57,0.39,0.74}
\definecolor{ppt_blue}{rgb}{0.27,0.45,0.77}
\definecolor{ppt_orange}{rgb}{0.93,0.49,0.19}

\begin{figure*}[t!]
    \centering
    \begin{minipage}{0.4\linewidth}
    \begin{algorithm}[H]
    \textbf{Input:} $(\RealShapes, p(\z | \x), E)$\\
    \textbf{Output:} $p(\z | \x)$ fine-tuned on $\RealShapes$~
    
    \begin{algorithmic}
        \STATE $\Programs^{\text{BEST}} \gets \{\}$  
        \FOR{$\text{Number of Rounds}$}
        \MYCOMMENT{Update Best Programs}
        \STATE $\Programs^{\text{BEST}} \gets \text{inferProgs}(p(\z|\x), \RealShapes, \Programs^{\text{BEST}})$ 
        \MYCOMMENT{Create Training Data}
        \IF{$\color{ppt_blue}\text{Self-Train}$}
        \STATE $\Z \gets \Programs^{\text{BEST}}$
        \STATE $\X \gets \RealShapes$
        \ELSIF{$\color{ppt_orange}\text{LEST}$}
        \STATE $\Z \gets \Programs^{\text{BEST}}$
        \STATE $\X \gets \{ E(\z) \mid \z \in \Z \}$
        \ELSIF{$\color{ppt_purple}\text{Wake-Sleep}$}
        \STATE $p(\z) \gets \text{trainGenerative}( \Programs^{\text{BEST}}) $
        \STATE $\Z \gets \text{sample}(p(\z), |\RealShapes|)$
        \STATE $\X \gets \{ E(\z) \mid \z \in \Z \}$
        \ENDIF
        \MYCOMMENT{Train inference model}
        \STATE $p(\z | \x) \gets \text{trainMLE}(\X, \Z)$
        \ENDFOR
    
    \end{algorithmic}
\end{algorithm}
    \end{minipage}%
    \begin{minipage}{0.6\linewidth}
        \centering
        \includegraphics[width=\linewidth]{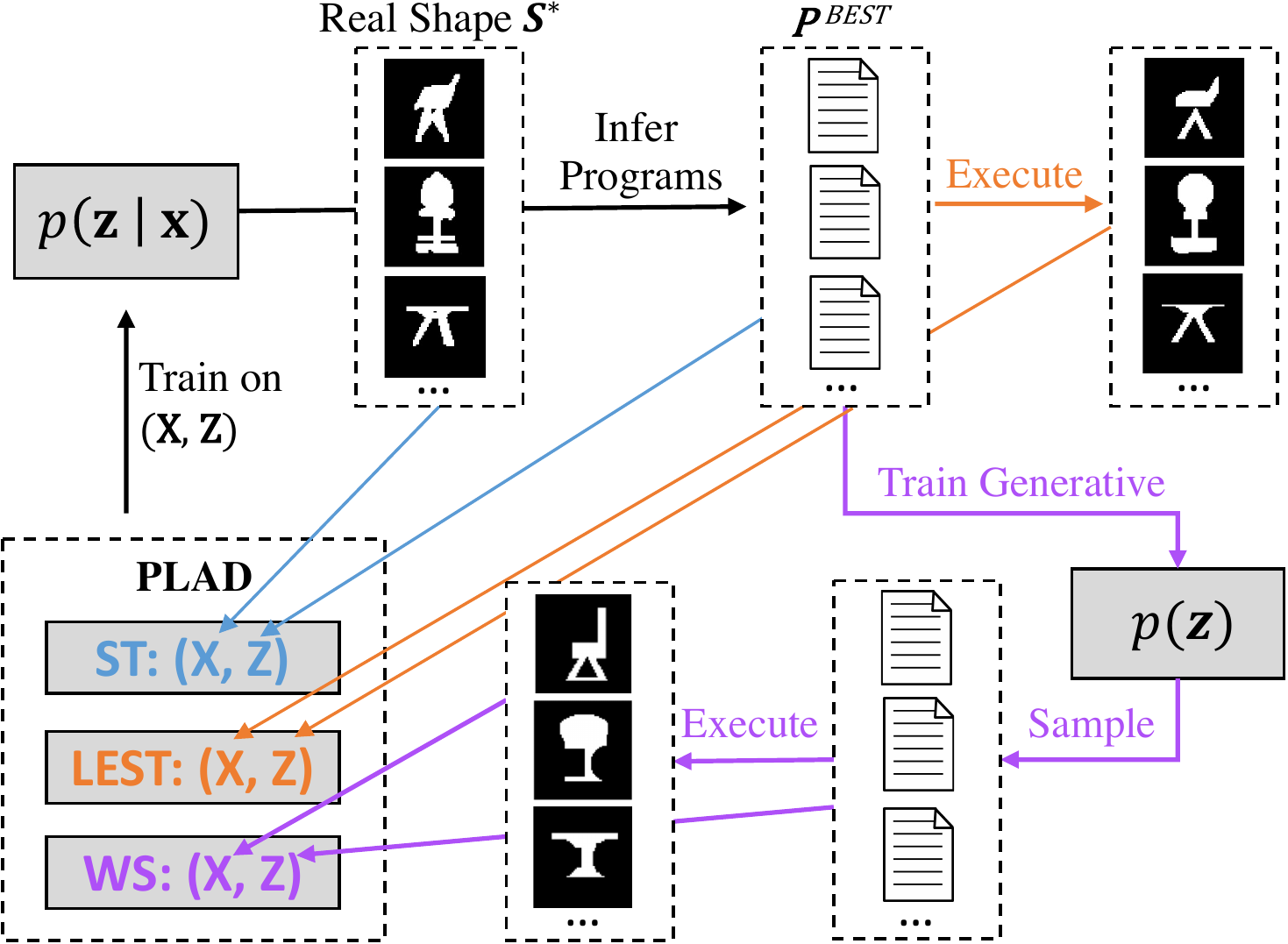}
    \end{minipage}
    \caption{\emph{(Left)} Pseudocode for fine-tuning shape program inference models, $p(\z|\x)$, towards a shape distribution of interest, $\RealShapes$, with Pseudo-Labels and Approximate Distributions (PLAD). PLAD methods iterate through three steps: infer programs for $\RealShapes$ with $p(\z|\x)$, create a dataset of $(\X, \Z)$ shape-program pairs, and train $p(\z|\x)$ on batches from $(\X, \Z)$. \textcolor{ppt_blue}{Self-training}, \textcolor{ppt_orange}{latent execution self-training}, and  \textcolor{ppt_purple}{wake-sleep} differ in how $(\X, \Z)$ is constructed. \emph{(Right)} A visual illustration of the algorithm's dataflow. }
    \label{figure:approach}
\end{figure*}

In this section, we describe the PLAD framework: a conceptual grouping of fine-tuning methods for shape program inference models. 
Our formulation assumes three inputs: a training dataset of shapes from the distribution of interest, $\RealShapes$, a program inference model, $p(\z|\x)$, and a program executor that converts programs into shapes, $E$. 
Throughout this paper, we assume that the $p(\z|\x)$ passed as input has undergone supervised pretraining on a distribution of synthetically generated shapes.
Methods within the PLAD framework return a fine-tuned $p(\z|\x)$ specialized to the distribution of interest from which $\RealShapes$ was sampled.

We depict the PLAD procedure both algorithimcally and pictorially in Figure \ref{figure:approach}. 
To fine-tune $p(\z|\x)$ towards $\RealShapes$, PLAD methods iterate through the following steps: (1) use $p(\z|\x)$ to find visually similar programs to $\RealShapes$, (2) construct a dataset of shape and program pairs $(\X, \Z)$ using the inferred programs, and (3) fine-tune $p(\z|\x)$ with maximum likelihood estimation updates on batches from $(\X, \Z)$. 
Through successive iterations, these steps bootstrap one another, forming a virtuous cycle: improvements to $p(\z|\x)$ create $(\X, \Z)$ pairs that more closely match the statistics of $\RealShapes$, and training on better $(\X, \Z)$ pairs specializes $p(\z|\x)$ to $\RealShapes$. 

Methods that fall within the PLAD framework differ in how the paired $(\X, \Z)$ data is created within each round. 
We detail this process for wake-sleep (Section \ref{sec:wake-sleep}) , self-training (Section \ref{sec:self-train}), and latent execution self-training (Section \ref{sec:lest}).
In Section \ref{sec:infer-progs} we explain our program inference procedure (inferProgs, Fig \ref{figure:approach}).
Finally, in Section \ref{sec:comb-plad} we discuss how a single $p(\z|\x)$ can be fine-tuned by multiple PLAD methods.

\subsection{Wake-Sleep $(X, Z)$ Construction}
\label{sec:wake-sleep}

Wake-sleep uses a generative model, $p(\z)$ to construct $(\X, \Z)$. 
In our implementation, we choose to model $p(\z)$ as a variational auto-encoder (VAE) \cite{kingma2014auto}, where the encoder consumes visual data and the decoder outputs a program. 
To create data for $p(\z)$, we take the current best programs discovered for $\RealShapes$, $\Programs^{\text{BEST}}$, and execute each program to form a set of shapes $\X^G$.
$p(\z)$ is then trained on pairs from $(\X^G, \Programs^{\text{BEST}})$ in the typical VAE framework. 
Note that the design space for $p(\z)$ is quite flexible, for instance, $p(\z)$ can trained without access to $\X^G$ if implemented with a program encoder.

Once $p(\z)$ has converged, we use it to sample $|\RealShapes|$ programs by decoding normally distributed random vectors. 
This set of programs becomes $\Z$, and $\X$ is formed by executing each program in $\Z$. 
In this set of $(\X, \Z)$ programs, $\Z$ is always the correct label for $\X$, so the gradient estimates during $p(\z|\x)$ training will be low-variance and unbiased. 
However, $\X$ is not guaranteed to be close to $\RealShapes$, it is only an approximate distribution. 
Note though, that as $\Programs^{\text{BEST}}$ better approximates $\RealShapes$, the distributional mismatch should become smaller, as long as the generative model has enough capacity to properly model $p(\z)$.

\subsection{Self-Training $(X, Z)$ Construction}
\label{sec:self-train}

Self-training constructs $(\X, \Z)$ by assigning labels from the current best program set, $\Programs^{\text{BEST}}$, to shape instances from $\RealShapes$.
Formally, $\X \gets \RealShapes$ and $\Z \gets \Programs^{\text{BEST}}$.
This framing maintains the nice property that $\X = \RealShapes$, so there will never be distributional mismatch between these two sets. 
The downside is that unless programs from $\Programs^{\text{BEST}}$ exactly recreate their paired shapes from $\RealShapes$ when executed, we know that the pseudo-labels from $\Programs^{\text{BEST}}$ are `incorrect'. 
From this perspective, we can consider gradient estimates that come from such $(\X, \Z)$ pairs to be biased.
However, as we will show experimentally, when $X$ forms a good approximation to $\RealShapes$, sourcing gradient estimates from these pseudo-labels leads to strong reconstruction performance. 

\subsection{LEST $(X, Z)$ Construction}
\label{sec:lest}

A unique property of shape program inference is that the distribution $p(\x | \z)$ is readily available in the form of the program executor $E$.
We leverage this property to propose LEST, a variant of self-training that does not create mismatch between pseudo-labels and their associated visual data.
Similar to the self-training paradigm, LEST first constructs $\Z$ as the current best program set, $\Programs^{\text{BEST}}$.
Then, differing from self-training, LEST constructs $\X$ as the executed version of each program in $\Z$. 
By construction, the labels in $\Z$ will now be correct for their paired shapes in $\X$. 
The downside is that, like wake-sleep, LEST may introduce a distributional mismatch between $\X$ and $\RealShapes$. 
But once again, as $\Programs^{\text{BEST}}$ better approximates $\RealShapes$, the mismatch between the two distributions will decrease. 

\subsection{Inferring Programs with $p(z|x)$}
\label{sec:infer-progs}

During each round of fine-tuning, PLAD methods rely on $p(\z|\x)$ to infer programs that approximate $\RealShapes$.
We propose to train PLAD methods such that the best matching inferred programs for $\RealShapes$ are maintained across rounds. 
Specifically, we construct a data structure $\Programs^{\text{BEST}}$ that maintains a program for each training shape in $\RealShapes$.
In this way, as more iterations are run, $\Programs^{\text{BEST}}$ always forms a closer approximation to  $\RealShapes$.
There is an alternative framing, where $\Programs^{\text{BEST}}$ is reset each epoch, but we show experimental results in the supplemental material that this can lead to worse generalization.

To update $\Programs^{\text{BEST}}$ each round, we employ an inner-loop search procedure.
For each shape in $\RealShapes$, $p(\z|\x)$ suggests high-likelihood programs, and the $\Programs^{\text{BEST}}$ entry is updated to keep the program whose execution obtains the highest similarity to the input shape; the specific similarity metric varies by domain.
While there are many ways to structure this inner-loop search, we choose beam-search, as we find it offers a good trade-off between speed and performance.
Experimentally, we demonstrate that PLAD methods are capable of performing well even as the time spent on inner-loop search is varied (Section \ref{sec:abl_search}).

\subsection{Training $p(z|x)$ with multiple PLAD methods}
\label{sec:comb-plad}

As detailed in the preceding sections, the main difference between PLAD approaches is in how they construct the $(\X, \Z)$ dataset used for fine-tuning $p(\z|\x)$. 
However, there is no strict requirement that these different $(\X, \Z)$ distributions be kept separate. 
We explore how $p(\z|\x)$ behaves under fine-tuning from multiple PLAD methods, such as combining LEST and self-training.
We implement these mixtures on a per-batch basis.
Before $p(\z|\x)$ training, we construct distinct $(\X, \Z)$ distributions for each method in the combination.
Then, during training, each batch is randomly sampled from one of the $(\X, \Z)$ distributions.
We experimentally validate the effectiveness of this approach in the next section.

%% file: 04-results.tex
\section{Results}
\label{sec:results}

\begin{table*}[t!]
    \centering    
    \setlength{\tabcolsep}{10pt}
    \begin{tabular}{@{}l ccc@{}}
        \toprule
        \textbf{Method} & \textbf{2D CSG CD} $\Downarrow$ & \textbf{3D CSG IoU }$\Uparrow$ & \textbf{ShapeAssembly IoU} $\Uparrow$ \\
        \midrule
        Supervised Pretraining (SP) & 1.580 & 41.0 & 37.6 \\
        \midrule
        REINFORCE (RL)     & 1.097  &  53.4   &   50.8    \\
        Wake-Sleep (WS)    & 1.118  &  67.4   &   57.2   \\
        Self-Training (ST) & 0.841  &  67.3   &   61.3   \\
        LEST               & 0.976  &  69.8   &   56.5 \\
        \midrule
        LEST+ST            & 0.829  &  70.8   &   66.0    \\
        LEST+ST+WS         & \textbf{0.811} & \textbf{74.3} & \textbf{66.4} \\
        \bottomrule
    \end{tabular}
    \caption{Test-set reconstruction performance across multiple shape program inference domains. The top row contains results for the pretrained $p(\z|\x)$ model fine-tuned by the other methods. For 2D CSG the metric is Chamfer distance (CD, lower is better). For 3D CSG and ShapeAssembly the metric is intersection over union (IoU, higher is better). Individual PLAD methods outperform RL, and combining PLAD methods achieves the best performance across all domains (LEST+ST+WS).}
    \label{tab:model_perf}
    \vspace{-2.0mm}
\end{table*}

We evaluate a series of methods on their ability to fine-tune shape program inference models across multiple domains. 
We describe the different domains in Section \ref{sec:domains} and details of our experimental design in Section \ref{sec:exp-design}.
In Section \ref{sec:recon}, we compare the reconstruction accuracy of each method, and study how they are affected by varying the time spent on inner-loop search (Section \ref{sec:abl_search}) and the size of the training set (Section \ref{sec:abl_size}).
Finally, we explore the convergence speed of each method in Section \ref{sec:speed}.

\subsection{Shape Program Domains}
\label{sec:domains}

We run experiments across three shape program domains: 2D Constructive Solid Geometry (CSG), 3D CSG, and ShapeAssembly. Details can be found in the supplemental.

In CSG, shapes are created by declaring parametric primitives (e.g. circles, boxes) and combining them with Boolean operations (union, intersection, difference).
CSG inference is non-trivial: as CSG uses non-additive operations (intersection, difference), inferring a CSG program does not simply reduce to primitive detection.
For 2D CSG, we follow the grammar defined by CSGNet \cite{sharma2018csgnet}, using 400 shape tokens that correspond to randomly placed circles, triangles and rectangles on a 64 x 64 grid.
For 3D CSG, we use a grammar that has individual tokens for defining primitives (ellipsoids and cuboids), setting primitive attributes (position and scales), and the three Boolean operators. Attributes are discretized into 32 bins.

ShapeAssembly is designed for specifying the part structure of manufactured 3D objects. It creates objects by declaring cuboid part geometries and assembling those parts together via attachment and symmetry operators.
Our grammar contains tokens for each command type and parameter value; to handle continuous values, we discretize them into 32 bins.

\subsection{Experimental Design}
\label{sec:exp-design}

{\bf Fine-Tuning Methods }
We compare the ability of the following training schemes to fine-tune a model on a specific domain of interest:
\begin{packed_itemize}
    \item \textbf{SP:} $p(\z|\x)$ with supervised pretraining.
    \item \textbf{RL:} Fine-tuning with REINFORCE.
    \item \textbf{WS:} Fine-tuning with wake-sleep.
    \item \textbf{ST:} Fine-tuning with self-training.
    \item \textbf{LEST:} Fine-tuning with latent execution self-training.
    \item \textbf{LEST+ST:} combining LEST and ST. 
    \item \textbf{LEST+ST+WS:} combining LEST, ST and WS.
\end{packed_itemize}

{\bf Shape Datasets }
~~ Fine-tuning methods learn to specialize $p(\z|\x)$ against a distribution of real shapes $\RealShapes$. For each domain, we construct a dataset of shapes $\RealShapes$, and split it into train, validation, and test sets. We perform early-stopping with respect to the validation set. 
For 2DCSG, we use the CAD dataset from CSGNet~\cite{sharma2018csgnet}, which consists of front and side views of chairs, desks, and lamps from the Trimble 3D warehouse. We split the dataset into $10K$ shapes for training, $3K$ shapes for validation, and $3K$ shapes for testing.
For 3D CSG and ShapeAssembly, we use CAD shapes from the chair, table, couches, and benches categories of ShapeNet; voxelizations are provided by \cite{chen2019bae_net}. We split the dataset into $10K$ shapes for training, $1K$ shapes for validation, and $1K$ shapes for testing.

{\bf Model Architectures }
For all experiments, we model $p(\z|\x)$ in an encoder-decoder framework, although the particular architectures vary by domain.
In all cases, the encoder is a CNN that converts visual data into a latent variable, and the decoder is an auto-regressive model that decodes the latent variables into a sequence of tokens.
For 2D CSG, we use the same $p(\z|\x)$ architecture as CSGNet.
A CNN consumes $64 \times 64$ binary mask shape images to produce a latent code that initializes a GRU-based recurrent decoder.
For 3D CSG and ShapeAssembly, we use a 3D CNN that consumes $32 \times 32 \times 32$ occupancy voxels. This CNN outputs a latent code that is attended to by a Transformer decoder network \cite{10.5555/3295222.3295349}; this network also attends over token sequences in a typical auto-regressive fashion.

{\bf Supervised Pretraining }
Before fine-tuning, $p(\z|\x)$ undergoes supervised pretraining on synthetically generated programs until it has converged on that set.
For 2D CSG, we follow CSGNet's approach.
For 3D CSG, we construct valid programs by (i) sampling a set of primitives within the allotted grid (ii) identifying potential overlaps (iii) constructing a binary tree of boolean operations using these overlaps. 
For ShapeAssembly, we propose programs by sampling random grammar expansions according to the language's typing system. 
We then employ a validation step where a program is rejected if any of its part are not the sole occupying part of at least 8 voxels (to discourage excessive part overlaps). 
For 3D CSG and ShapeAssembly we sample 2 million synthetic programs and train until convergence on a validation set of 1000 programs. 
Full details provided in the supplemental.

\begin{figure*}[]
    \centering
   \includegraphics[width=\linewidth]{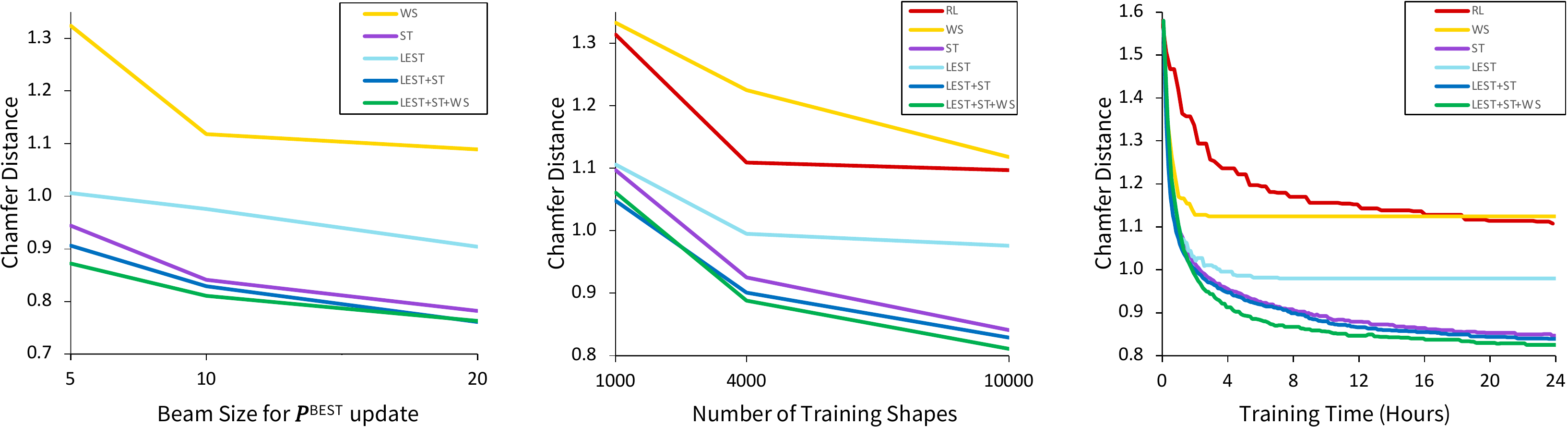}
    \caption{Experiments exploring properties of PLAD methods on 2D CSG. On the X-axis we plot the beam size used during the $\Programs^{\text{BEST}}$ update \emph{(Left)}, the number of training shapes \emph{(Middle)}, and the training time \emph{(Right)}. 
    The Y-axis of each plot measures reconstruction accuracy on test-set shapes.}
    \label{fig:plot_beam_size} 
\end{figure*}

\begin{figure*}[t!]
    \centering
    \footnotesize
    \setlength{\tabcolsep}{1pt}
    \begin{tabular}{ccccccccc}
        & \textbf{SP} &  \textbf{WS} &  \textbf{RL} &  \textbf{ST} &  \textbf{LEST} & \textbf{LEST+ST} & \textbf{LEST+ST+WS} & \textbf{Target} \\ \\
        \multirow{2}{*}{{\rotatebox{90}{\textbf{2D CSG}}}} 
        &
        \includegraphics[trim={2.5cm 1.0cm 2.5cm, 1.0cm}, clip,{width=.115\linewidth}]{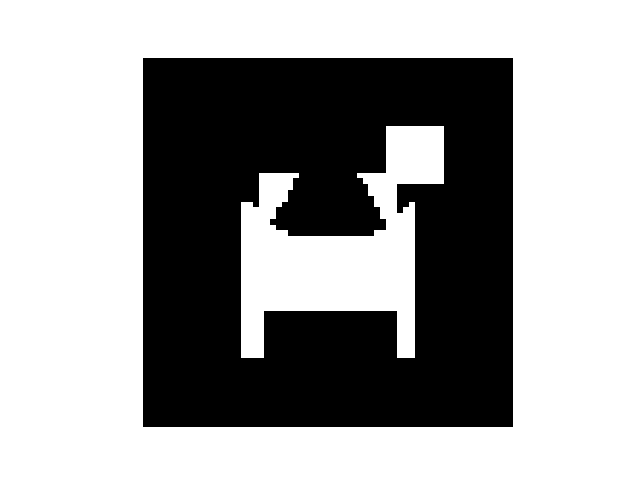} &
        \includegraphics[trim={2.5cm 1.0cm 2.5cm, 1.0cm}, clip,{width=.115\linewidth}]{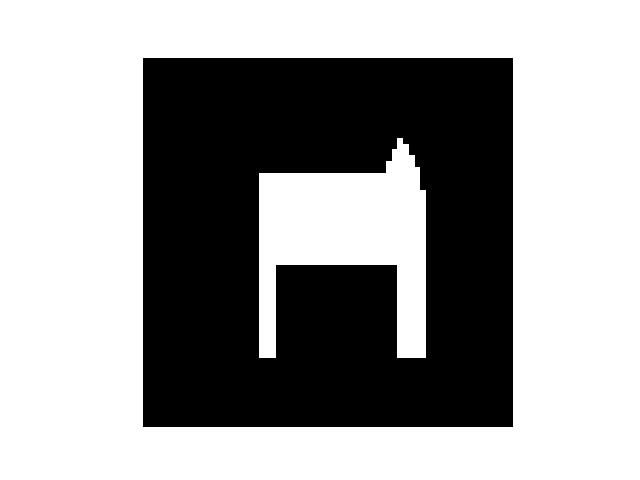} &
        \includegraphics[trim={2.5cm 1.0cm 2.5cm, 1.0cm}, clip,{width=.115\linewidth}]{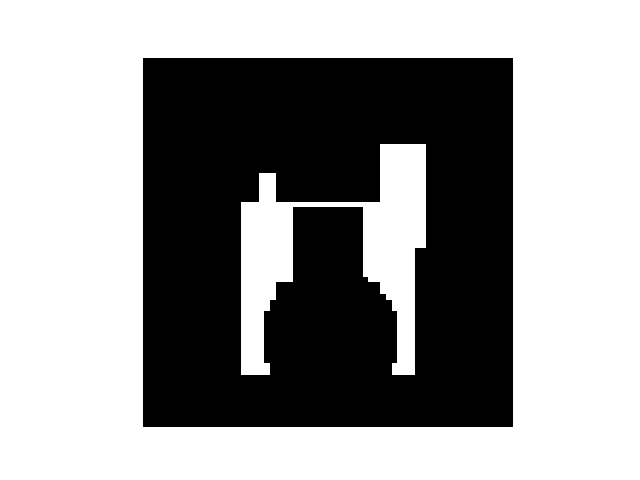} &
        \includegraphics[trim={2.5cm 1.0cm 2.5cm, 1.0cm}, clip,{width=.115\linewidth}]{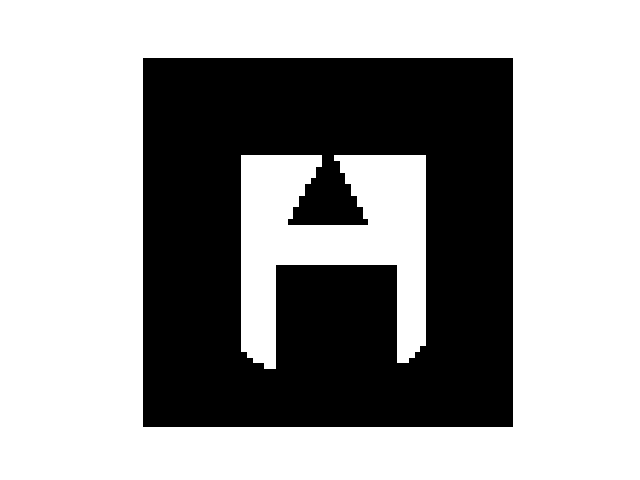} &
        \includegraphics[trim={2.5cm 1.0cm 2.5cm, 1.0cm}, clip,{width=.115\linewidth}]{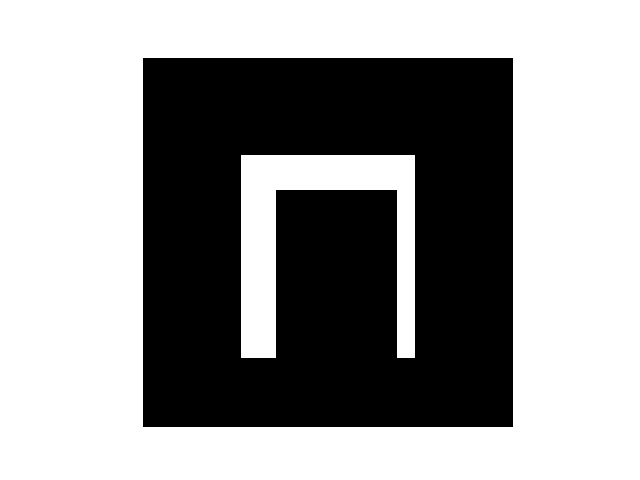} &
        \includegraphics[trim={2.5cm 1.0cm 2.5cm, 1.0cm}, clip,{width=.115\linewidth}]{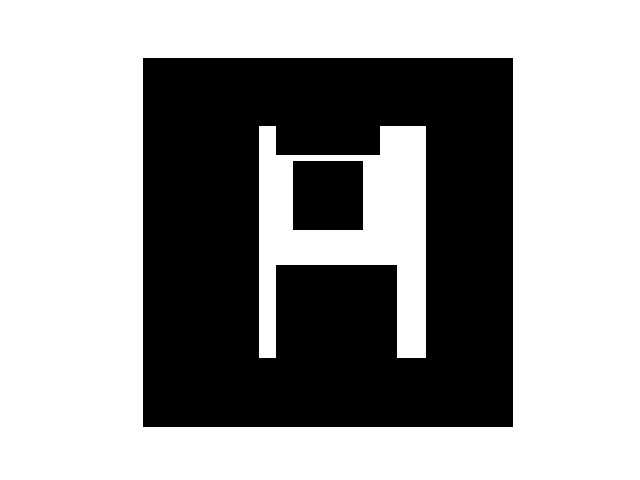} &
        \includegraphics[trim={2.5cm 1.0cm 2.5cm, 1.0cm}, clip,{width=.115\linewidth}]{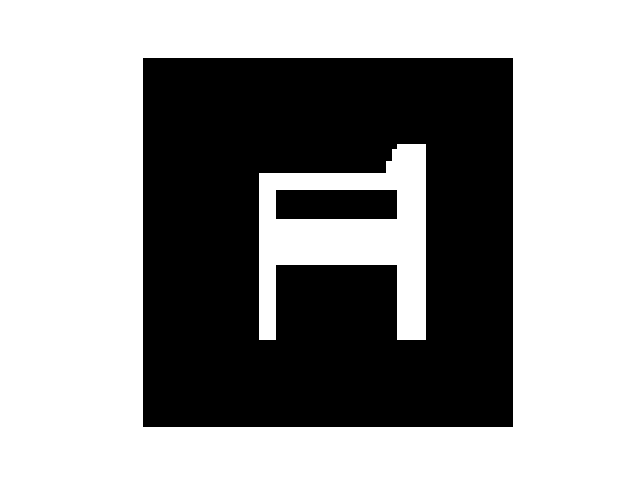} &
        \includegraphics[trim={2.5cm 1.0cm 2.5cm, 1.0cm}, clip,{width=.115\linewidth}]{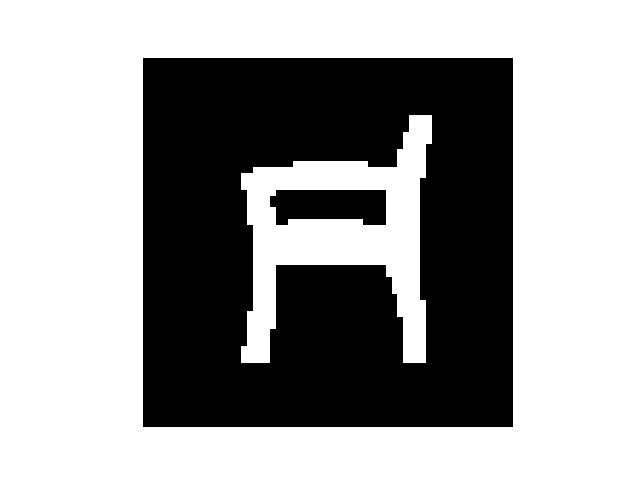} \\
        &
        \includegraphics[trim={2.5cm 1.0cm 2.5cm, 1.0cm}, clip,{width=.115\linewidth}]{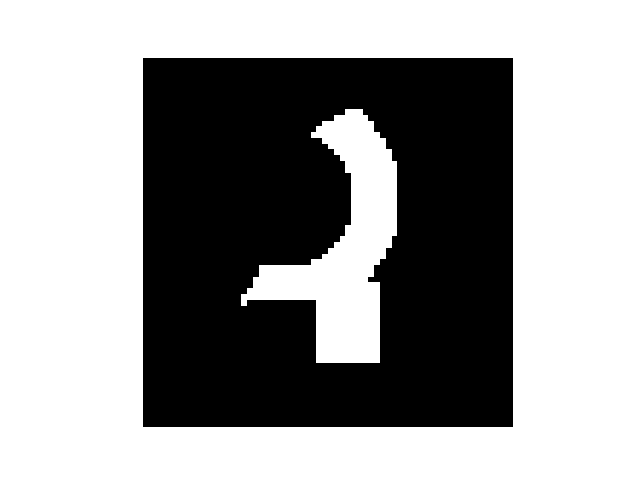} &
        \includegraphics[trim={2.5cm 1.0cm 2.5cm, 1.0cm}, clip,{width=.115\linewidth}]{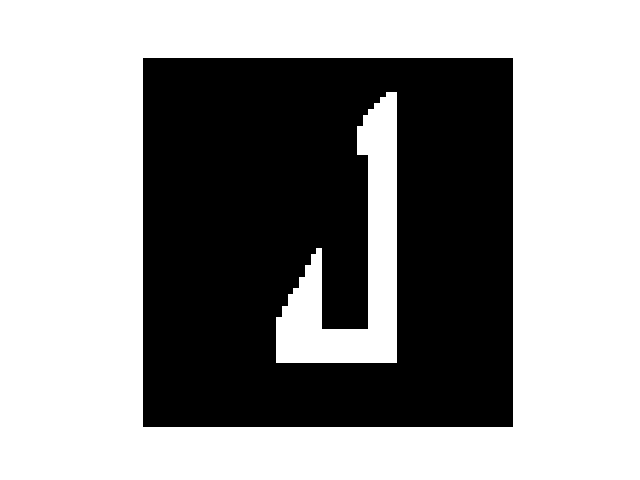} &
        \includegraphics[trim={2.5cm 1.0cm 2.5cm, 1.0cm}, clip,{width=.115\linewidth}]{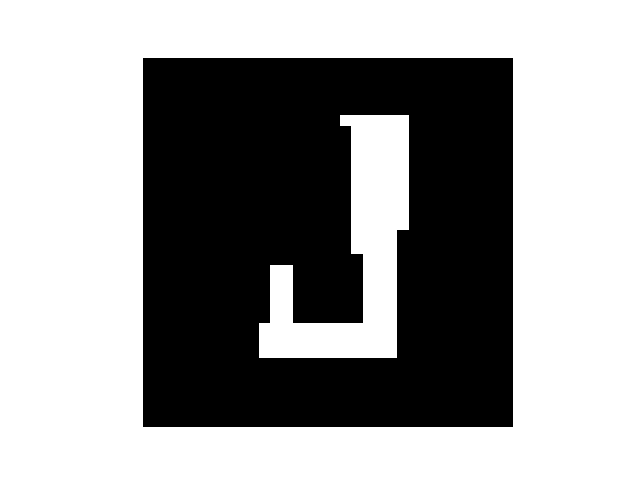} &
        \includegraphics[trim={2.5cm 1.0cm 2.5cm, 1.0cm}, clip,{width=.115\linewidth}]{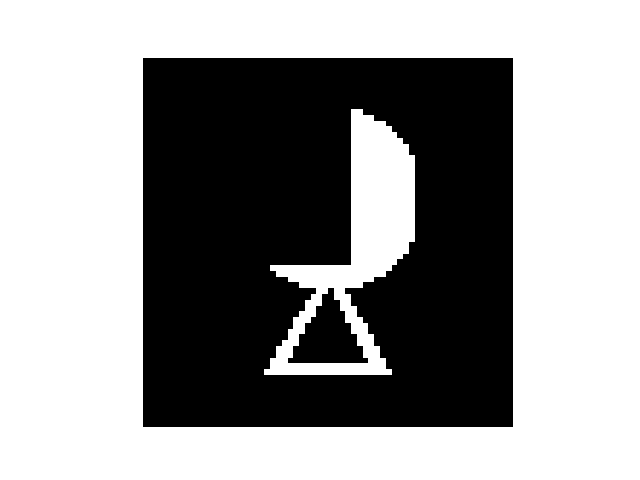} &
        \includegraphics[trim={2.5cm 1.0cm 2.5cm, 1.0cm}, clip,{width=.115\linewidth}]{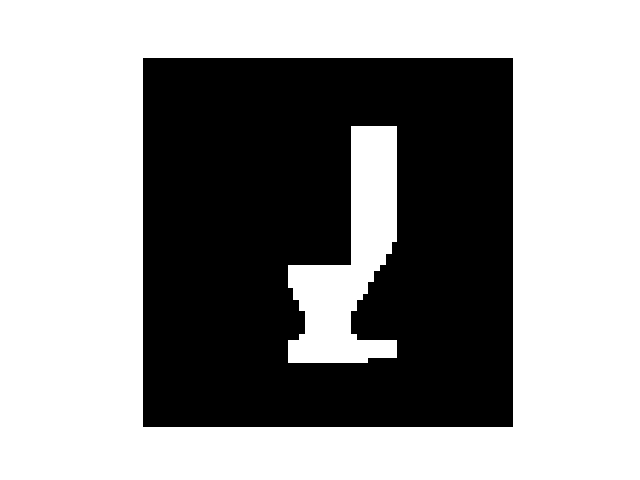} &
        \includegraphics[trim={2.5cm 1.0cm 2.5cm, 1.0cm}, clip,{width=.115\linewidth}]{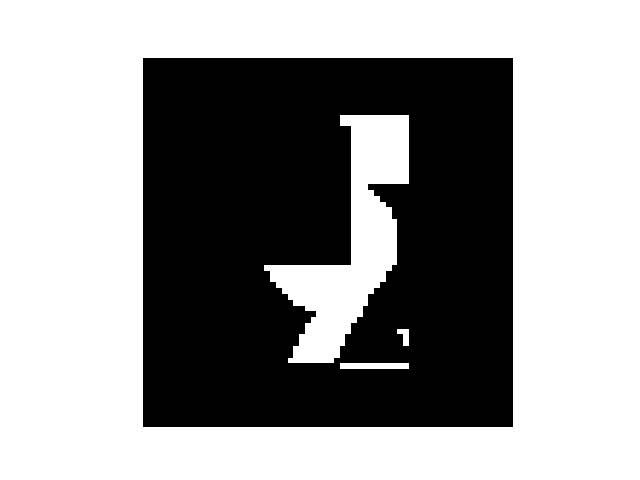} &
        \includegraphics[trim={2.5cm 1.0cm 2.5cm, 1.0cm}, clip,{width=.115\linewidth}]{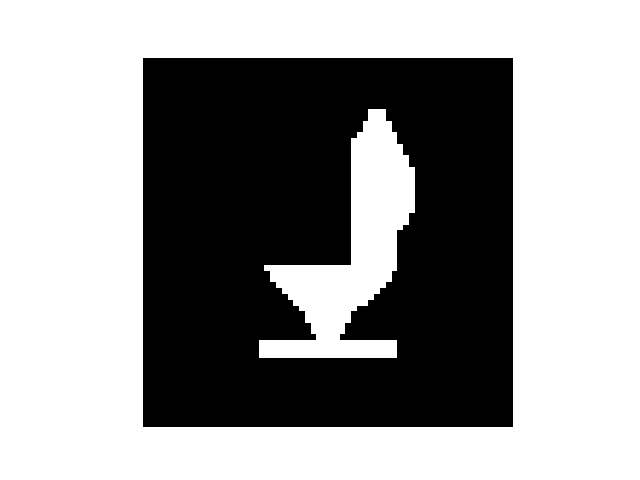} &
        \includegraphics[trim={2.5cm 1.0cm 2.5cm, 1.0cm}, clip,{width=.115\linewidth}]{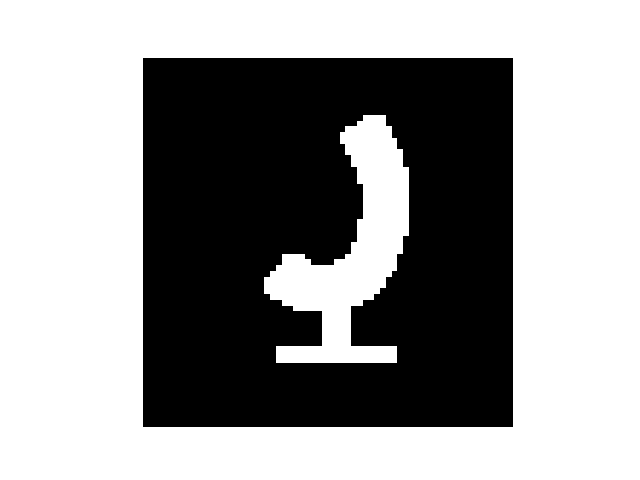} \\
        \\
        \multirow{2}{*}{{\rotatebox{90}{\textbf{3D CSG}}}} 
        &
        \includegraphics[trim={2.5cm 2.5cm 2.5cm 2.5cm},clip,{width=.115\linewidth}]{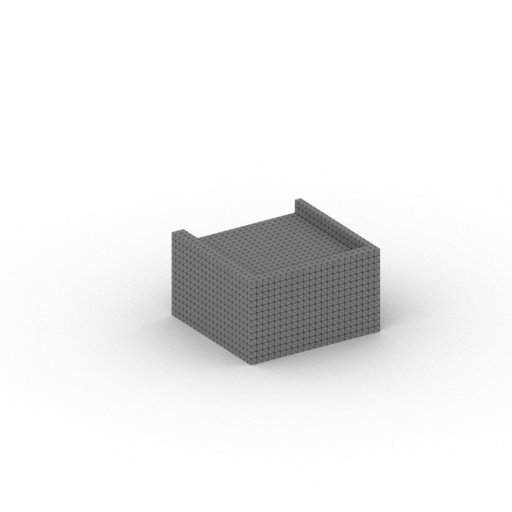} &
        \includegraphics[trim={2.5cm 2.5cm 2.5cm 2.5cm},clip,{width=.115\linewidth}]{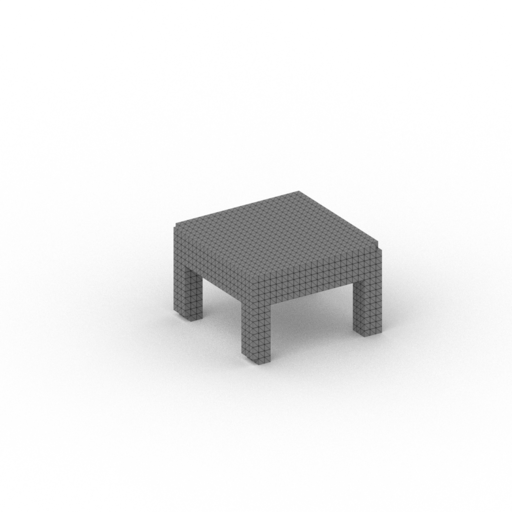} &
        \includegraphics[trim={2.5cm 2.5cm 2.5cm 2.5cm},clip,{width=.115\linewidth}]{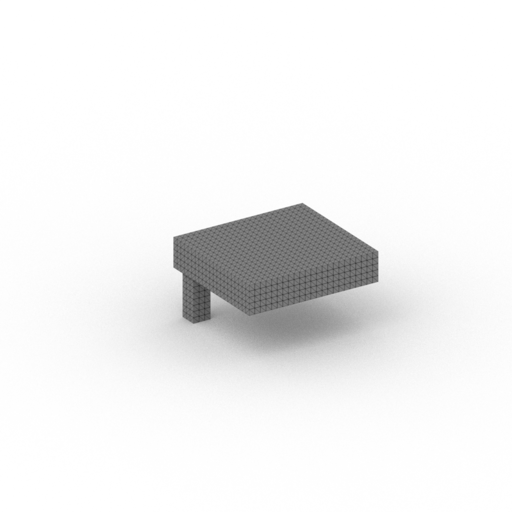} &
        \includegraphics[trim={2.5cm 2.5cm 2.5cm 2.5cm},clip,{width=.115\linewidth}]{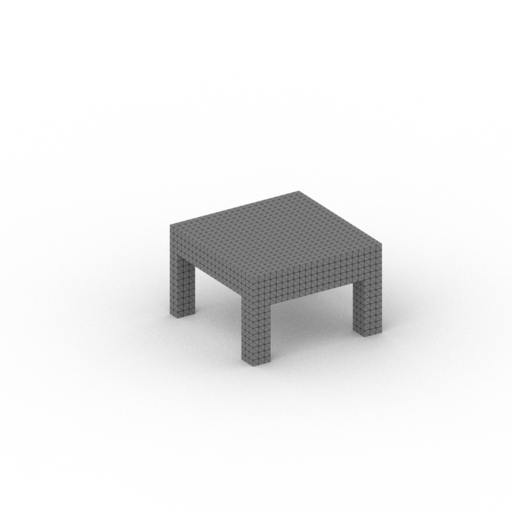} &
        \includegraphics[trim={2.5cm 2.5cm 2.5cm 2.5cm},clip,{width=.115\linewidth}]{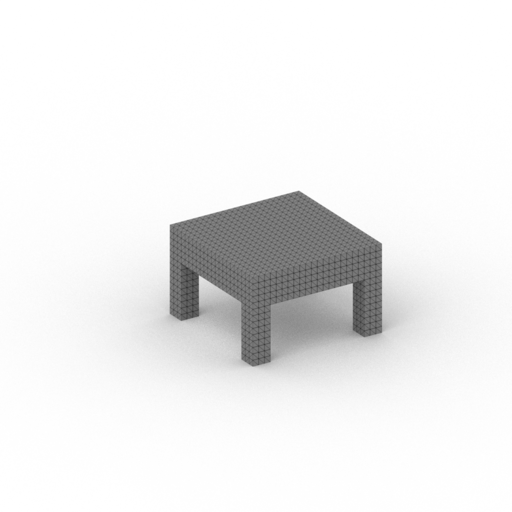} &
        \includegraphics[trim={2.5cm 2.5cm 2.5cm 2.5cm},clip,{width=.115\linewidth}]{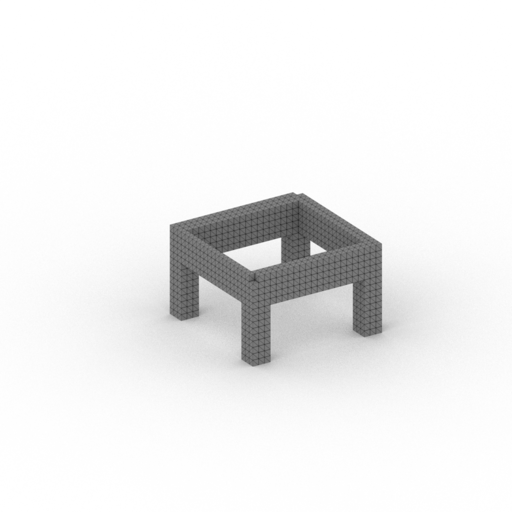} &
        \includegraphics[trim={2.5cm 2.5cm 2.5cm 2.5cm},clip,{width=.115\linewidth}]{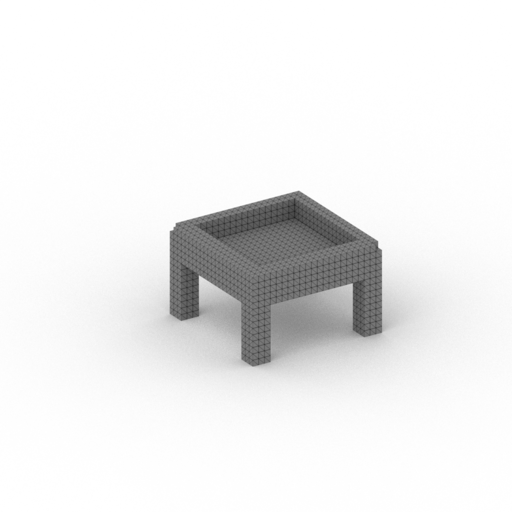} &
        \includegraphics[trim={2.5cm 2.5cm 2.5cm 2.5cm},clip,{width=.115\linewidth}]{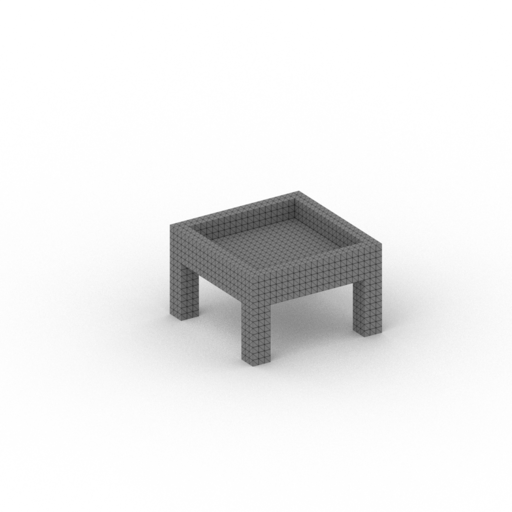} \\
        &
        \includegraphics[trim={3.5cm 2.5cm 3.5cm 2.5cm},clip,{width=.115\linewidth}]{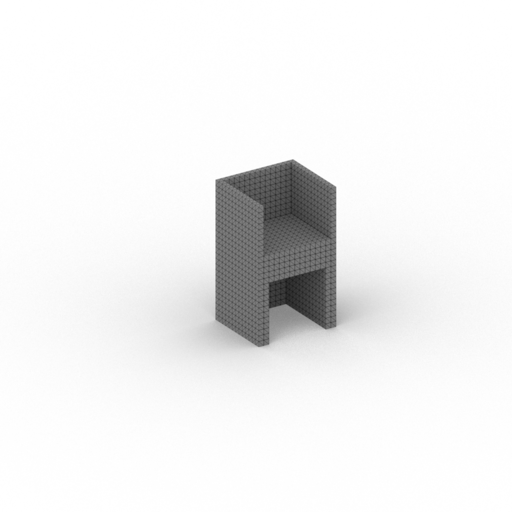} &
        \includegraphics[trim={3.5cm 2.5cm 3.5cm 2.5cm},clip,{width=.115\linewidth}]{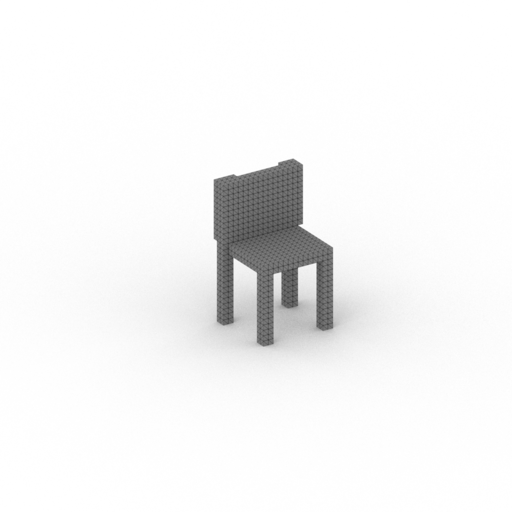} &
        \includegraphics[trim={2.5cm 2.5cm 2.5cm 2.5cm},clip,{width=.115\linewidth}]{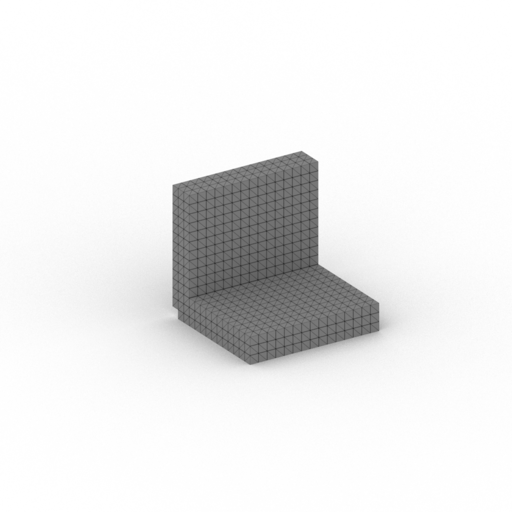} &
        \includegraphics[trim={3.5cm 2.5cm 3.5cm 2.5cm},clip,{width=.115\linewidth}]{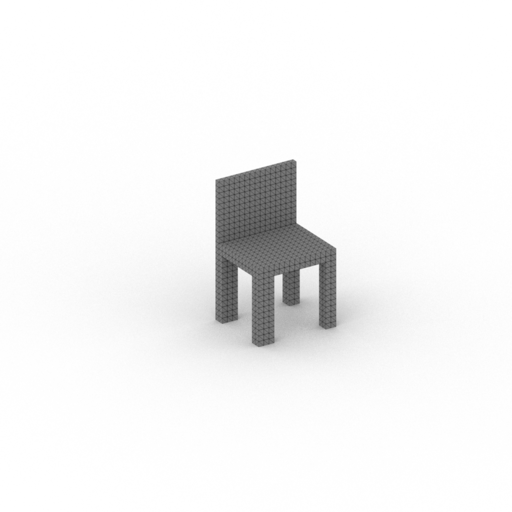} &
        \includegraphics[trim={3.5cm 2.5cm 3.5cm 2.5cm},clip,{width=.115\linewidth}]{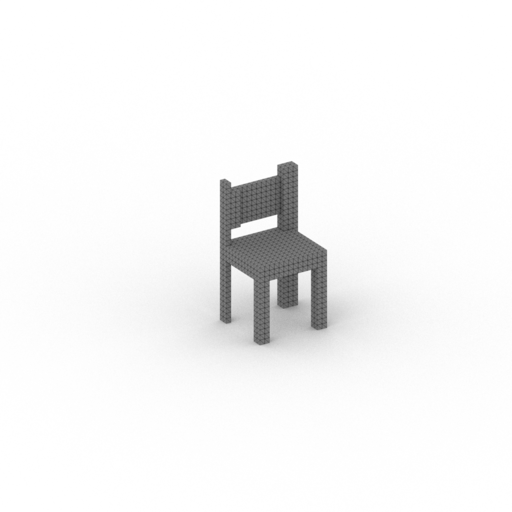} &
        \includegraphics[trim={3.5cm 2.5cm 3.5cm 2.5cm},clip,{width=.115\linewidth}]{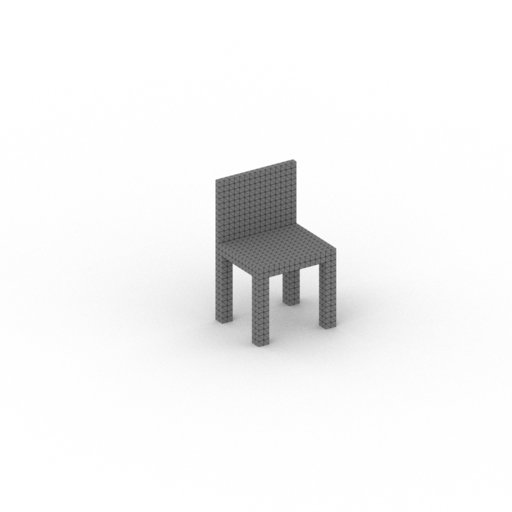} &
        \includegraphics[trim={3.5cm 2.5cm 3.5cm 2.5cm},clip,{width=.115\linewidth}]{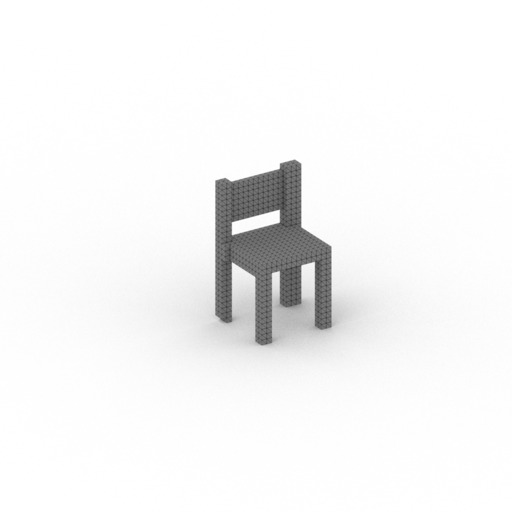} &
        \includegraphics[trim={3.5cm 2.5cm 3.5cm 2.5cm},clip,{width=.115\linewidth}]{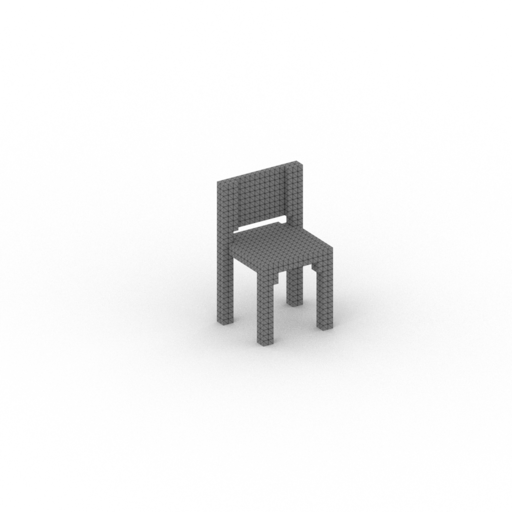}
        \\
        \multirow{2}{*}{{\rotatebox{90}{\textbf{SA}}}} 
        &
        \includegraphics[trim={.5cm .5cm .5cm .5cm},clip,{width=.115\linewidth}]{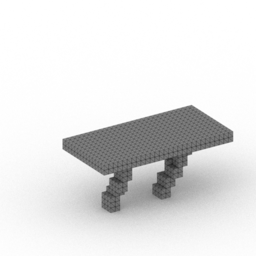} &
        \includegraphics[trim={.5cm .5cm .5cm .5cm},clip,{width=.115\linewidth}]{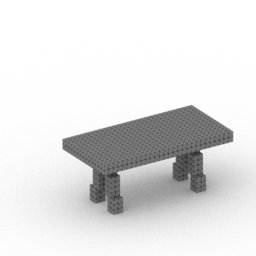} &
        \includegraphics[trim={.5cm .5cm .5cm .5cm},clip,{width=.115\linewidth}]{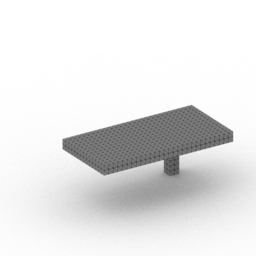} &
        \includegraphics[trim={.5cm .5cm .5cm .5cm},clip,{width=.115\linewidth}]{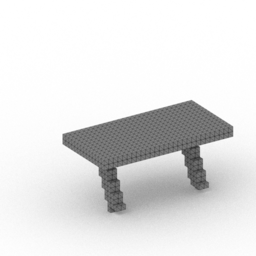} &
        \includegraphics[trim={.5cm .5cm .5cm .5cm},clip,{width=.115\linewidth}]{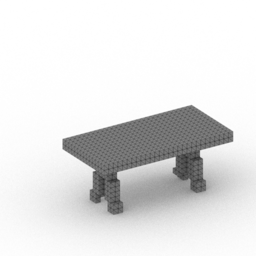} &
        \includegraphics[trim={.5cm .5cm .5cm .5cm},clip,{width=.115\linewidth}]{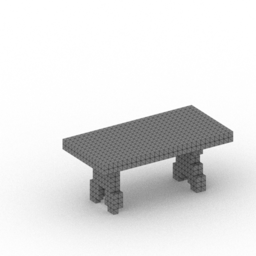} &
        \includegraphics[trim={.5cm .5cm .5cm .5cm},clip,{width=.115\linewidth}]{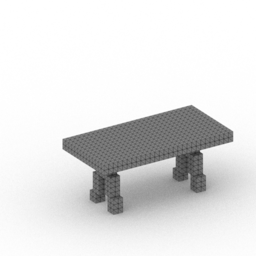} &
        \includegraphics[trim={.5cm .5cm .5cm .5cm},clip,{width=.115\linewidth}]{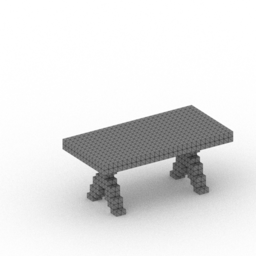} \\
        & 
        \includegraphics[{width=.115\linewidth}]{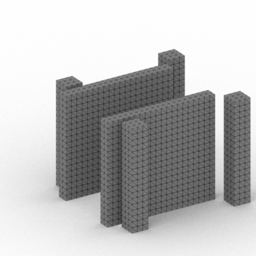} &
        \includegraphics[{width=.115\linewidth}]{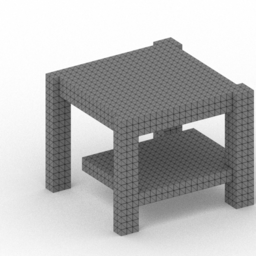} &
        \includegraphics[{width=.115\linewidth}]{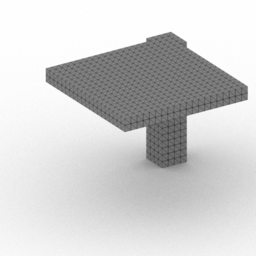} &
        \includegraphics[{width=.115\linewidth}]{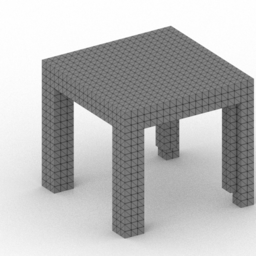} &
        \includegraphics[{width=.115\linewidth}]{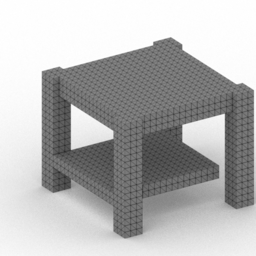} &
        \includegraphics[{width=.115\linewidth}]{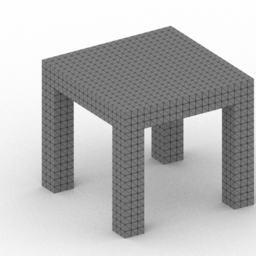} &
        \includegraphics[{width=.115\linewidth}]{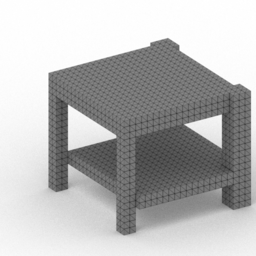} &
        \includegraphics[{width=.115\linewidth}]{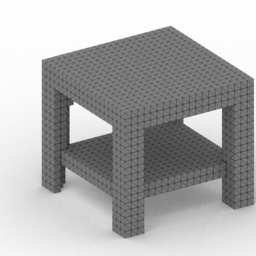} \\
        \\
    \end{tabular}
    \caption{Qualitative comparisons of shape programs inferred for test-set shapes made by different fine-tuning methods for 2D CSG \emph{(Top)}, 3D CSG \emph{(Middle)}, and ShapeAssembly \emph{(Bottom)}. We provide additional qualitative results in the supplemental. } 
    \label{fig:qual_seg_comp}
\end{figure*}

\subsection{Reconstruction Accuracy }
\label{sec:recon}

We evaluate the performance of each fine-tuning method according to reconstruction accuracy: how closely the output of a shape program matches the input shape from which it was inferred, on a held out set of test shapes. The specific metric varies by domain.
For 2D CSG, we follow CSGNet and use Chamfer Distance (CD), where lower distances indicate more similar shapes.
For 3D CSG and ShapeAssembly, we use volumetric intersection over union (IoU).

For each domain, we run each fine-tuning method to convergence, starting with the same $p(\z|\x)$ model that has undergone supervised pretraining. 
The reward for RL models follows the similarity metric in each domain: CD for 2D CSG; IoU for 3D CSG and ShapeAssembly.
For PLAD fine-tuning methods, the similiarity metric in each domain determines which program is kept during updates to $\Programs^{\text{BEST}}$.
At inference time, when evaluating the reconstruction performance of each $p(\z|\x)$, we employ a beam search procedure, decoding multiple programs in parallel, and choosing the program that achieves the highest similarity to the target shape. We use a beam size of 10, unless otherwise stated.

We present quantitative results in Table \ref{tab:model_perf}.
Looking at the middle four rows, the two self-training variants (ST and LEST) outperform RL as a fine-tuning method in all the domains we studied.
The wake-sleep variant (WS) also outperforms RL for both 3D CSG and ShapeAssembly. These are more challenging domains with larger token spaces, posing difficulties for policy gradient fine-tuning.
As demonstrated by the last two rows, further improvement can be had by combining multiple methods: for each domain, the best performance is achieved by LEST+ST+WS. 
In fact, for 2DCSG, the test-set reconstruction accuracy achieved by LEST+ST+WS (0.811) outperforms previous state-of-the-art results, CSGNet (1.14) \cite{sharma2018csgnet} and CSGNetStack (1.02) \cite{sharma2020neural}, for paradigms where the executor is treated as a black-box. 

Mixing updates from multiple PLAD methods is beneficial because, in this joint paradigm, each method can cover the other's weaknesses.
For instance, employing ST ensures that some samples of $\X$ are sourced from $\RealShapes$, and employing LEST ensures that some samples of $\X$ have paired $\Z$ programs which are exact labels. 
We present qualitative results in Figure \ref{fig:qual_seg_comp}. 
The reconstructions from the PLAD combination methods better reflect the input shapes, reinforcing the quantitative trends.

\subsection{Inner-loop Search Time}
\label{sec:abl_search}

PLAD methods make use of $\Programs^{\text{BEST}}$ to generate $(\X, \Z)$ datasets that train $p(\z|\x)$. 
To study how time spent on inner-loop search affects each technique, we ran an experiment using different beam sizes to update $\Programs^{\text{BEST}}$.
We present results in Figure \ref{fig:plot_beam_size}, left.
On the X-axis we plot beam size; on the Y-axis we plot test-set reconstruction Chamfer distance.
Unsurprisingly, spending more time on the inner-loop search leads to better performance; finding better programs for training shapes improves test time generalization. 
That said, across all beam sizes, we find that it is always best to train under a combination of PLAD methods; the LEST+ST and the LEST+ST+WS variants always outperform any individual fine-tuning scheme.
Note, RL is not included in this experiment because REINFORCE, as defined, has no inner-loop search mechanism.
In this way, PLAD provides an additional control lever, where time spent on inner-loop search modulates a trade-off between convergence speed and test-set reconstruction performance.

\subsection{Number of Training Shapes from $\RealShapes$}
\label{sec:abl_size}

All the fine-tuning methods make use of a training distribution of shapes that are sampled from $\RealShapes$.
For some domains, the size of available samples from $\RealShapes$ may be limited.
We run an experiment on 2D CSG to see how different fine-tuning methods are affected by training data size.
We present the results of this experiment in Figure \ref{fig:plot_beam_size}, middle.
We plot the number of training shapes on the X-axis and test set reconstruction accuracy on the Y-axis.
All fine-tuning methods improve as the training size of $\RealShapes$ increases, but once again, combining multiple PLAD methods leads to the best performance in all regimes.
This study also demonstrates the sample efficiency of PLAD combinations: LEST+ST and LEST+ST+WS trained on 1,000 shapes achieve better test set generalization than RL trained on 10,000 shapes.   
\subsection{Convergence Speed}
\label{sec:speed}

Beyond reconstruction accuracy, we are also interested in the convergence properties of a fine-tuning method.
Policy gradient RL is notoriously unstable and slow to converge, which is undesirable.
For 2D CSG, we record the convergence speed of each method and present these results in Figure \ref{fig:plot_beam_size}, right.
We plot reconstruction accuracy (Y-axis) as a function of training wall-clock time (X-axis); all timing information was collected on a machine with a GeForce RTX 2080 Ti GPU and an Intel i9-9900K CPU. 
All PLAD techniques converge faster than policy gradient RL. 
For instance, RL took 36 hours to reach its converged test-set CD of 1.097, while LEST matched this performance at 1.1 hours (32x faster) and LEST+ST matched this performance at 0.85 hours (42x faster). 

%% file: 05-conclusion.tex
\section{Conclusion}
\label{sec:conc}

We presented the PLAD framework to group a family of techniques for fine-tuning shape program inference models with Pseudo-Labels and Approximate Distributions.
Within this framework, we proposed LEST: a self-training variant that creates a shape distribution $\X$ approximating the real distribution $\RealShapes$ by executing inferred latent programs.
Experiments on 2D CSG, 3D CSG, and ShapeAssembly demonstrate that PLAD methods achieve better reconstruction accuracy and converge faster than policy gradient RL, the current standard approach for black-box fine-tuning.
Finally, we found that combining updates from
multiple PLAD methods outperforms any individual technique. 

While fine-tuning $p(\z | \x)$, PLAD methods construct $(\X, \Z)$ sets approximating the statistics of $\RealShapes$, specializing $p(\z | \x)$ towards $\RealShapes$.
As a consequence, $p(\z | \x)$ may not generalize as well to shapes outside of $\RealShapes$; we explore this phenomenon in the supplemental material.
Training a general-purpose inference model for all shapes expressible under the grammar is an interesting line of future work.

While our work focuses on reconstruction quality, producing programs with `good' structure matters just as much, if the program is to be used for editing tasks.
Currently, the synthetic pretraining data is the only place where knowledge about what constitutes ``good program structure'' can be injected.
Such knowledge must be expressed in procedural form, which may be harder to elicit from domain experts than declarative knowledge (i.e. ``a good program has these properties'' vs. ``this is how you write a good program'').
Finding efficient ways to elicit and inject such knowledge is an important future direction.

Finally, we believe that ideas from the PLAD framework are applicable to a broader class of program inference problems than those we evaluated in this paper. In principle, these approaches can be used to train an approximate inference model $p(\z | \x)$ for any domain in which (1) an executor $p(\x | \z)$ is available and (2) the executed output $\x$ takes the form of some concrete artifact which can be encoded via neural network (image, audio, text, etc.).
For example, one could imagine using PLAD techniques to infer graphics shader programs which produce certain textures or audio synthesizers which sound like certain real-world instruments. 

%% file: 06-acks.tex
\section*{Acknowledgments}
We would like to thank the anonymous reviewers for their helpful suggestions. This work was funded in parts by NSF award \#1941808 and a Brown University Presidential Fellowship. Daniel Ritchie is an advisor to Geopipe and owns equity in the company. Geopipe is a start-up that is developing 3D technology to build immersive virtual copies of the real world with applications in various fields, including games and architecture

%% file: supplemental.tex
\section{Details of Domain Grammars}

\paragraph{2D CSG}

We follow the grammar from CSGNet \cite{sharma2018csgnet}. 
This grammar contains 3 Boolean operations (intersect, union, subtract), 3 primitive types (square, circle, triangle), and parameters to initialize each primitive (L and R tuples). Please refer to the CSGNet paper for details.

\vspace{-2mm}
\begin{align*}
&S \rightarrow E;\\
&E \rightarrow EET \mid P(L, R);\\
&T \rightarrow intersect \mid union \mid subtract;  \\
&P \rightarrow square \mid circle  \mid triangle;  \\
&L \rightarrow \big[8:8:56\big]^2;~~ R \rightarrow \big[8:4:32\big].
\end{align*}

\paragraph{3D CSG}

We design our own grammar for 3D CSG similar in spirit to the grammar of CSGNet. 
While CSGNet does contain a 3D CSG grammar, we find that it overly discretizes the possible spacing and positioning of primitives.
Therefore in our grammar, we allow each primitive to be parameterized at the same granularity as the voxel grid (32 bins). In this way, each primitive takes in 6 parameters (instead of 2 parameter tuples), where the 6 parameters control the position and scaling of the primitive.

\vspace{-2mm}
\begin{align*}
&S \rightarrow E;\\
&E \rightarrow EET \mid P(F, F, F, F, F, F);\\
&T \rightarrow intersect \mid union \mid subtract;  \\
&P \rightarrow cuboid \mid ellipsoid;  \\
&F \rightarrow \big[1:32\big]
\end{align*}

\paragraph{ShapeAssembly}

ShapeAssembly is a domain-specific language for creating structures of 3D Shapes \cite{ShapeAssembly}. 
It creates structures by instantiating parts (\textit{Cuboid} command), and then attaching parts to one another (\textit{attach} command). 
It further includes macro operators that capture higher-order spatial patterns (\textit{squeeze}, \textit{reflect}, \textit{translate} commands). 
To remain consistent with our CSG experiments, we further modify the grammar such that all continuous parameters are discretized. 

\vspace{-2mm}
\begin{align*}
&S \xrightarrow{}  BBoxBlock; ShapeBlock; \\
&BBoxBlock \xrightarrow{} \text{bbox} = \texttt{Cuboid}(1.0, x, 1.0)  \\
&ShapeBlock \xrightarrow{} PBlock ; ShapeBlock \mid None  \\
&PBlock \xrightarrow{}  c_n = \texttt{Cuboid}(x, x, x); ABlock; SBlock \\
&ABlock \xrightarrow{} Attach \mid Attach ; Attach \mid Squeeze \\
&SBlock \xrightarrow{} Reflect \mid Translate \mid None \\
&Attach \xrightarrow{} \texttt{attach}(cube_{n}, f, uv, uv) \\
&Squeeze \xrightarrow{} \texttt{squeeze}(cube_{n}, cube_{n}, face, uv) \\
&Reflect \xrightarrow{} \texttt{reflect}(\text{axis}) \\
&Translate \xrightarrow{} \texttt{translate}(\text{axis}, m, x) \\
&f \xrightarrow{} right\: \mid\: left \:\mid\: top\: \mid\: bot\: \mid\: front\: \mid\: back \\
&\text{axis} \xrightarrow{} X\: \mid\: Y \:\mid\: Z\: \\
& x \in [1, 32] / 32. \\
& uv \in [1, 10]^2 / 10. \\
& n \in [0, 10] \\
& m \in [1, 4] \\
\end{align*}

\section{Details of Synthetic Pretraining}

\paragraph{2DCSG} We follow the synthetic pretraining steps from CSGNet and directly use their released pretrained model weights. Please refer to their paper and code for further details.

\paragraph{3DCSG} 

We generate synthetic programs for 3D CSG with the following procedure. 
First, we sample K primitives, where K is randomly chosen between 2 and 12. 
To sample a primitive, we sample a center position within the voxel space, and then we sample a scale, such that the scale is constrained so that the primitive will not extend past the borders of the voxel grid.
We then find if the bounding boxes of any two primitives overlap in space (using the position and scale of each primitive).
We then construct a binary tree of Boolean operations by randomly merging the K primitives together, until only one group remains. 
Each Boolean operation merges two primitive groups into a single primitive group.
The type of semantically valid Boolean operation depends on the overlaps between primitives of the two groups. 
When a group of primitives A and a group of primitives B is merging:
union is always a valid operation, difference is a valid operation if each primitive in group B shares an overlap with some primitive in group A, and intersection is a valid operation if each primitive in group A shares an overlap with some primitive in group B and each primitive in group B shares an overlap with some primitive in group A.
We can then unroll this binary tree of boolean operations into a sequence of tokens from the CSG grammar, forming a synthetic program.
We sample 2,000,000 synthetic programs according to this procedure, that are used during supervised pretraining, and we sample another 1000 synthetic programs that we use a validation set. 
We pretrain our model for 40 epochs, where each epoch takes around 1.5 hours to complete.
At this check-point, the model had converged to a reconstruction IoU of ~90 on both train and validation synthetic data.

\paragraph{ShapeAssembly}

We generate synthetic programs for ShapeAssembly with the following procedure.
We first sample the number of primitive blocks K (PBlock), where K is randomly chosen between 2 and 8; note that the number of cuboids created can be greater then K, when symmetry operations are applied.
Each PBlock is filled in with random samples according to the grammar syntax.
First a cuboid is created, then an attach block is applied, then a symmetry block is applied. 
An attach block can contain either one attach operation, one squeeze operation, or two attach operations. 
A symmetry block can contain either a reflect operation, a translation operation, or no operation.
Command parameters are randomly sampled according to simple heuristics (e.g. reflections are more common than translations) and in order to maintain language semantics (e.g. attaches can only be made to previously instantiated cuboid indices). 
A final validation step occurs after a complete set of program tokens has been synthetically generated; we execute the synthetic program, and check how many voxels are uniquely occupied by each cuboid in the executed output.
If any cuboid uniquely occupies less than 8 voxels, the entire synthetic sample is rejected. 
We sample 2,000,000 synthetic programs according to this procedure, that are used during supervised pretraining, and we sample another 1000 synthetic programs that we use as a validation set.
We pretrain our model for 26 epochs, where each epoch takes around 40 minutes to complete.
At this check-point the model had converged to reconstruction IoU of ~70 on both train and validation synthetic data.

\section{Experiment Hyperparameters}

\paragraph{3D Experiments}

For 3D CSG and ShapeAssembly, we use the following model hyper-parameters. 

The encoder for both cases is a 3D CNN that consumes a 32 x 32 x 32 voxel grid. 
It has four layers of convolution, ReLU, max-pooling, and dropout. Each convolution layer uses kernel size of 4, stride of 1, padding of 2, with channels (32, 64, 128, 256). 
The output of the CNN is a (2x2x2x256) dimensional vector, which we transform into a (8 x 256) vector.
This vector is then sent through a 3-layer MLP with ReLU and dropout to produce a final (8 x 256) vector that acts as an 8-token embedding of the voxel grid. 

The decoder for both cases is a Transformer Decoder module \cite{10.5555/3295222.3295349}. It uses 8 layers and 16 heads, with a hidden dimension size of 256. It attends over the 8-token CNN voxel encoding and up to 100 additional sequence tokens, with an auto-regressive attention mask.  
We use a learned positional embedding for each sequence position.
An embedding layer lifts each token into an embedding space, consumed by the transformer, and a 2-layer MLP converts Transformer outputs into a probability distribution over tokens.

In all cases we set dropout to 0.1 . We use a learning rate of 0.0005 with the Adam optimizer \cite{Adam} for all training modes, except for RL, where following CSGNet we use SGD with a learning rate of 0.01 . 
During supervised pretraining we use a batch size of 400. During PLAD method fine-tuning we use batch size of 100. During RL fine-tuning we use a batch size of 4, due to memory limitations (a batch size of 4 takes up 10GB of GPU memory). 
Early stopping on the validation set is performed to determine when to end each round and when to stop introducing additional rounds. 
For deciding when to stop introducing additional rounds, we use a patience of 100 epochs.
For deciding when to stop each round, we use a patience of 10 epochs.
In both cases we employ a patience threshold of 0.001 IoU improvement (e.g. we must see at least this much improvement to reset the patience). 
Within each round of PLAD training, we check validation set reconstruction performance with a beam size of 3; between rounds of PLAD training we check validation set reconstruction performance with a beam size of 5; final reconstruction performance of converged models is computed with a beam size of 10.

For RL runs, we make a gradient update after every 10 batches, following CSGNet. For runs that involve VAE training (all Wake-Sleep runs), we add an additional module in-between the encoder and the decoder. 
This module uses an MLP to convert the output of the encoder into a 128 x 2 latent vector (representing 128 means and standard deviations).
This module then samples an 128 dimensional vector from a normal distribution described by the means and standard deviations, and further lifts this encoding into the dimension that the decoder expects with a sequence of linear layers. 
For each round of VAE training, we allow the VAE to update for no more than 100 epochs. We perform early-stopping for VAE training with respect to its loss, where the loss is a combination of reconstruction (cross-entropy on token predictions) and KL divergence, both weighed equally.

\paragraph{2D Experiments}

For 2DCSG, we follow the network architecture and hyper-parameters of CSGNet. 
All training regimes use a dropout of 0.2 and a batch size of 100. 
PLAD methods use the Adam optimizer with a learning rate of 0.001.
For deciding when to stop introducing additional rounds, we use a patience of 1000 epochs.
For deciding when to stop each round, we use a patience of 10 epochs.
In both cases we employ a patience threshold of 0.005 CD improvement. 
The parameters for the RL runs and VAE training are the same as in the 3D Experiments.

\section{P Best Update mode}

\begin{table}[]
    \centering   
    \small
    \begin{tabular}{@{}l cccc@{}}
        \toprule
        \textbf{$\Programs^{BEST}$} mode & ST & LEST & LEST+ST & LEST+ST+WS   \\
        Per round & 0.881 & 1.011 & 0.853 & 0.845 \\
        All-time  & \textbf{0.841} & \textbf{0.976} & \textbf{0.829} & \textbf{0.811 }\\
        \bottomrule
    \end{tabular}
    \caption{Different ways to update $\Programs^{BEST}$ data structure. In the "Per round" row, the data structure is cleared in between rounds. In the "All-time" row, the data structure maintains the best program for each input shape across multiple rounds. }
    \label{tab:pbest_upd}
\end{table}

During updates to $\Programs^{BEST}$, we choose to update each entry in $\Programs^{BEST}$ according to which inferred program has achieved the best reconstruction similarity with respect to the input shape. 
The entries of this data structure are maintained across rounds. 
There is another framing where the entries of this data structure are reset each round, so that the best program for each shape is reset each epoch. 
This is similar to traditional self-training framing. 

We run experiments on 2D CSG with this variant of $\Programs^{BEST}$ update and present results in Table \ref{tab:pbest_upd}. When the best program is maintained across rounds (All-time, bottom row) each fine-tuning strategy reaches a better converged reconstruction accuracy compared with when the best program is reset after each round (Per round, top row).  

\section{Failure to generalize beyond $\RealShapes$} 

\begin{figure}[]
    \centering
    \setlength{\tabcolsep}{1pt}
    \begin{tabular}{ccc}
        \textbf{SP} &  \textbf{LEST+ST+WS} & \textbf{Target} \\
        
        \includegraphics[trim={2.5cm 1.0cm 2.5cm, 1.0cm}, clip,{width=.33\linewidth}]{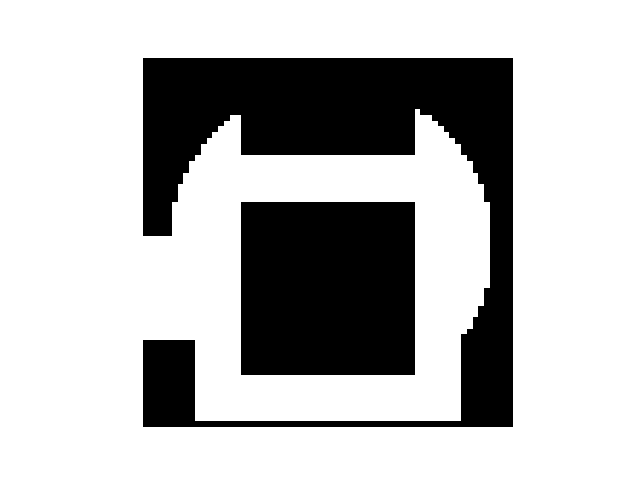} &
        \includegraphics[trim={2.5cm 1.0cm 2.5cm, 1.0cm}, clip,{width=.33\linewidth}]{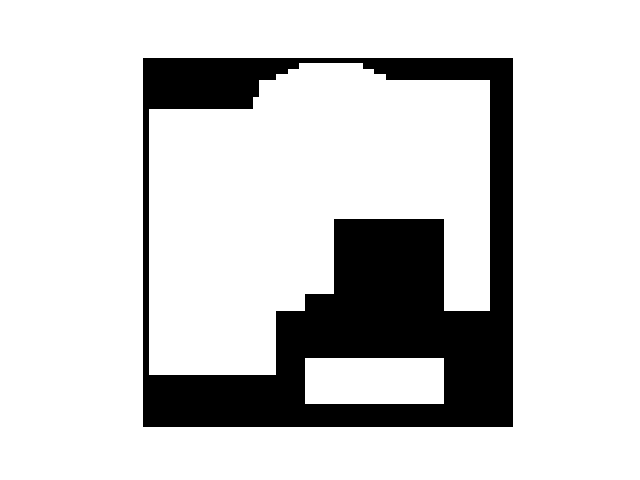} &
        \includegraphics[trim={2.5cm 1.0cm 2.5cm, 1.0cm}, clip,{width=.33\linewidth}]{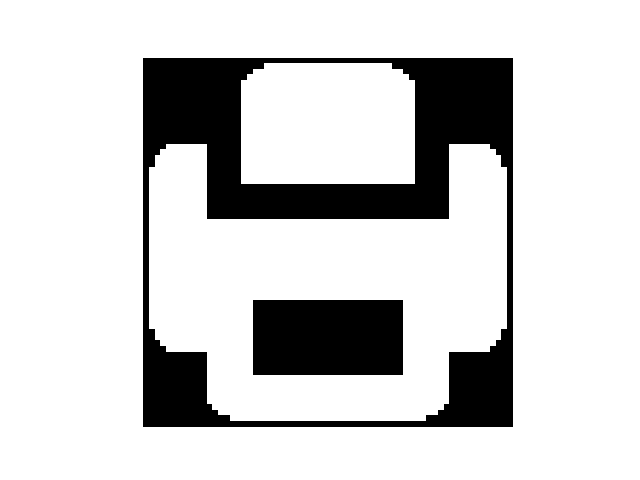} \\
        
        \includegraphics[trim={2.5cm 1.0cm 2.5cm, 1.0cm}, clip,{width=.33\linewidth}]{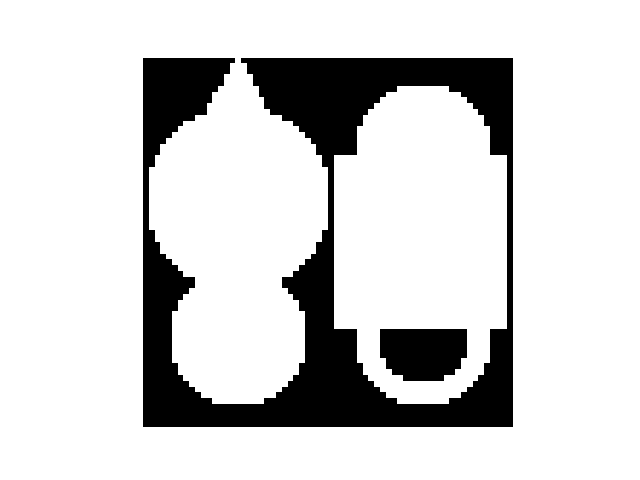} &
        \includegraphics[trim={2.5cm 1.0cm 2.5cm, 1.0cm}, clip,{width=.33\linewidth}]{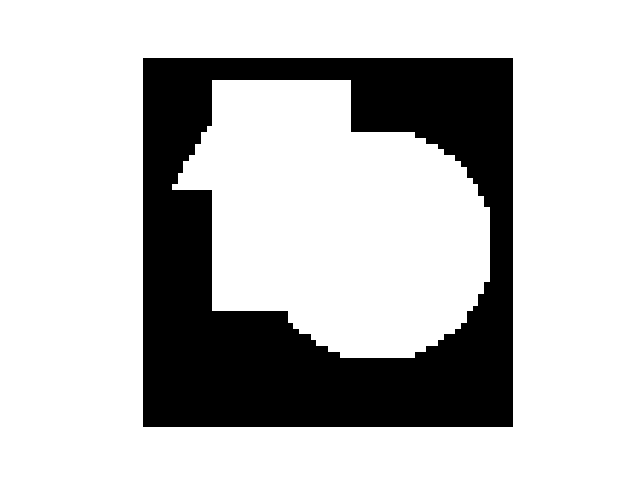} &
        \includegraphics[trim={2.5cm 1.0cm 2.5cm, 1.0cm}, clip,{width=.33\linewidth}]{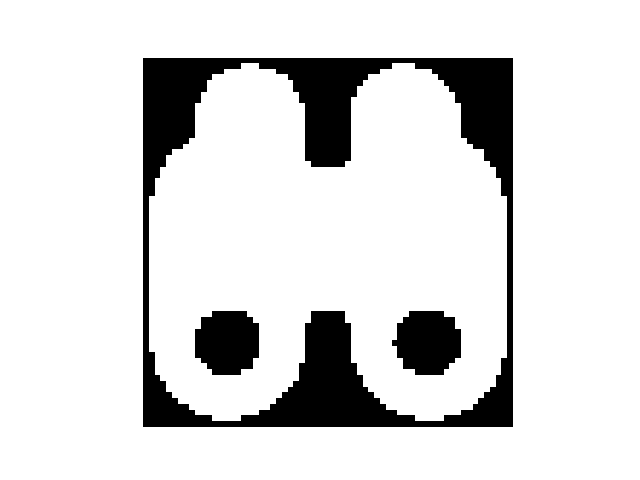} \\
        
        \includegraphics[trim={2.5cm 1.0cm 2.5cm, 1.0cm}, clip,{width=.33\linewidth}]{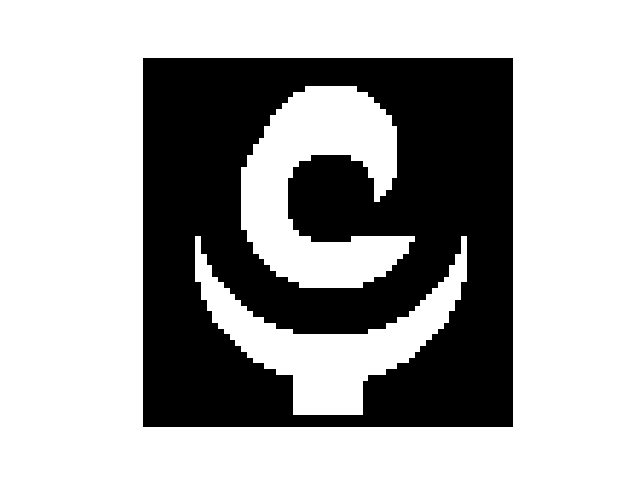} &
        \includegraphics[trim={2.5cm 1.0cm 2.5cm, 1.0cm}, clip,{width=.33\linewidth}]{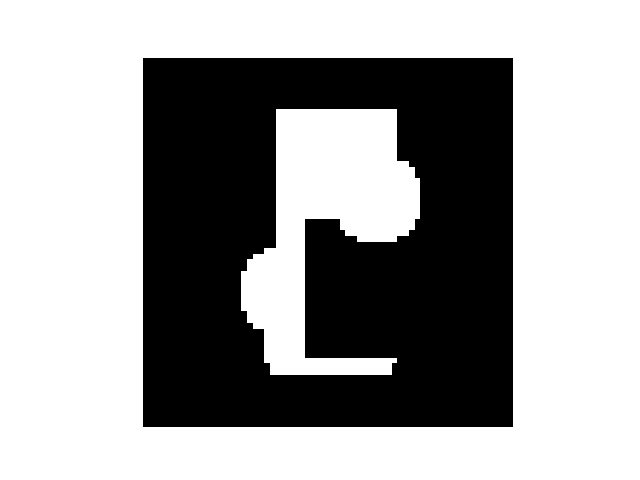} &
        \includegraphics[trim={2.5cm 1.0cm 2.5cm, 1.0cm}, clip,{width=.33\linewidth}]{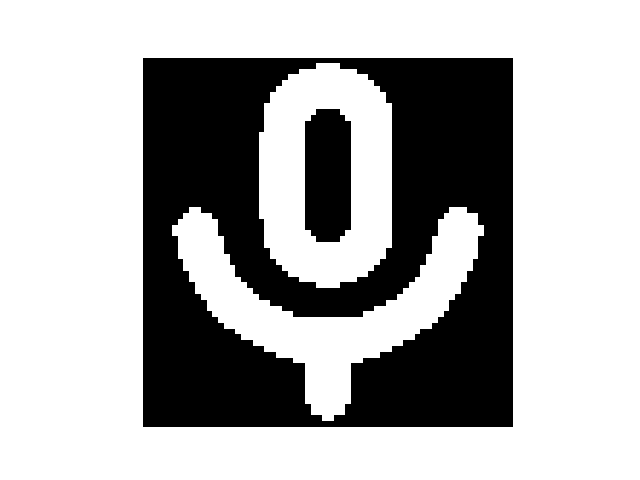} \\
        
        \includegraphics[trim={2.5cm 1.0cm 2.5cm, 1.0cm}, clip,{width=.33\linewidth}]{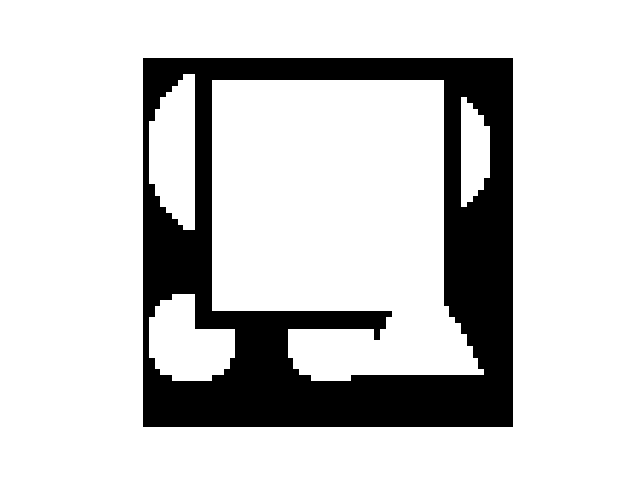} &
        \includegraphics[trim={2.5cm 1.0cm 2.5cm, 1.0cm}, clip,{width=.33\linewidth}]{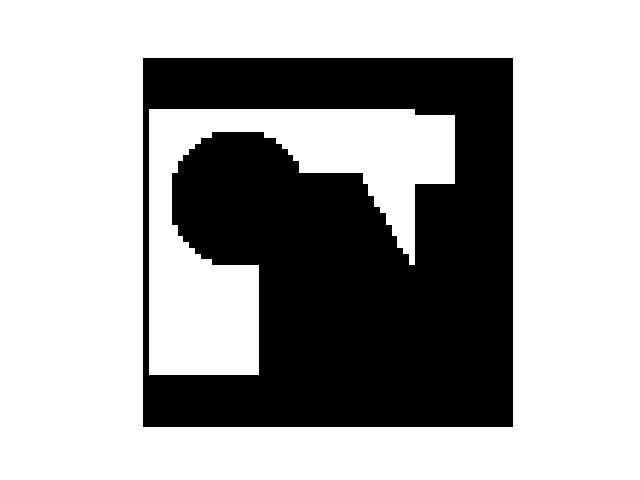} &
        \includegraphics[trim={2.5cm 1.0cm 2.5cm, 1.0cm}, clip,{width=.33\linewidth}]{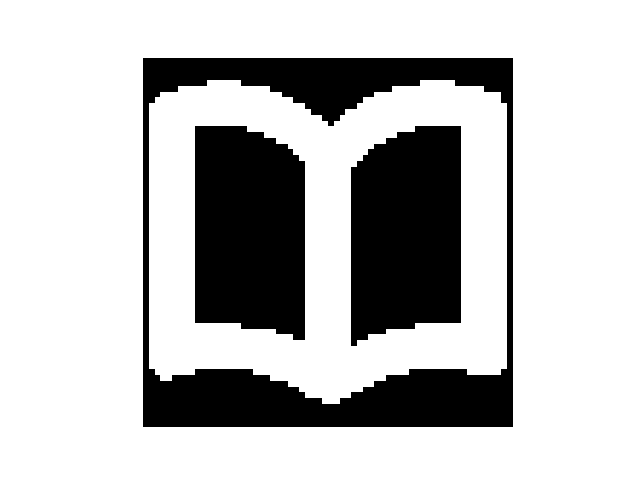} \\

    \end{tabular}
    \caption{Qualitative examples of inferring 2D CSG programs for 2D icons. Both SP and LEST+ST+WS fail to infer representative programs, but the reconstructions from LEST+ST+WS are even less accurate than those from SP.} 
    \label{fig:noun_qual}
\end{figure}

As demonstrated by our experiments, PLAD fine-tuning methods are able to successfully specialize $p(\z | \x)$ towards a distribution of interest $\RealShapes$. 
Unfortunately, this specialization comes at a cost; the fine-tuned $p(\z | \x)$ may actually generalize worse to out of distribution samples. 
To demonstrate this, we collected a small dataset of 2D icons from the The Noun Project\footnote{\url{https://thenounproject.com}}.
We tested the shape program inference abilities of the initial $p(\z | \x)$ trained under supervised pretraining (SP) and of the fine-tuned $p(\z | \x)$ trained under PLAD regimes (LEST+ST+WS) and specialized to CAD shapes.
We show qualitative examples of this experiment in Figure ~\ref{fig:noun_qual}.
While both methods fail to accurately represent the 2D icons, fine-tuning $p(\z | \x)$ on CAD shapes lowers the reconstruction accuracy significantly; the SP variant achieves an average CD of 1.9 while the LEST+ST+WS variant achieves a CD of 4.1
Developing $p(\z | \x)$ models capable of out-of-domain generalization is an important area of future research. 

\section{Potential Societal Impacts}

Fine-tuning our deep neural networks $p(\z | \x)$ requires a relatively large amount of electricity, which can have a significant environmental impact \cite{strubell2019energy}. Reducing the energy consumption of deep learning is an active research area \cite{you2019drawing, spring2017scalable}. Notably, PLAD techniques place no restrictions on the inference model, making it easy to adopt more efficient deep learning techniques. 
Moreover, shape program inference procedures may also allow the reverse engineering of protected intellectual property.
Thus, improvements in shape program inference may impact the content and enforcement of copyright law.

\section{Additional Qualitative Results}

\begin{figure*}[t!]
    \centering
    \footnotesize
    \setlength{\tabcolsep}{1pt}
    \begin{tabular}{cccccccc}
        \textbf{SP} &  \textbf{WS} &  \textbf{RL} &  \textbf{ST} &  \textbf{LEST} & \textbf{LEST+ST} & \textbf{LEST+ST+WS} & \textbf{Target} \\
        
        \includegraphics[trim={2.5cm 1.0cm 2.5cm, 1.0cm}, clip,{width=.115\linewidth}]{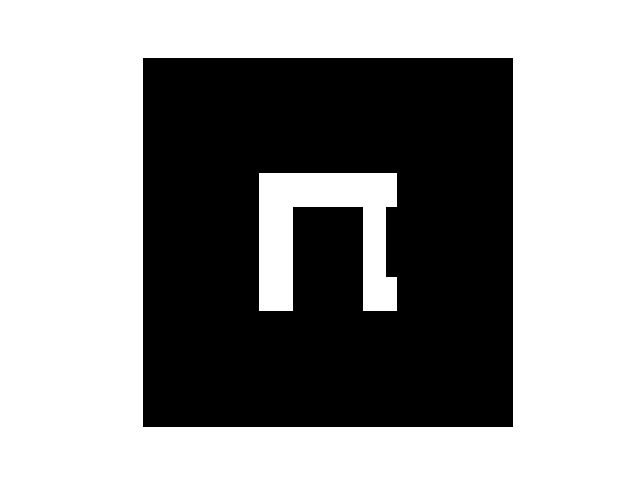} &
        \includegraphics[trim={2.5cm 1.0cm 2.5cm, 1.0cm}, clip,{width=.115\linewidth}]{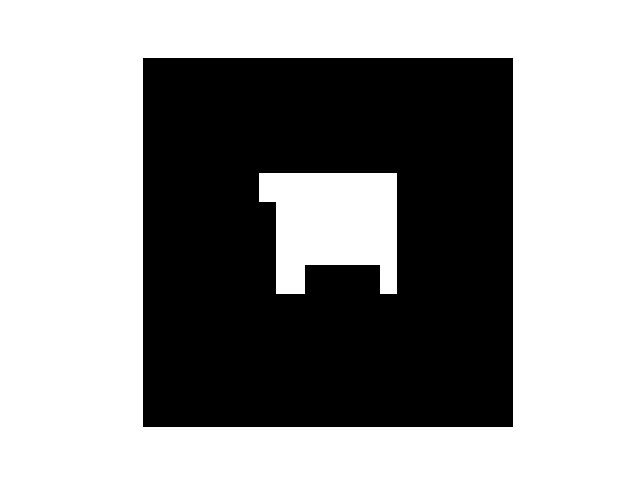} &
        \includegraphics[trim={2.5cm 1.0cm 2.5cm, 1.0cm}, clip,{width=.115\linewidth}]{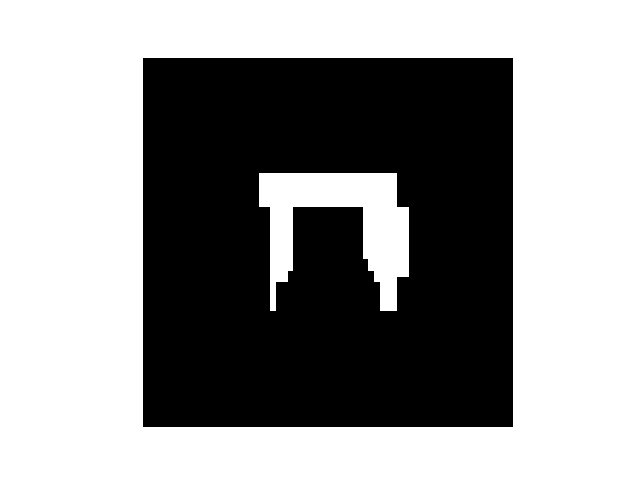} &
        \includegraphics[trim={2.5cm 1.0cm 2.5cm, 1.0cm}, clip,{width=.115\linewidth}]{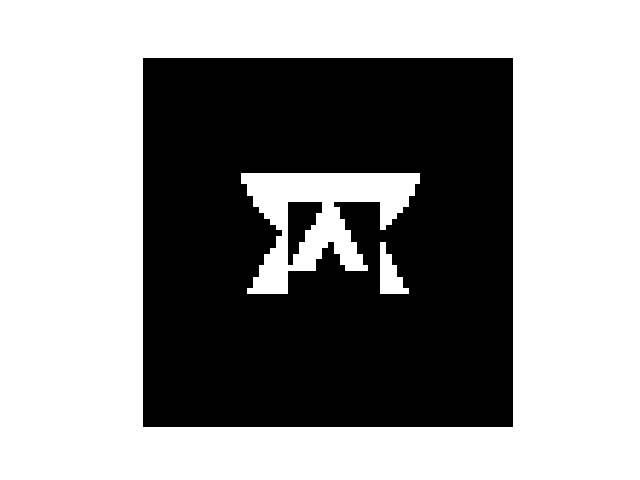} &
        \includegraphics[trim={2.5cm 1.0cm 2.5cm, 1.0cm}, clip,{width=.115\linewidth}]{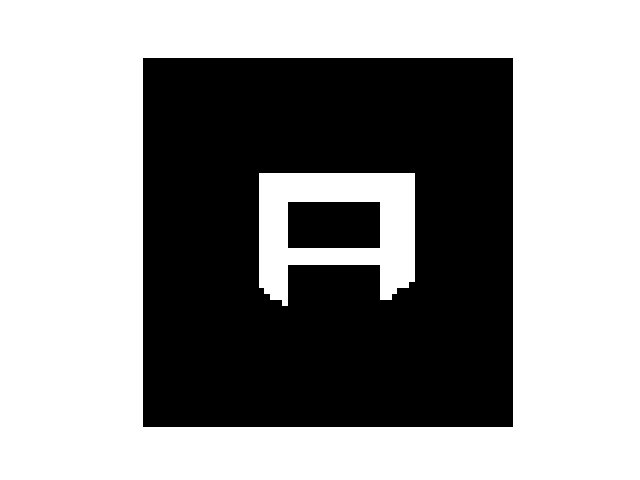} &
        \includegraphics[trim={2.5cm 1.0cm 2.5cm, 1.0cm}, clip,{width=.115\linewidth}]{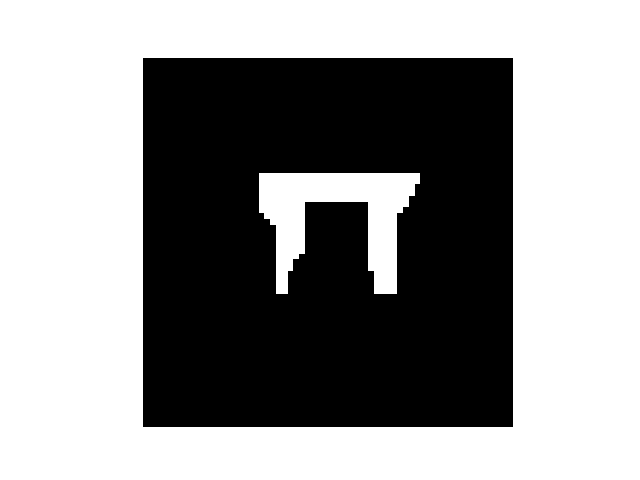} &
        \includegraphics[trim={2.5cm 1.0cm 2.5cm, 1.0cm}, clip,{width=.115\linewidth}]{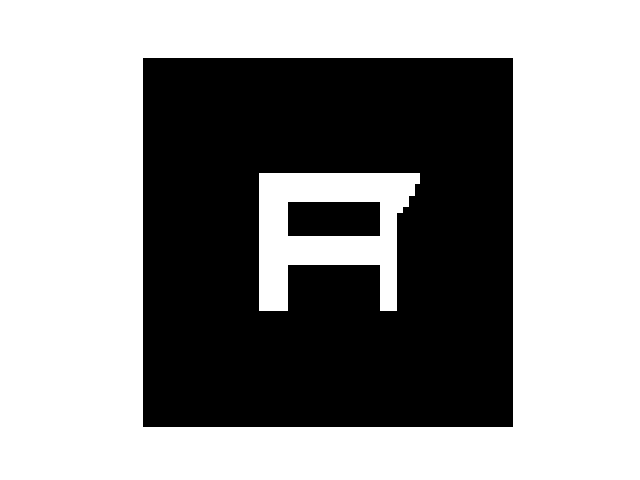} &
        \includegraphics[trim={2.5cm 1.0cm 2.5cm, 1.0cm}, clip,{width=.115\linewidth}]{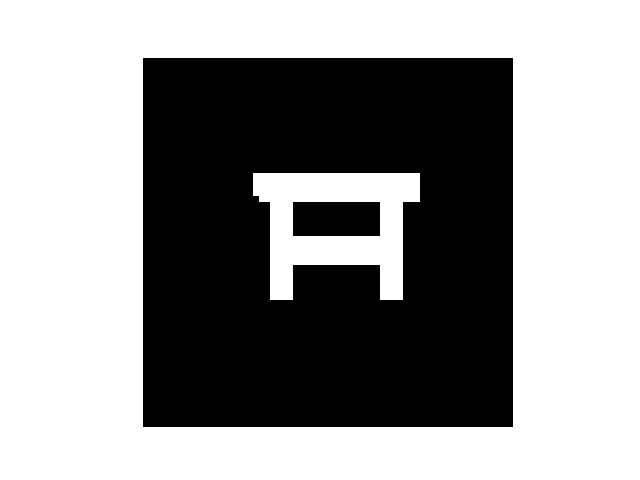} \\
        
        \includegraphics[trim={2.5cm 1.0cm 2.5cm, 1.0cm}, clip,{width=.115\linewidth}]{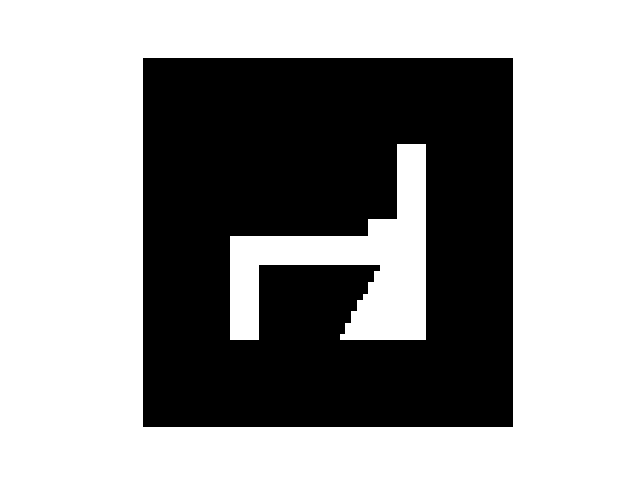} &
        \includegraphics[trim={2.5cm 1.0cm 2.5cm, 1.0cm}, clip,{width=.115\linewidth}]{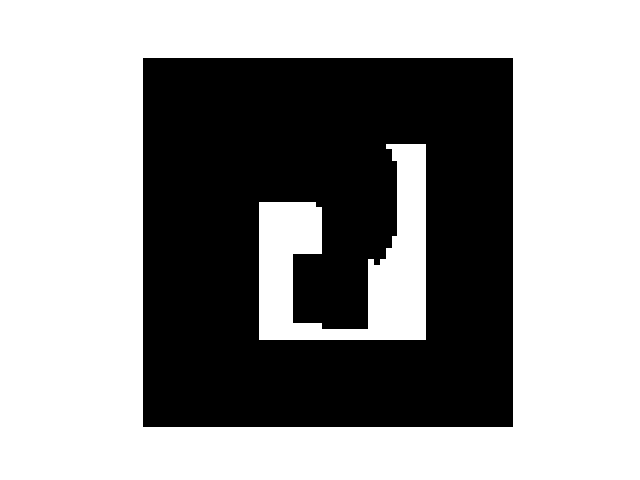} &
        \includegraphics[trim={2.5cm 1.0cm 2.5cm, 1.0cm}, clip,{width=.115\linewidth}]{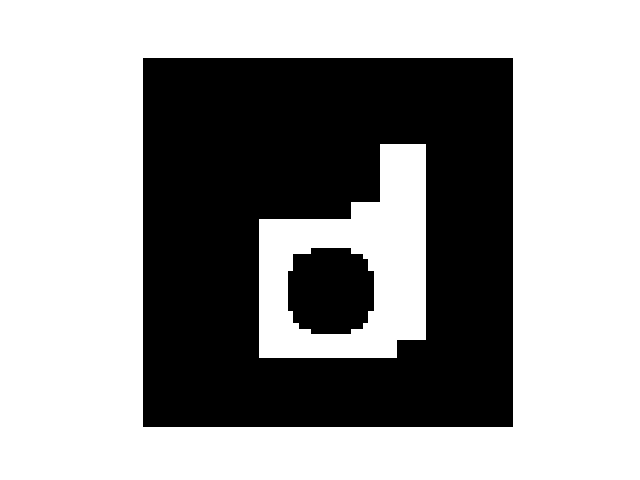} &
        \includegraphics[trim={2.5cm 1.0cm 2.5cm, 1.0cm}, clip,{width=.115\linewidth}]{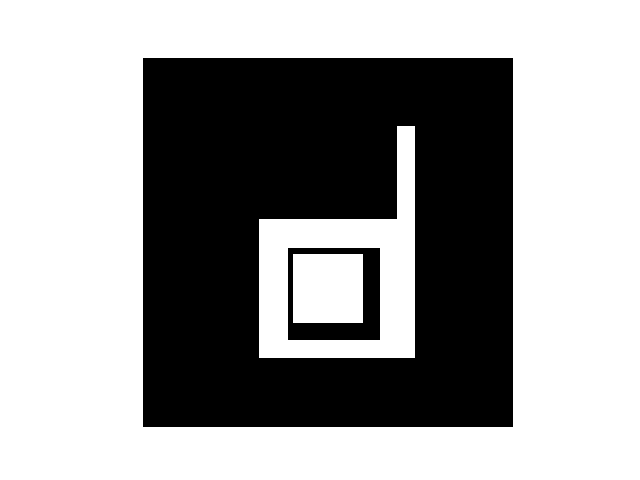} &
        \includegraphics[trim={2.5cm 1.0cm 2.5cm, 1.0cm}, clip,{width=.115\linewidth}]{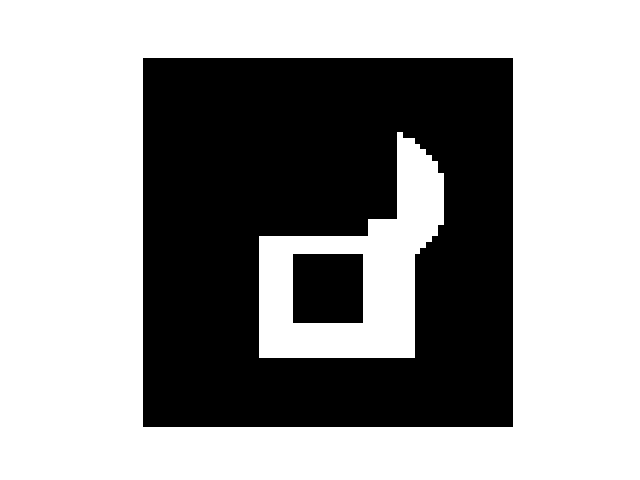} &
        \includegraphics[trim={2.5cm 1.0cm 2.5cm, 1.0cm}, clip,{width=.115\linewidth}]{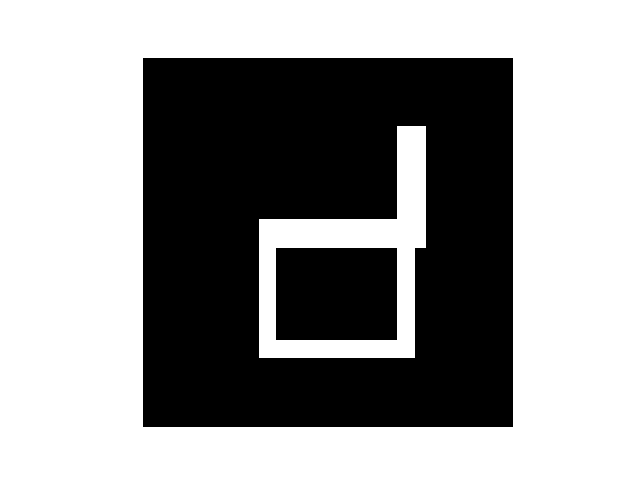} &
        \includegraphics[trim={2.5cm 1.0cm 2.5cm, 1.0cm}, clip,{width=.115\linewidth}]{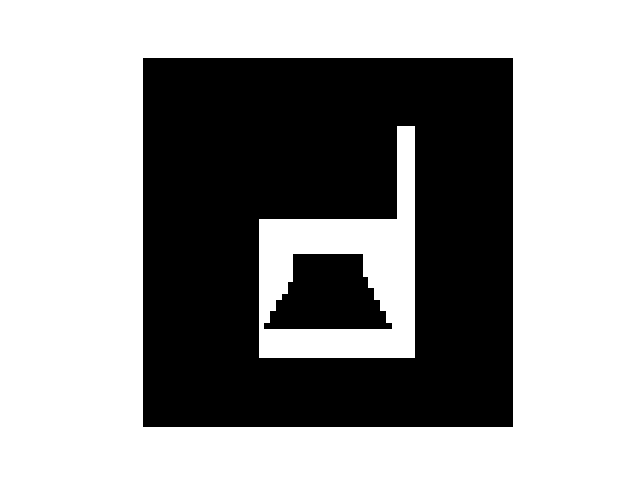} &
        \includegraphics[trim={2.5cm 1.0cm 2.5cm, 1.0cm}, clip,{width=.115\linewidth}]{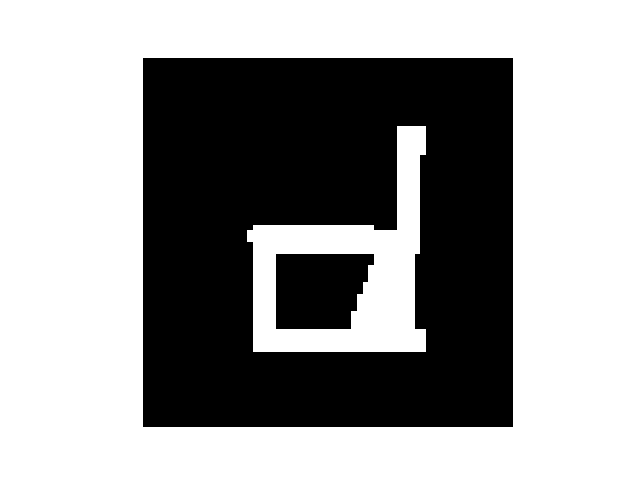} \\
        
        \includegraphics[trim={2.5cm 1.0cm 2.5cm, 1.0cm}, clip,{width=.115\linewidth}]{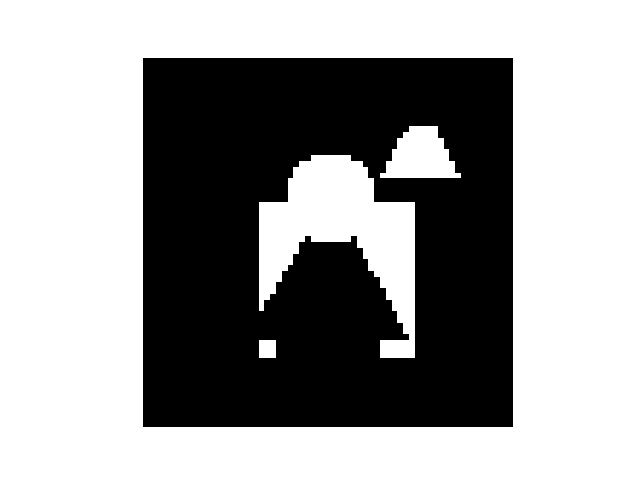} &
        \includegraphics[trim={2.5cm 1.0cm 2.5cm, 1.0cm}, clip,{width=.115\linewidth}]{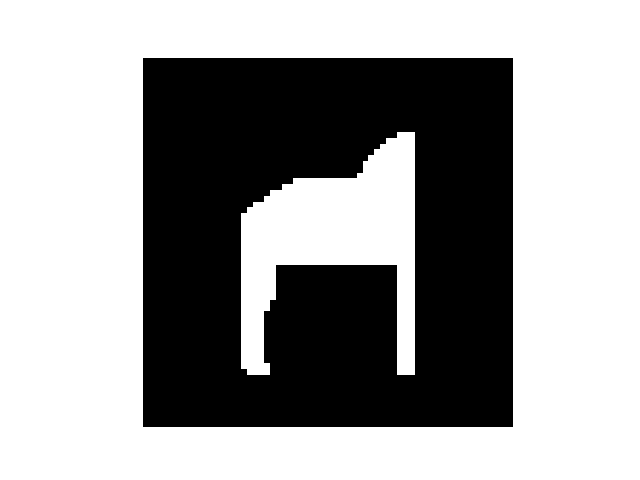} &
        \includegraphics[trim={2.5cm 1.0cm 2.5cm, 1.0cm}, clip,{width=.115\linewidth}]{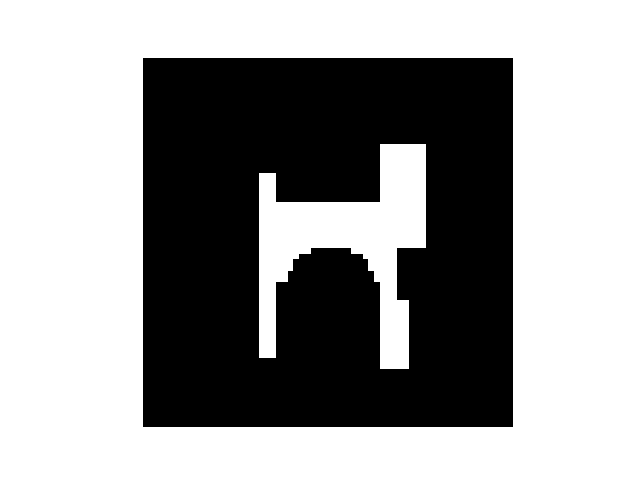} &
        \includegraphics[trim={2.5cm 1.0cm 2.5cm, 1.0cm}, clip,{width=.115\linewidth}]{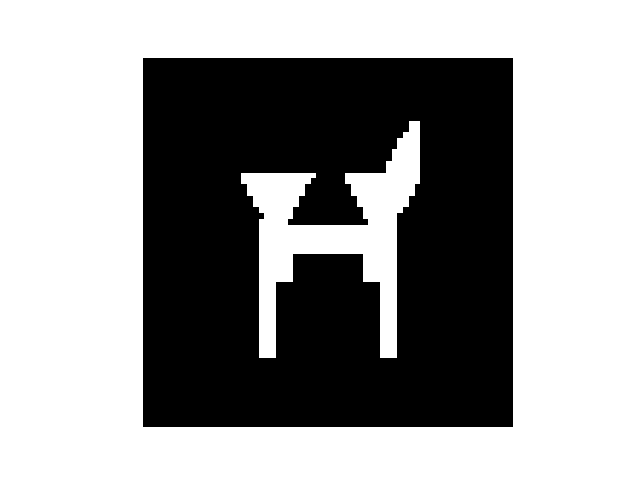} &
        \includegraphics[trim={2.5cm 1.0cm 2.5cm, 1.0cm}, clip,{width=.115\linewidth}]{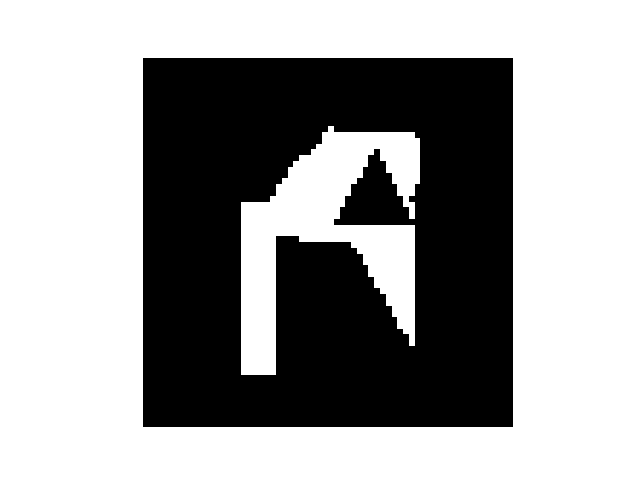} &
        \includegraphics[trim={2.5cm 1.0cm 2.5cm, 1.0cm}, clip,{width=.115\linewidth}]{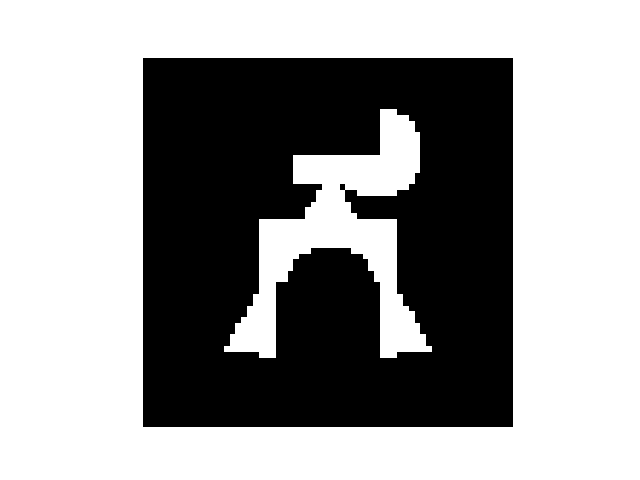} &
        \includegraphics[trim={2.5cm 1.0cm 2.5cm, 1.0cm}, clip,{width=.115\linewidth}]{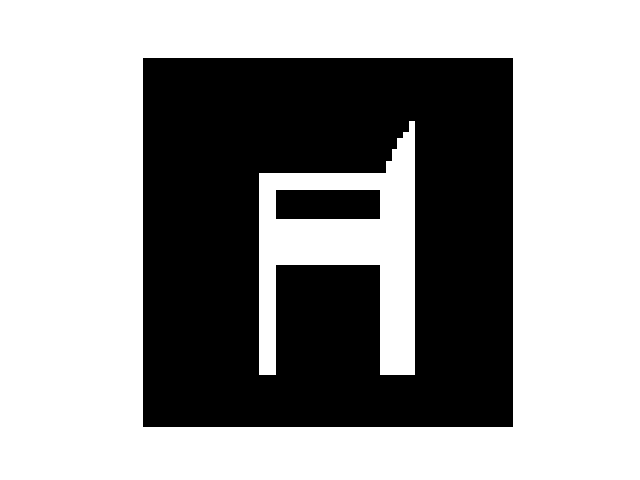} &
        \includegraphics[trim={2.5cm 1.0cm 2.5cm, 1.0cm}, clip,{width=.115\linewidth}]{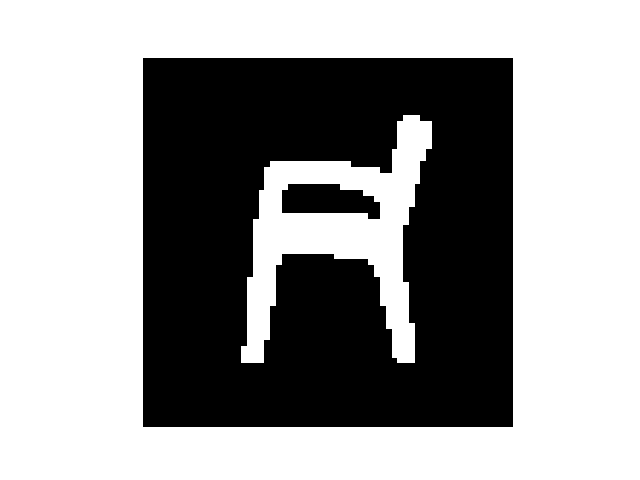} \\
        
        \includegraphics[trim={2.5cm 1.0cm 2.5cm, 1.0cm}, clip,{width=.115\linewidth}]{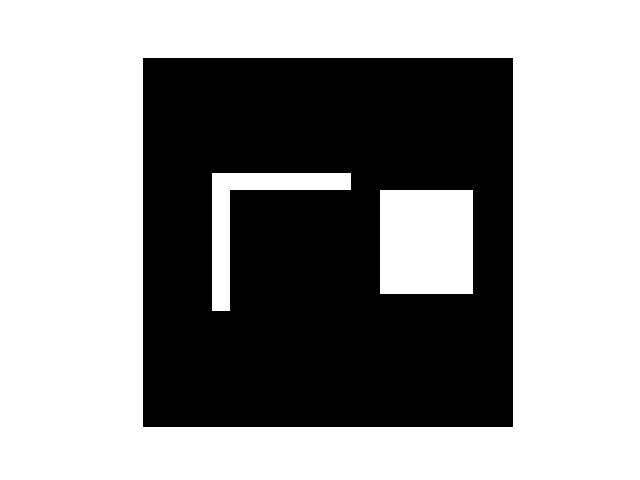} &
        \includegraphics[trim={2.5cm 1.0cm 2.5cm, 1.0cm}, clip,{width=.115\linewidth}]{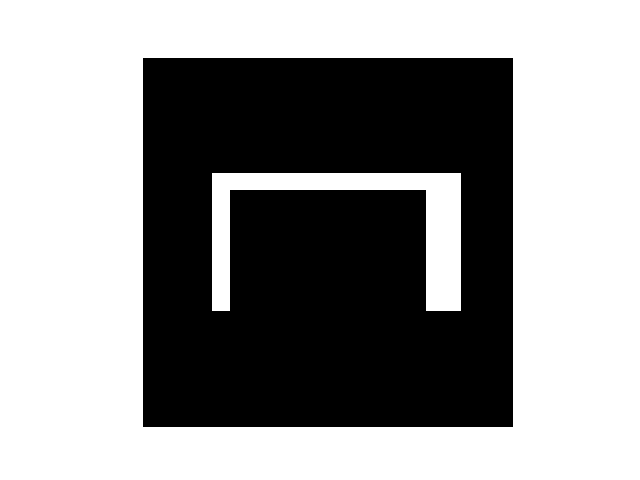} &
        \includegraphics[trim={2.5cm 1.0cm 2.5cm, 1.0cm}, clip,{width=.115\linewidth}]{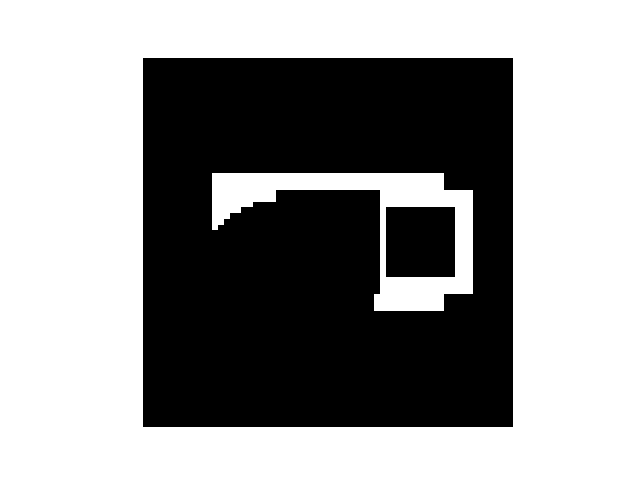} &
        \includegraphics[trim={2.5cm 1.0cm 2.5cm, 1.0cm}, clip,{width=.115\linewidth}]{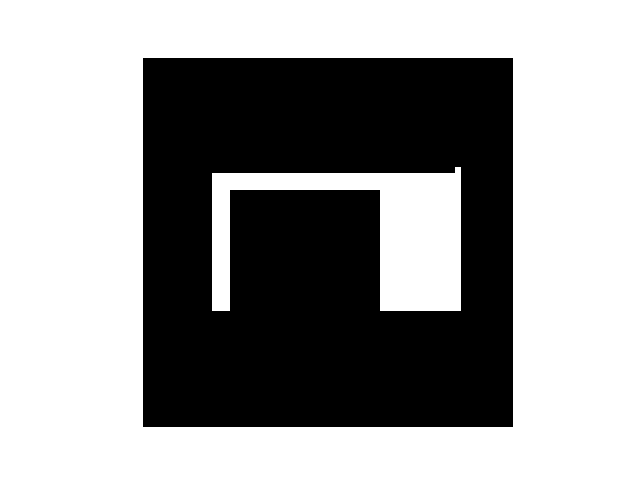} &
        \includegraphics[trim={2.5cm 1.0cm 2.5cm, 1.0cm}, clip,{width=.115\linewidth}]{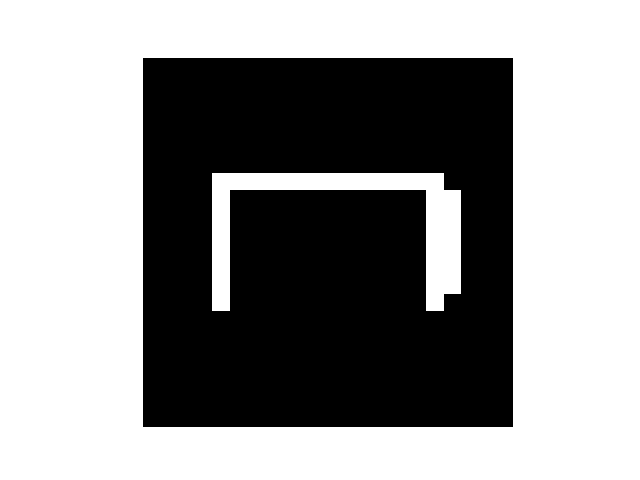} &
        \includegraphics[trim={2.5cm 1.0cm 2.5cm, 1.0cm}, clip,{width=.115\linewidth}]{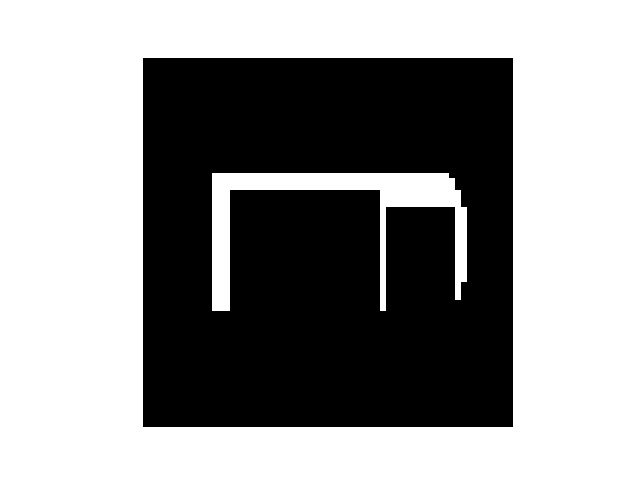} &
        \includegraphics[trim={2.5cm 1.0cm 2.5cm, 1.0cm}, clip,{width=.115\linewidth}]{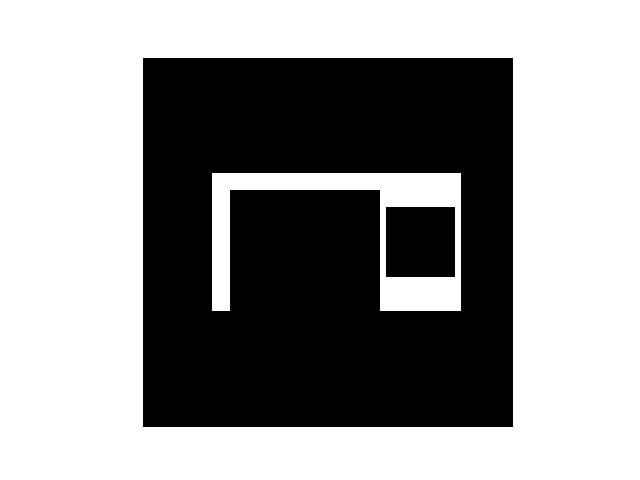} &
        \includegraphics[trim={2.5cm 1.0cm 2.5cm, 1.0cm}, clip,{width=.115\linewidth}]{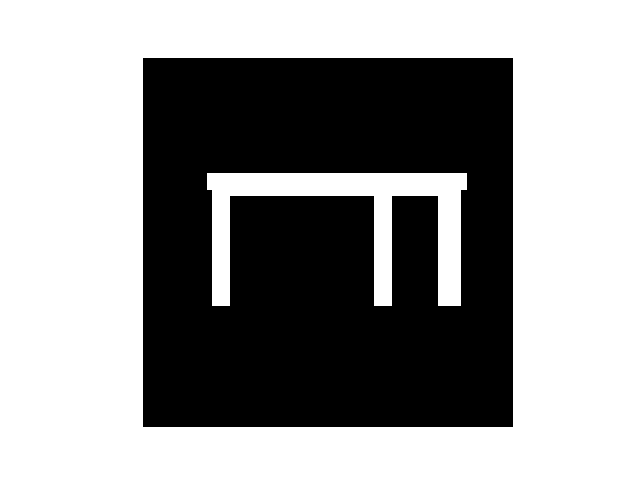} \\
        
        \includegraphics[trim={2.5cm 1.0cm 2.5cm, 1.0cm}, clip,{width=.115\linewidth}]{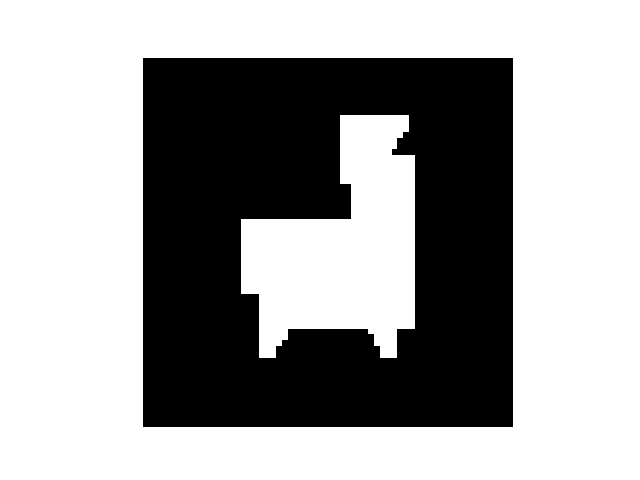} &
        \includegraphics[trim={2.5cm 1.0cm 2.5cm, 1.0cm}, clip,{width=.115\linewidth}]{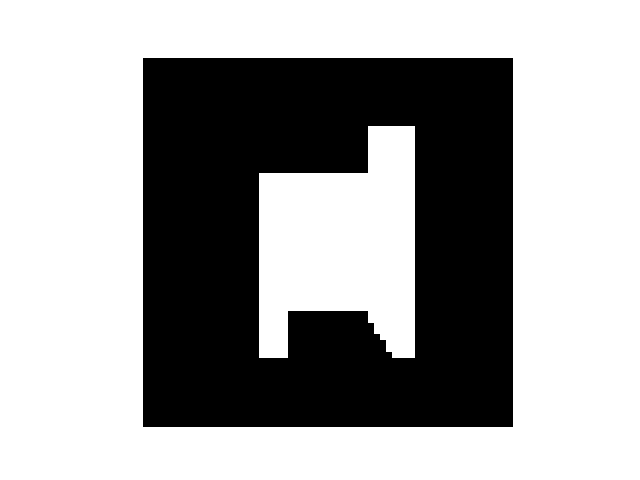} &
        \includegraphics[trim={2.5cm 1.0cm 2.5cm, 1.0cm}, clip,{width=.115\linewidth}]{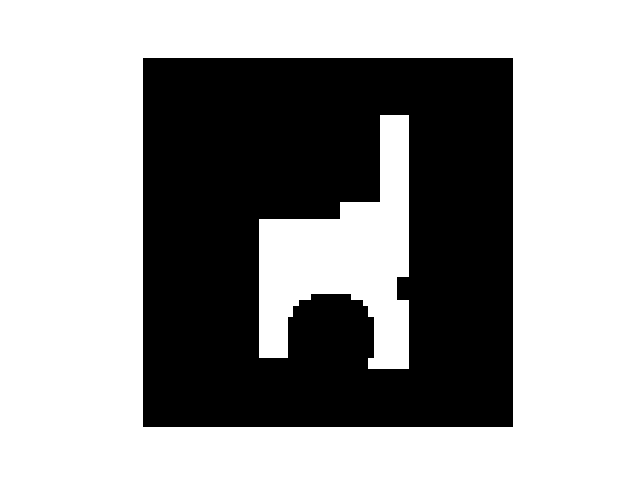} &
        \includegraphics[trim={2.5cm 1.0cm 2.5cm, 1.0cm}, clip,{width=.115\linewidth}]{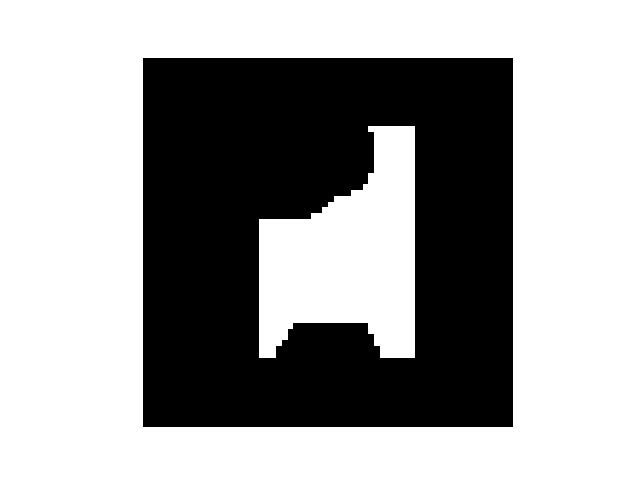} &
        \includegraphics[trim={2.5cm 1.0cm 2.5cm, 1.0cm}, clip,{width=.115\linewidth}]{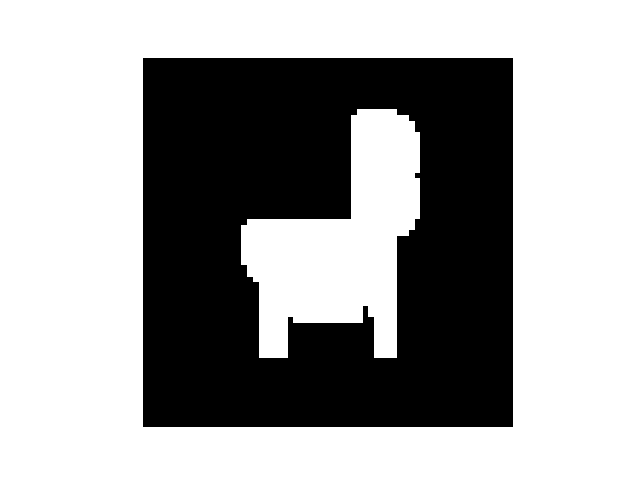} &
        \includegraphics[trim={2.5cm 1.0cm 2.5cm, 1.0cm}, clip,{width=.115\linewidth}]{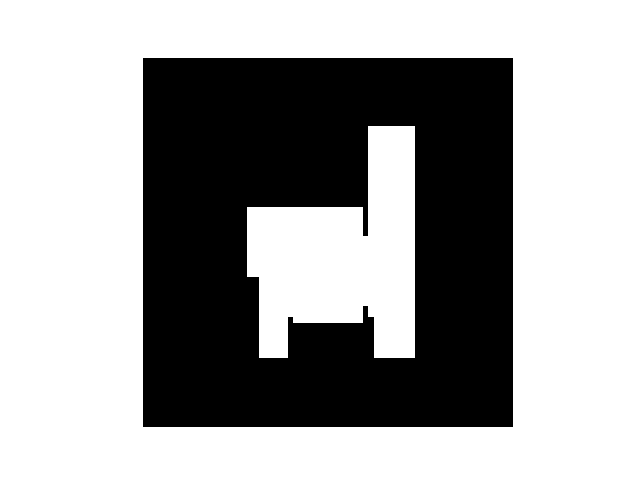} &
        \includegraphics[trim={2.5cm 1.0cm 2.5cm, 1.0cm}, clip,{width=.115\linewidth}]{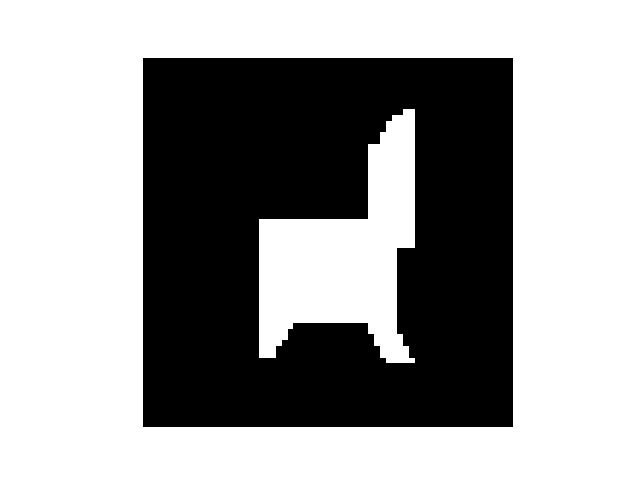} &
        \includegraphics[trim={2.5cm 1.0cm 2.5cm, 1.0cm}, clip,{width=.115\linewidth}]{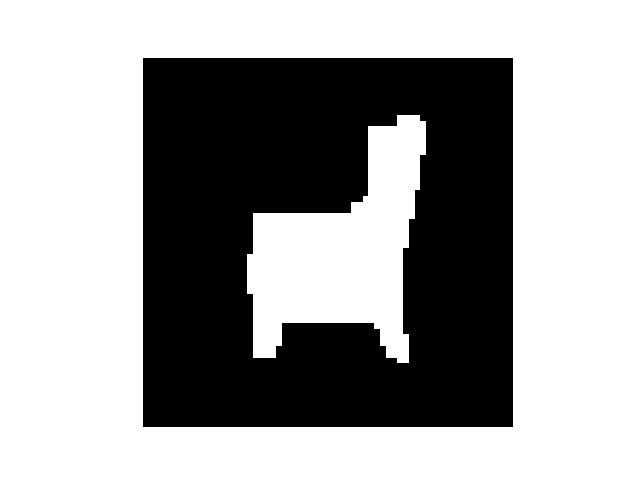} \\
        
        \includegraphics[trim={2.5cm 1.0cm 2.5cm, 1.0cm}, clip,{width=.115\linewidth}]{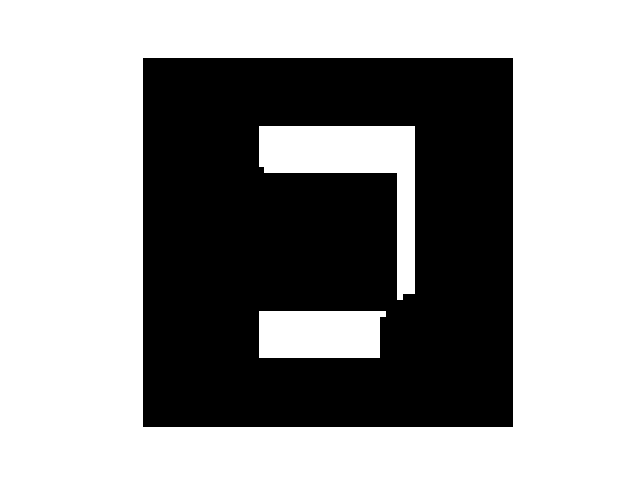} &
        \includegraphics[trim={2.5cm 1.0cm 2.5cm, 1.0cm}, clip,{width=.115\linewidth}]{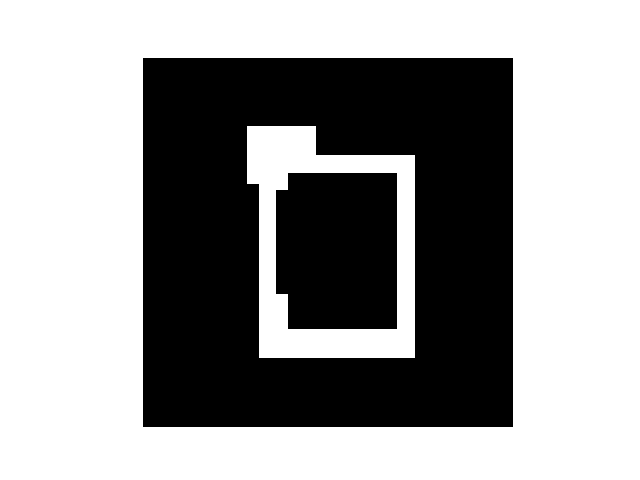} &
        \includegraphics[trim={2.5cm 1.0cm 2.5cm, 1.0cm}, clip,{width=.115\linewidth}]{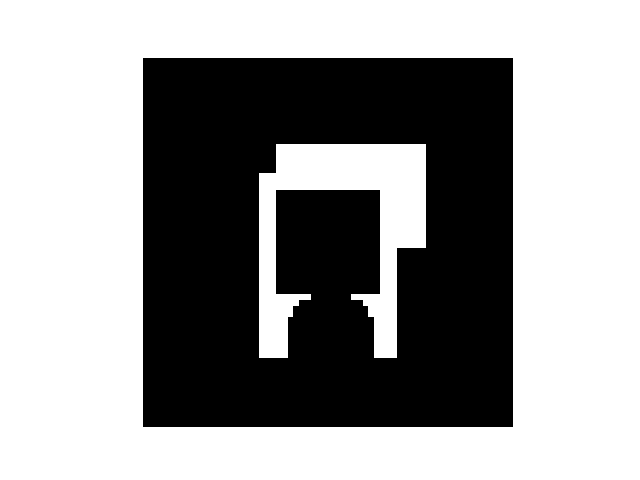} &
        \includegraphics[trim={2.5cm 1.0cm 2.5cm, 1.0cm}, clip,{width=.115\linewidth}]{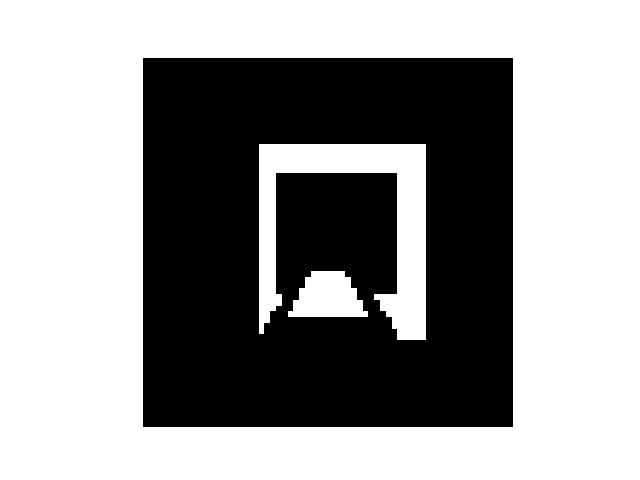} &
        \includegraphics[trim={2.5cm 1.0cm 2.5cm, 1.0cm}, clip,{width=.115\linewidth}]{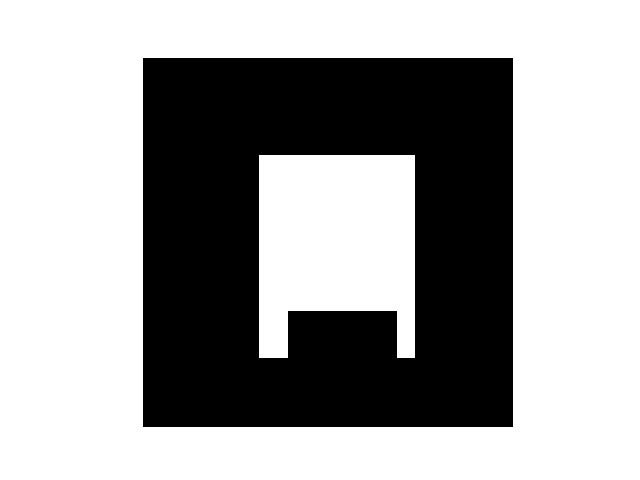} &
        \includegraphics[trim={2.5cm 1.0cm 2.5cm, 1.0cm}, clip,{width=.115\linewidth}]{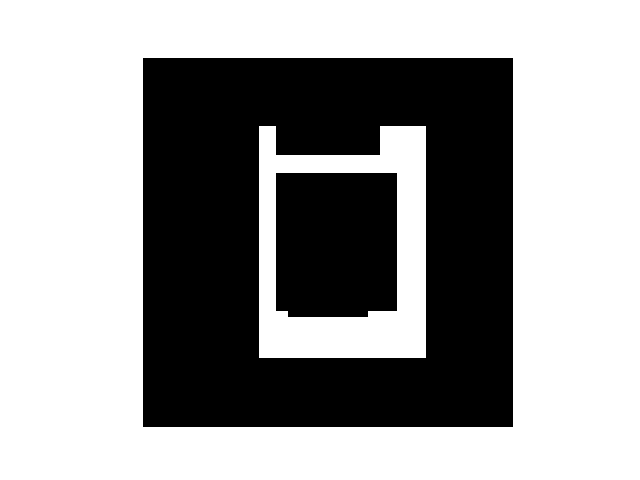} &
        \includegraphics[trim={2.5cm 1.0cm 2.5cm, 1.0cm}, clip,{width=.115\linewidth}]{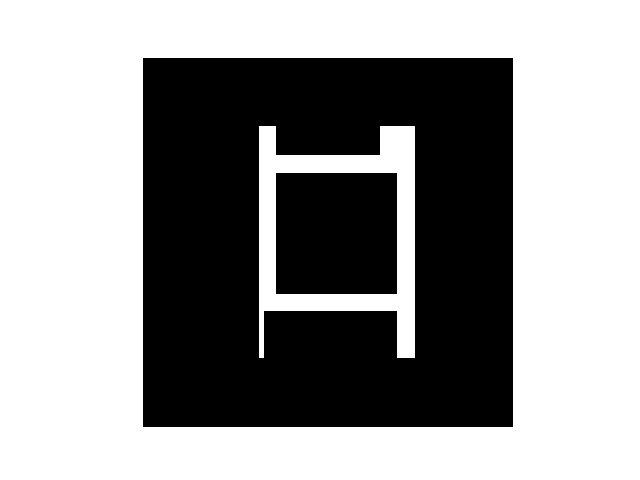} &
        \includegraphics[trim={2.5cm 1.0cm 2.5cm, 1.0cm}, clip,{width=.115\linewidth}]{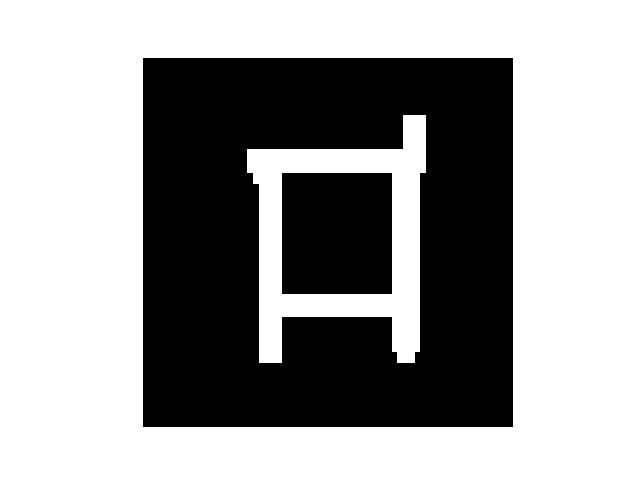} \\
        
        \includegraphics[trim={2.5cm 1.0cm 2.5cm, 1.0cm}, clip,{width=.115\linewidth}]{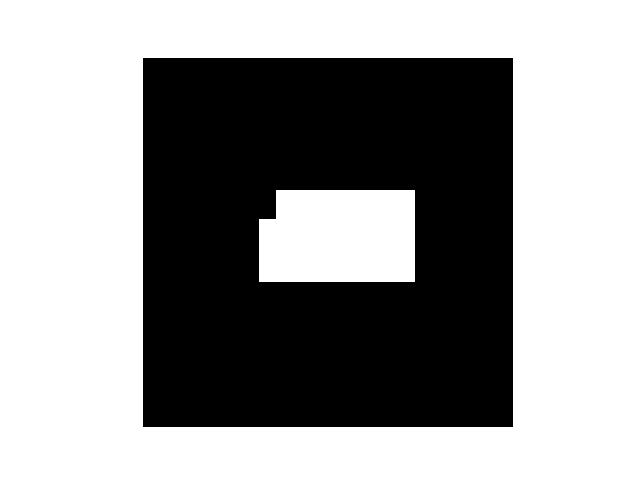} &
        \includegraphics[trim={2.5cm 1.0cm 2.5cm, 1.0cm}, clip,{width=.115\linewidth}]{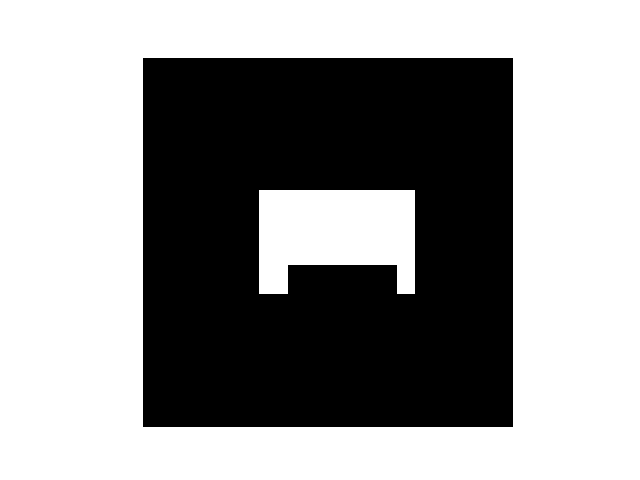} &
        \includegraphics[trim={2.5cm 1.0cm 2.5cm, 1.0cm}, clip,{width=.115\linewidth}]{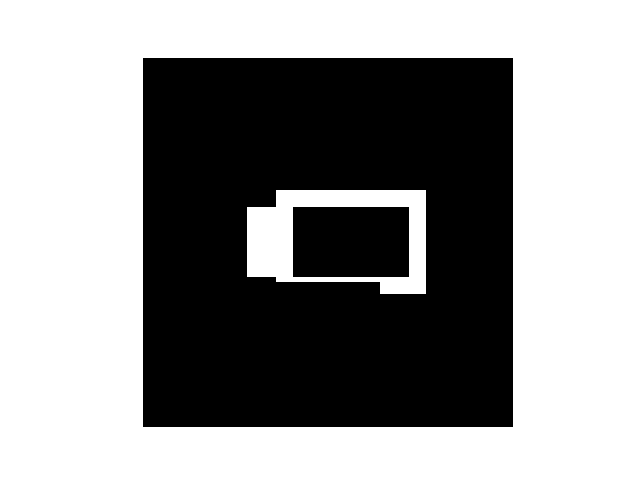} &
        \includegraphics[trim={2.5cm 1.0cm 2.5cm, 1.0cm}, clip,{width=.115\linewidth}]{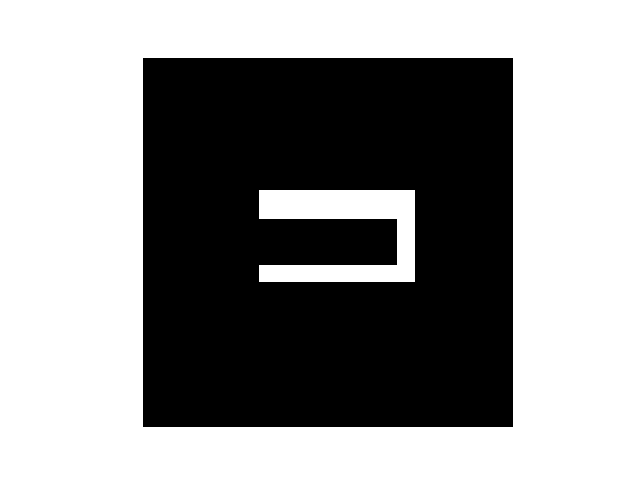} &
        \includegraphics[trim={2.5cm 1.0cm 2.5cm, 1.0cm}, clip,{width=.115\linewidth}]{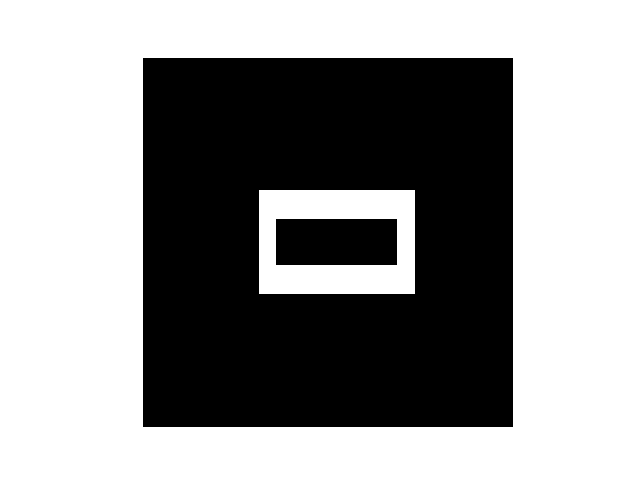} &
        \includegraphics[trim={2.5cm 1.0cm 2.5cm, 1.0cm}, clip,{width=.115\linewidth}]{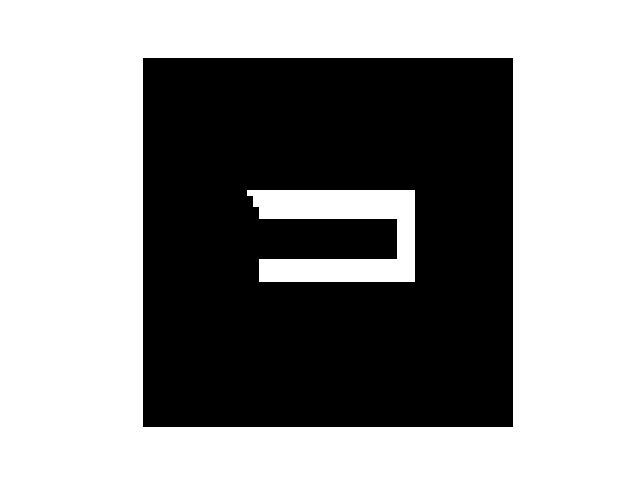} &
        \includegraphics[trim={2.5cm 1.0cm 2.5cm, 1.0cm}, clip,{width=.115\linewidth}]{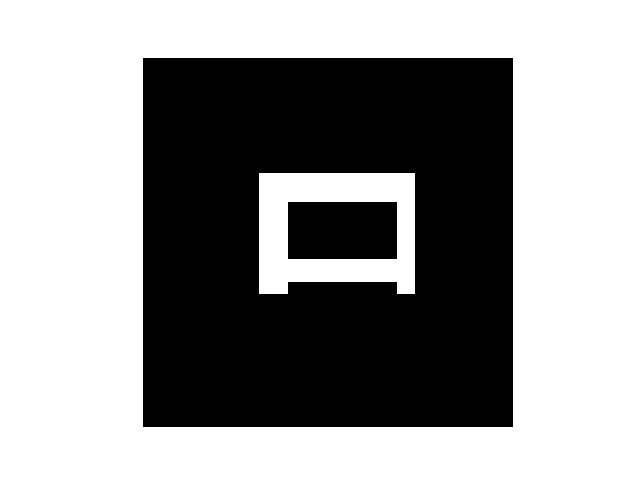} &
        \includegraphics[trim={2.5cm 1.0cm 2.5cm, 1.0cm}, clip,{width=.115\linewidth}]{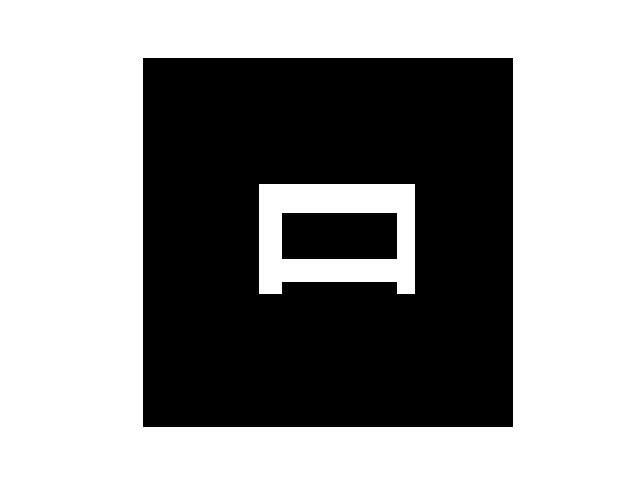} \\
        
        \includegraphics[trim={2.5cm 1.0cm 2.5cm, 1.0cm}, clip,{width=.115\linewidth}]{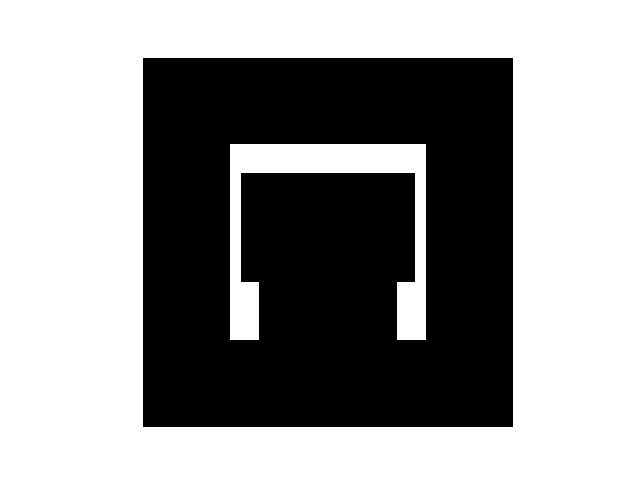} &
        \includegraphics[trim={2.5cm 1.0cm 2.5cm, 1.0cm}, clip,{width=.115\linewidth}]{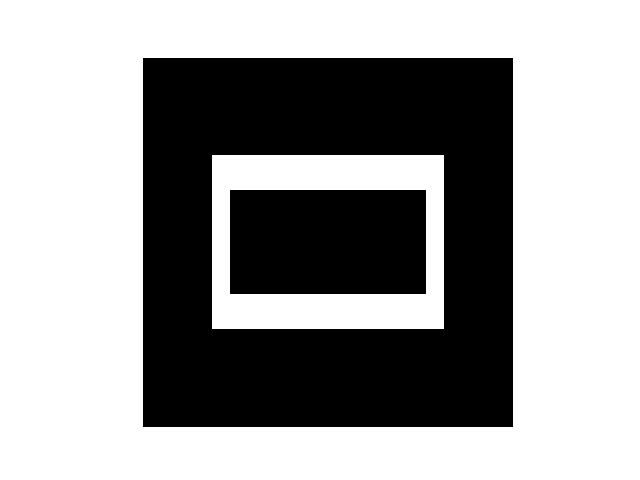} &
        \includegraphics[trim={2.5cm 1.0cm 2.5cm, 1.0cm}, clip,{width=.115\linewidth}]{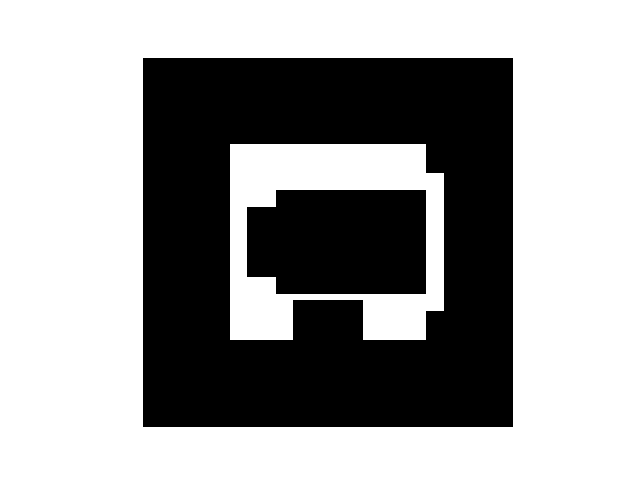} &
        \includegraphics[trim={2.5cm 1.0cm 2.5cm, 1.0cm}, clip,{width=.115\linewidth}]{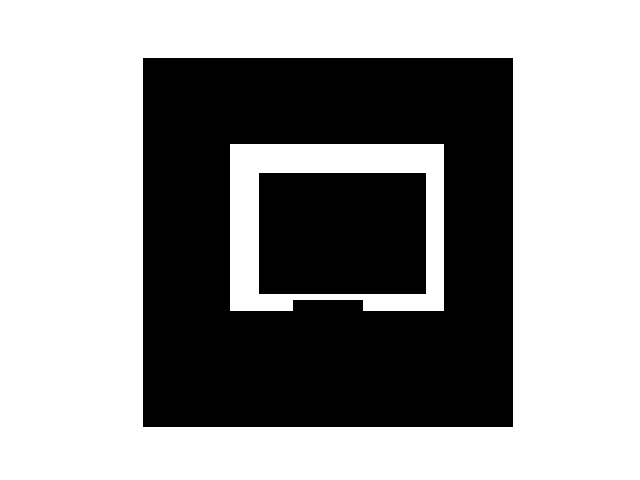} &
        \includegraphics[trim={2.5cm 1.0cm 2.5cm, 1.0cm}, clip,{width=.115\linewidth}]{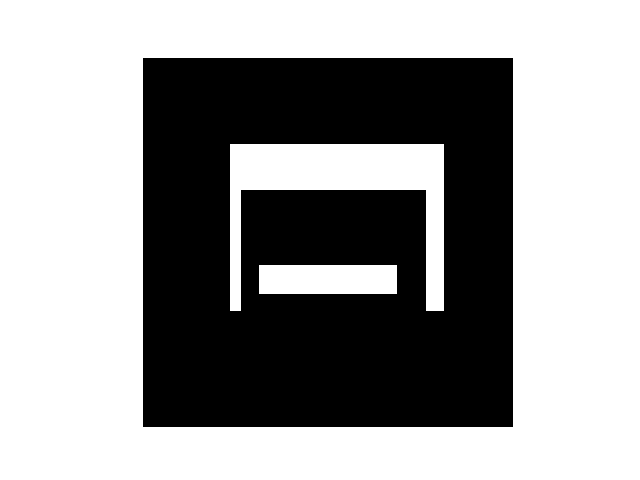} &
        \includegraphics[trim={2.5cm 1.0cm 2.5cm, 1.0cm}, clip,{width=.115\linewidth}]{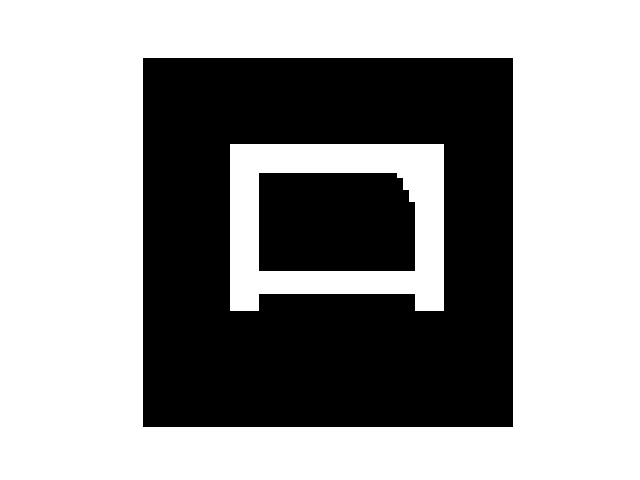} &
        \includegraphics[trim={2.5cm 1.0cm 2.5cm, 1.0cm}, clip,{width=.115\linewidth}]{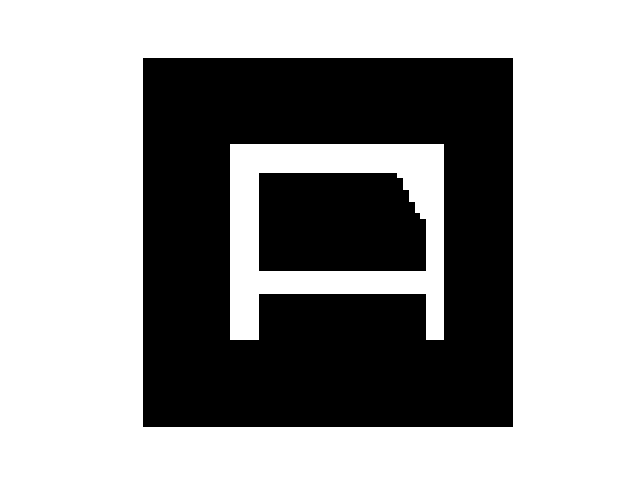} &
        \includegraphics[trim={2.5cm 1.0cm 2.5cm, 1.0cm}, clip,{width=.115\linewidth}]{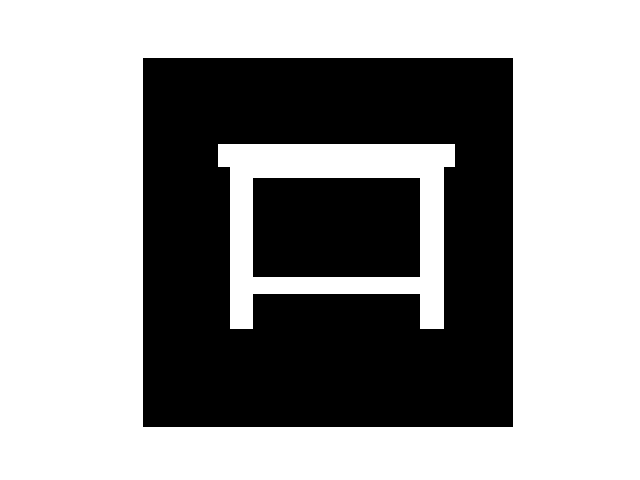} \\
        
        \includegraphics[trim={2.5cm 1.0cm 2.5cm, 1.0cm}, clip,{width=.115\linewidth}]{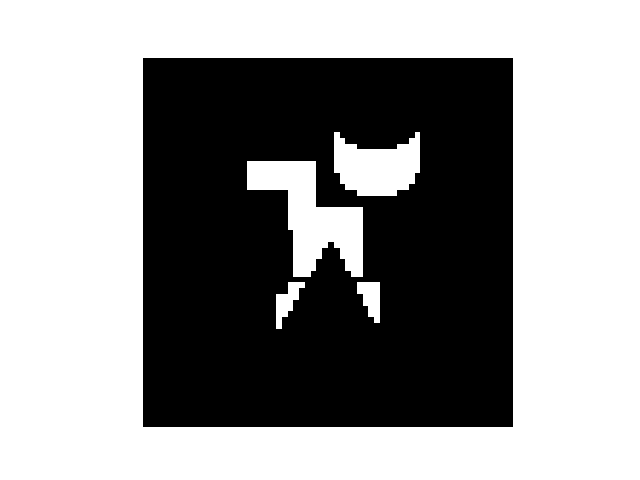} &
        \includegraphics[trim={2.5cm 1.0cm 2.5cm, 1.0cm}, clip,{width=.115\linewidth}]{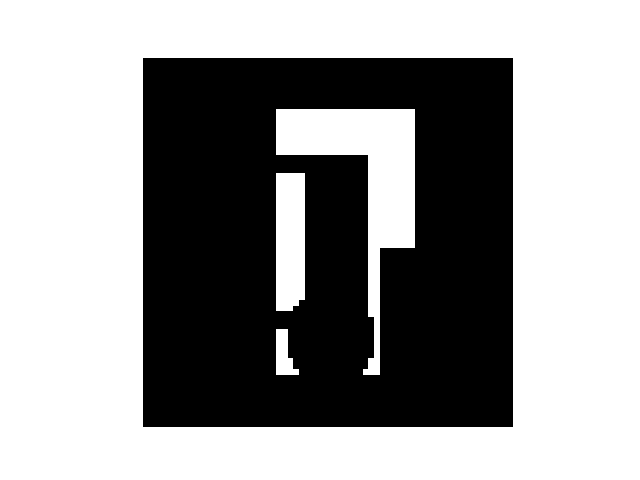} &
        \includegraphics[trim={2.5cm 1.0cm 2.5cm, 1.0cm}, clip,{width=.115\linewidth}]{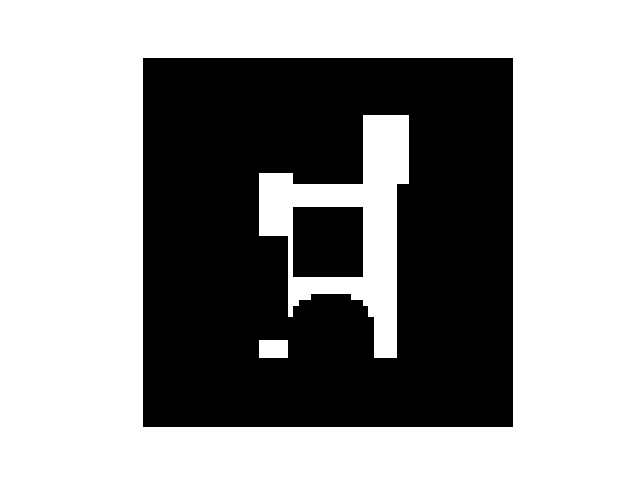} &
        \includegraphics[trim={2.5cm 1.0cm 2.5cm, 1.0cm}, clip,{width=.115\linewidth}]{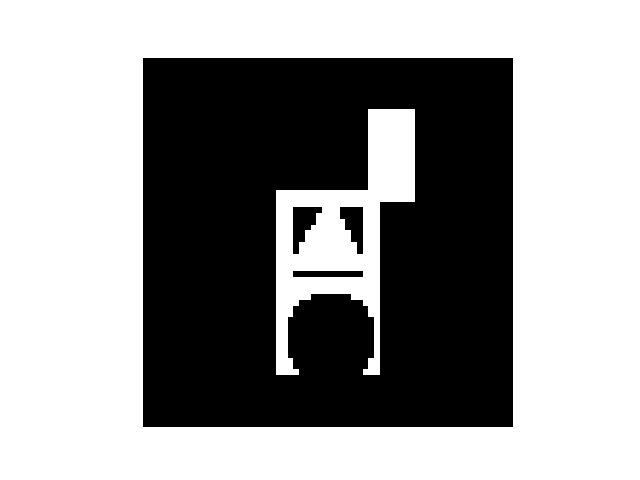} &
        \includegraphics[trim={2.5cm 1.0cm 2.5cm, 1.0cm}, clip,{width=.115\linewidth}]{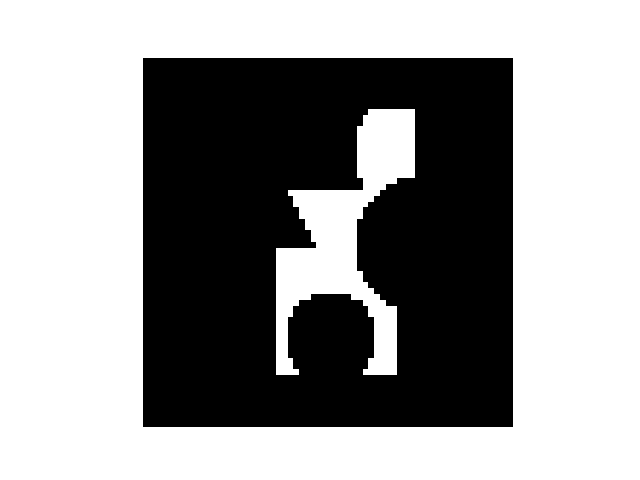} &
        \includegraphics[trim={2.5cm 1.0cm 2.5cm, 1.0cm}, clip,{width=.115\linewidth}]{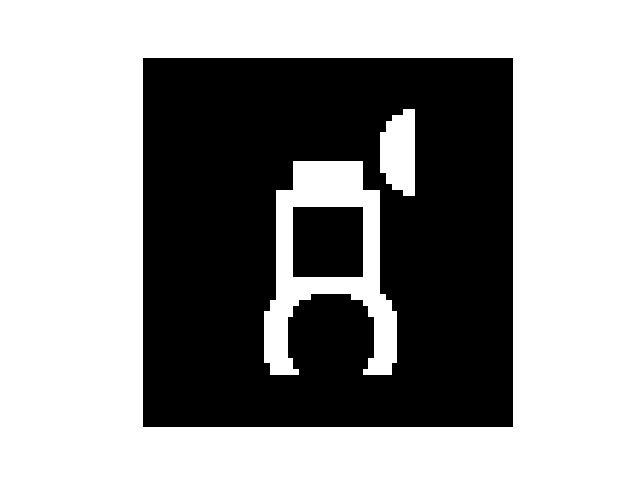} &
        \includegraphics[trim={2.5cm 1.0cm 2.5cm, 1.0cm}, clip,{width=.115\linewidth}]{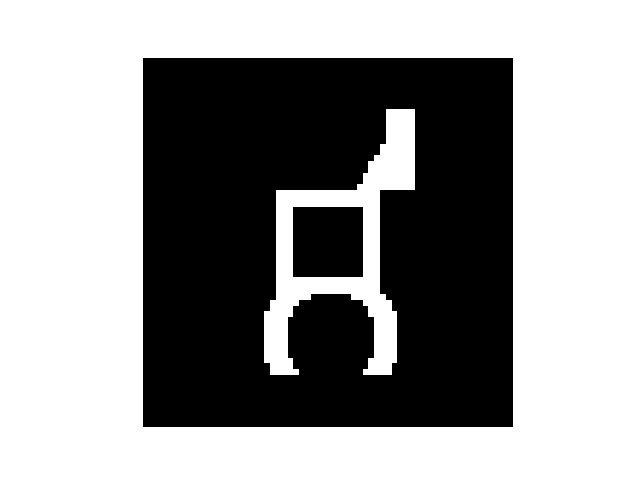} &
        \includegraphics[trim={2.5cm 1.0cm 2.5cm, 1.0cm}, clip,{width=.115\linewidth}]{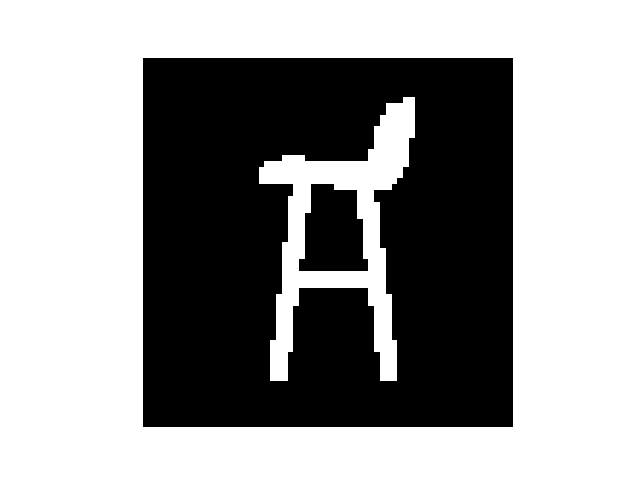} \\
        
        \includegraphics[trim={2.5cm 1.0cm 2.5cm, 1.0cm}, clip,{width=.115\linewidth}]{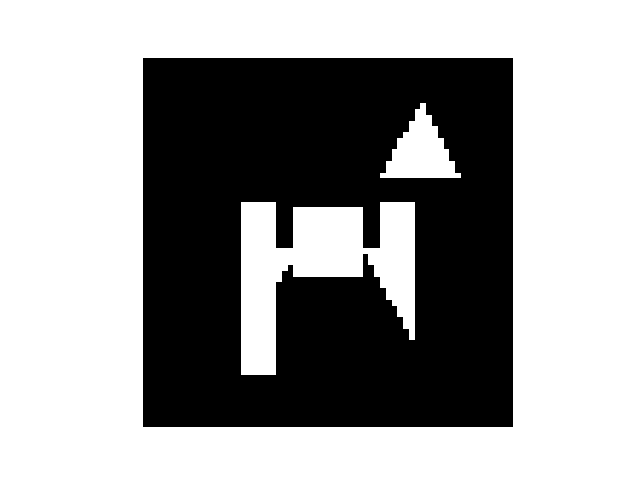} &
        \includegraphics[trim={2.5cm 1.0cm 2.5cm, 1.0cm}, clip,{width=.115\linewidth}]{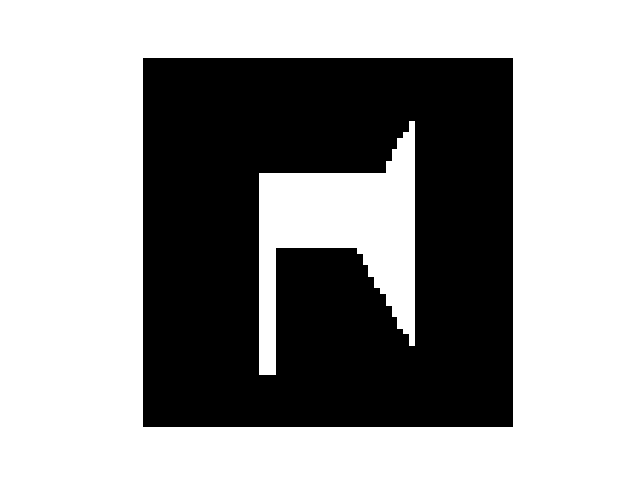} &
        \includegraphics[trim={2.5cm 1.0cm 2.5cm, 1.0cm}, clip,{width=.115\linewidth}]{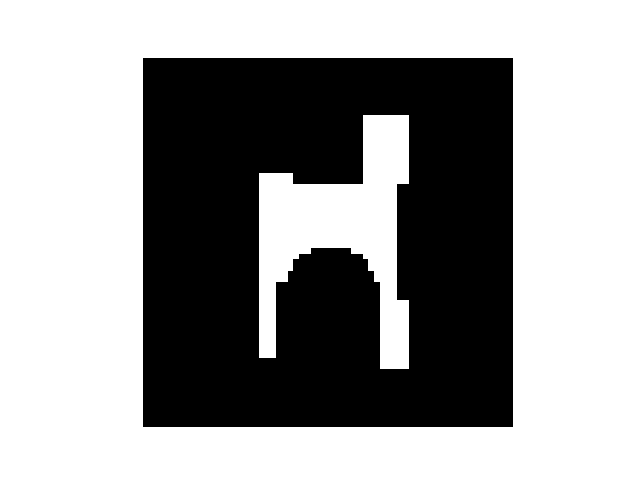} &
        \includegraphics[trim={2.5cm 1.0cm 2.5cm, 1.0cm}, clip,{width=.115\linewidth}]{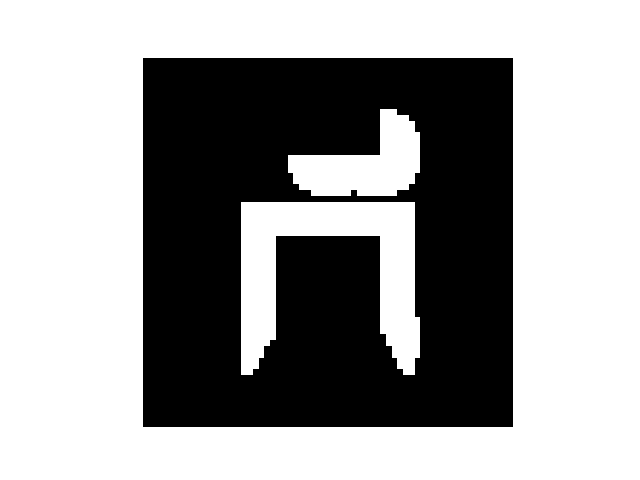} &
        \includegraphics[trim={2.5cm 1.0cm 2.5cm, 1.0cm}, clip,{width=.115\linewidth}]{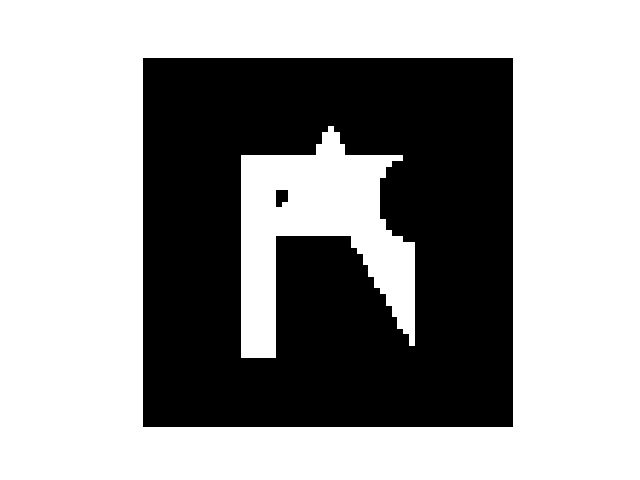} &
        \includegraphics[trim={2.5cm 1.0cm 2.5cm, 1.0cm}, clip,{width=.115\linewidth}]{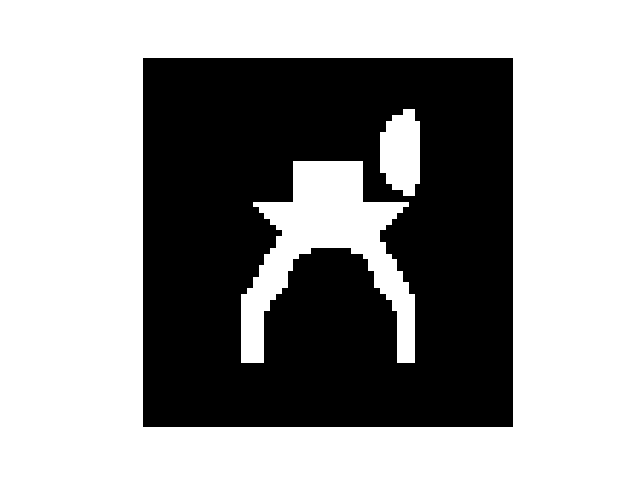} &
        \includegraphics[trim={2.5cm 1.0cm 2.5cm, 1.0cm}, clip,{width=.115\linewidth}]{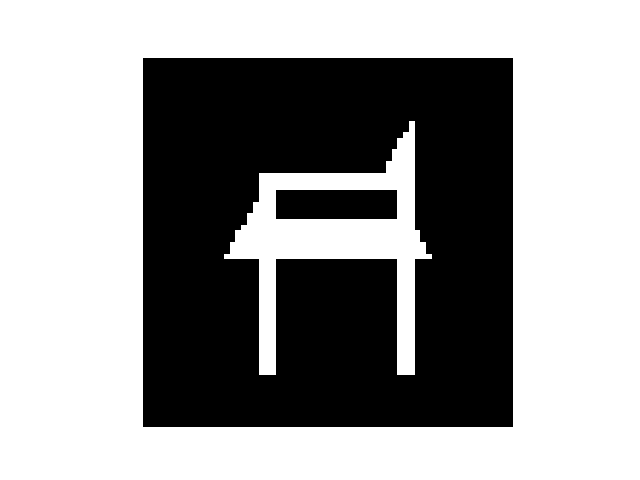} &
        \includegraphics[trim={2.5cm 1.0cm 2.5cm, 1.0cm}, clip,{width=.115\linewidth}]{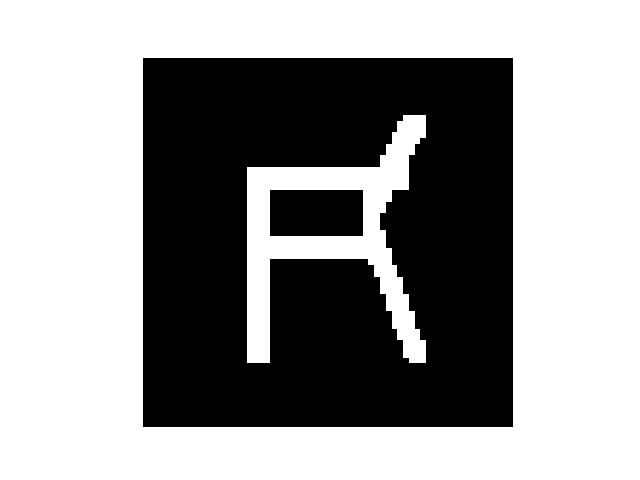} \\
        
        \includegraphics[trim={2.5cm 1.0cm 2.5cm, 1.0cm}, clip,{width=.115\linewidth}]{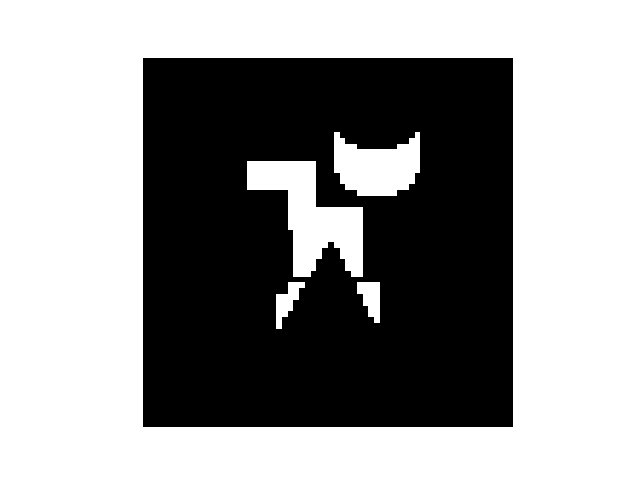} &
        \includegraphics[trim={2.5cm 1.0cm 2.5cm, 1.0cm}, clip,{width=.115\linewidth}]{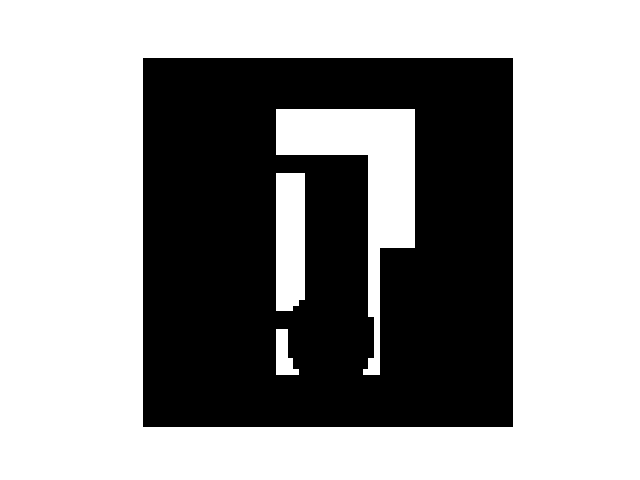} &
        \includegraphics[trim={2.5cm 1.0cm 2.5cm, 1.0cm}, clip,{width=.115\linewidth}]{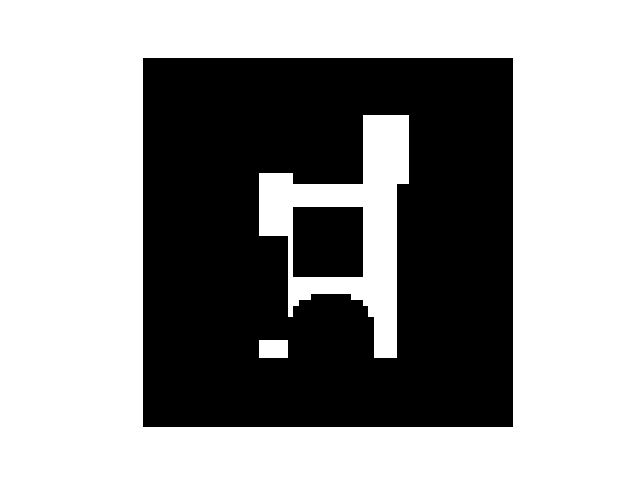} &
        \includegraphics[trim={2.5cm 1.0cm 2.5cm, 1.0cm}, clip,{width=.115\linewidth}]{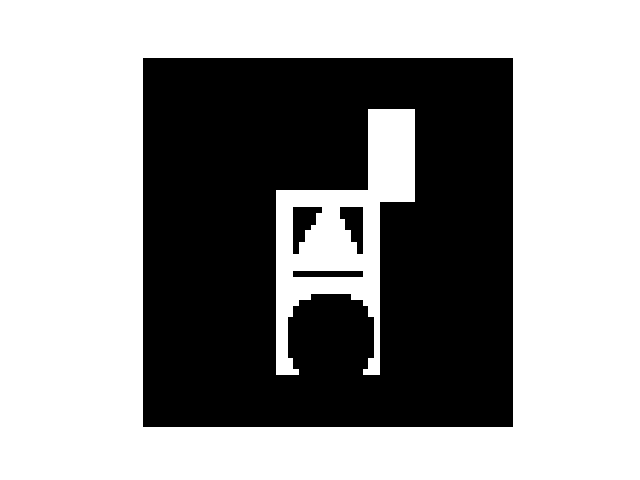} &
        \includegraphics[trim={2.5cm 1.0cm 2.5cm, 1.0cm}, clip,{width=.115\linewidth}]{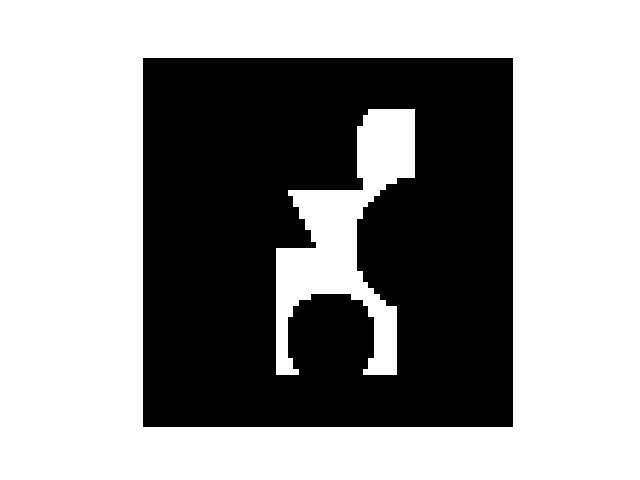} &
        \includegraphics[trim={2.5cm 1.0cm 2.5cm, 1.0cm}, clip,{width=.115\linewidth}]{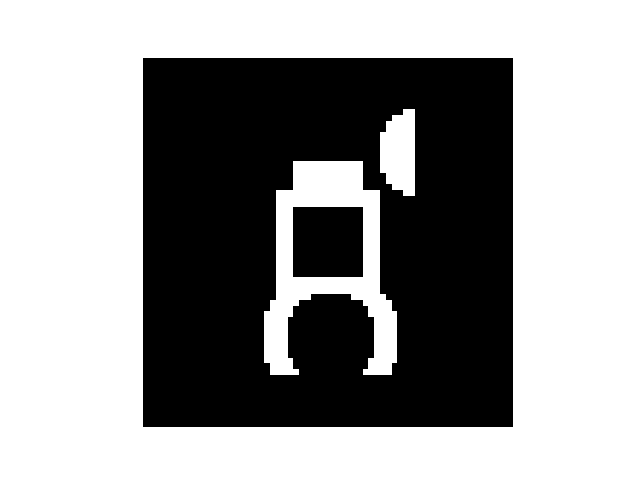} &
        \includegraphics[trim={2.5cm 1.0cm 2.5cm, 1.0cm}, clip,{width=.115\linewidth}]{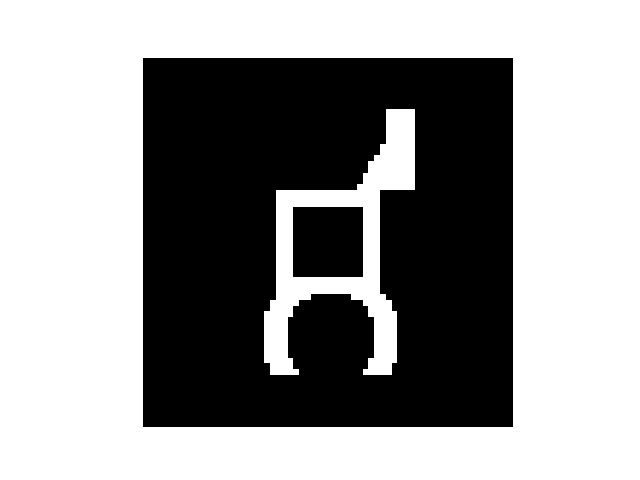} &
        \includegraphics[trim={2.5cm 1.0cm 2.5cm, 1.0cm}, clip,{width=.115\linewidth}]{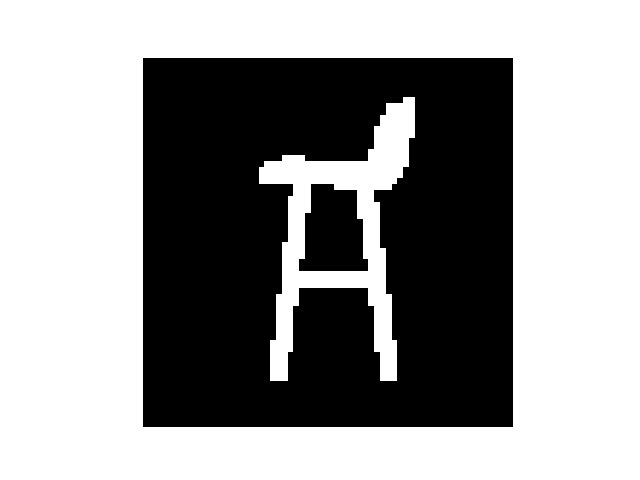} \\

    \end{tabular}
    \caption{2DCSG qualitative examples.} 
    \label{fig:qual_2d}
\end{figure*}

\begin{figure*}[t!]
    \centering
    \footnotesize
    \setlength{\tabcolsep}{1pt}
    \begin{tabular}{cccccccc}
        \textbf{SP} &  \textbf{WS} &  \textbf{RL} &  \textbf{ST} &  \textbf{LEST} & \textbf{LEST+ST} & \textbf{LEST+ST+WS} & \textbf{Target} \\
        
        \includegraphics[trim={3.5cm 4.5cm 3.5cm, 4.5cm}, clip,{width=.115\linewidth}]{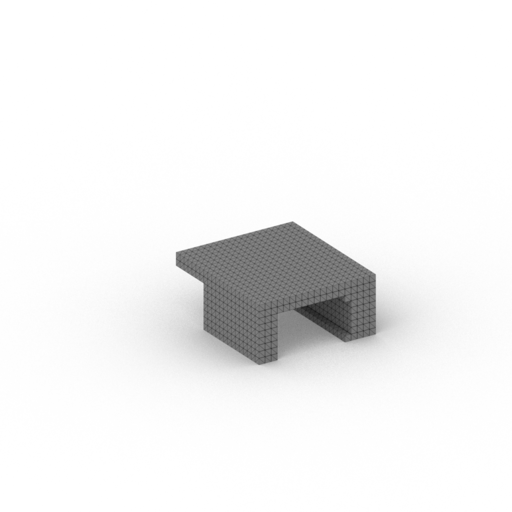} &
        \includegraphics[trim={3.5cm 4.5cm 3.5cm, 4.5cm}, clip,{width=.115\linewidth}]{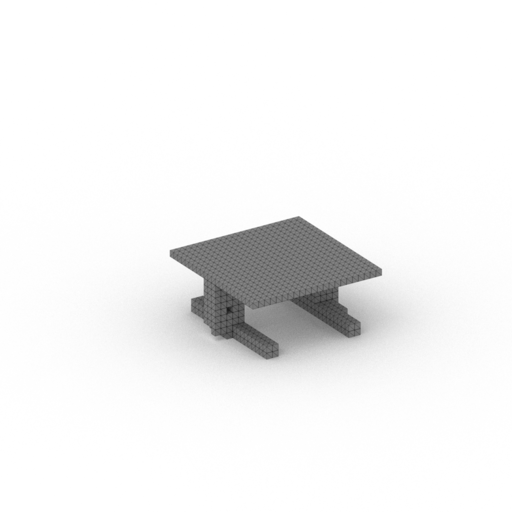} &
        \includegraphics[trim={3.5cm 4.5cm 3.5cm, 4.5cm}, clip,{width=.115\linewidth}]{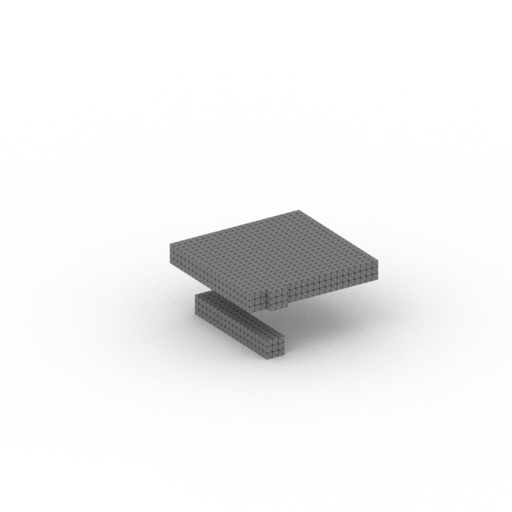} &
        \includegraphics[trim={3.5cm 4.5cm 3.5cm, 4.5cm}, clip,{width=.115\linewidth}]{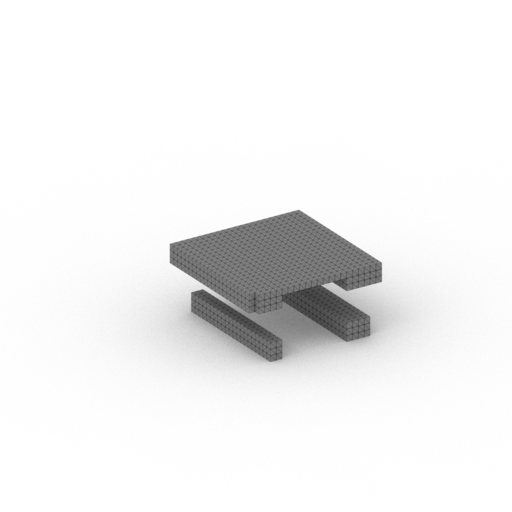} &
        \includegraphics[trim={3.5cm 4.5cm 3.5cm, 4.5cm}, clip,{width=.115\linewidth}]{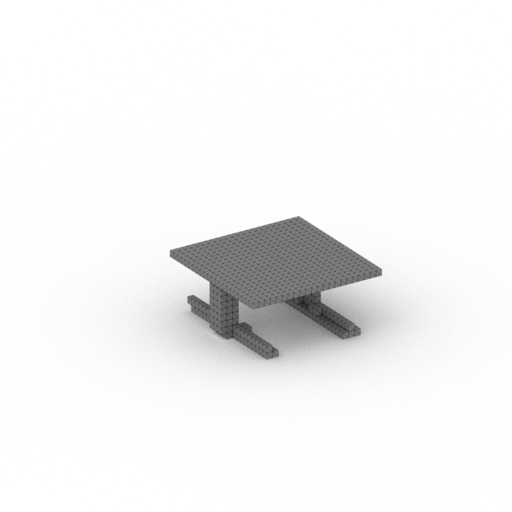} &
        \includegraphics[trim={3.5cm 4.5cm 3.5cm, 4.5cm}, clip,{width=.115\linewidth}]{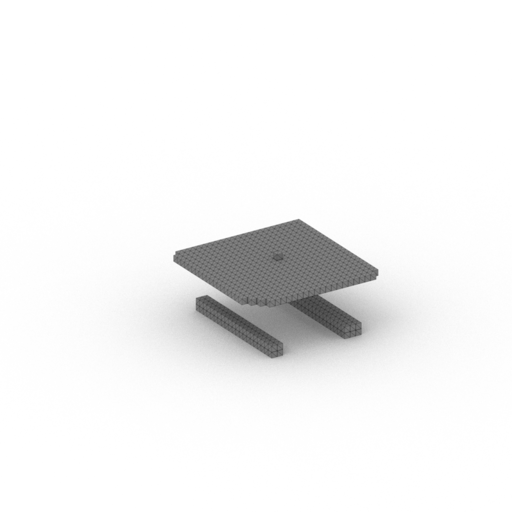} &
        \includegraphics[trim={3.5cm 4.5cm 3.5cm, 4.5cm}, clip,{width=.115\linewidth}]{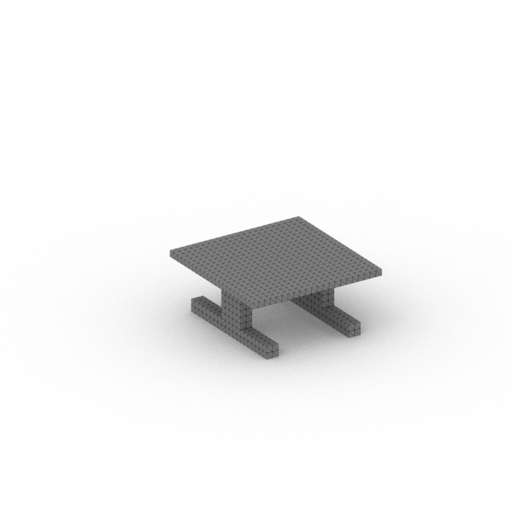} &
        \includegraphics[trim={3.5cm 4.5cm 3.5cm, 4.5cm}, clip,{width=.115\linewidth}]{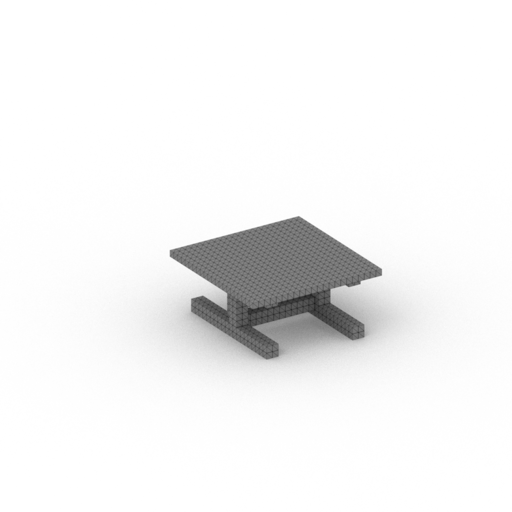} \\
        
        \includegraphics[trim={3.5cm 4.5cm 3.5cm, 4.5cm}, clip,{width=.115\linewidth}]{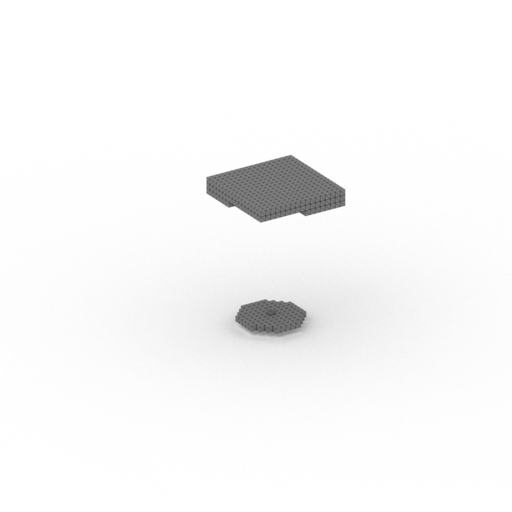} &
        \includegraphics[trim={3.5cm 4.5cm 3.5cm, 4.5cm}, clip,{width=.115\linewidth}]{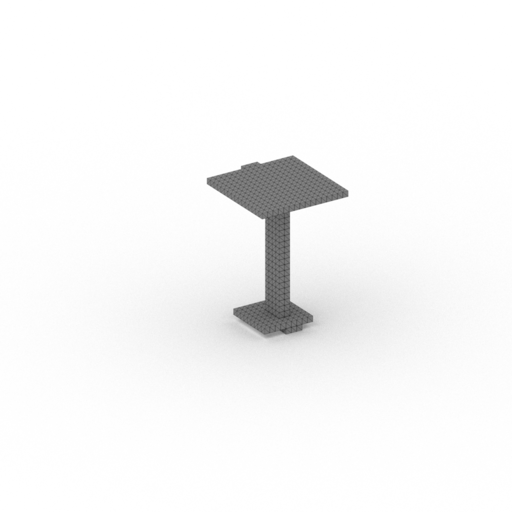} &
        \includegraphics[trim={3.5cm 4.5cm 3.5cm, 4.5cm}, clip,{width=.115\linewidth}]{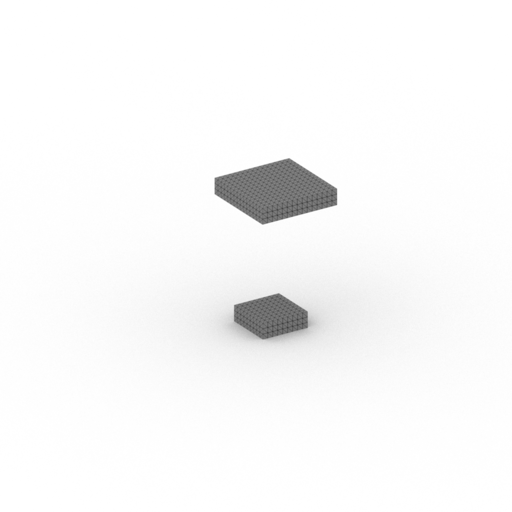} &
        \includegraphics[trim={3.5cm 4.5cm 3.5cm, 4.5cm}, clip,{width=.115\linewidth}]{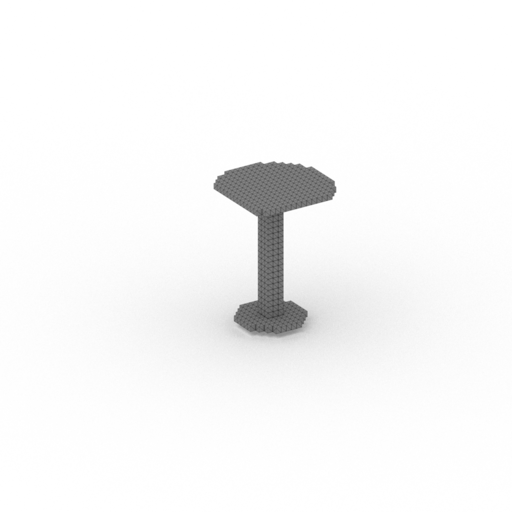} &
        \includegraphics[trim={3.5cm 4.5cm 3.5cm, 4.5cm}, clip,{width=.115\linewidth}]{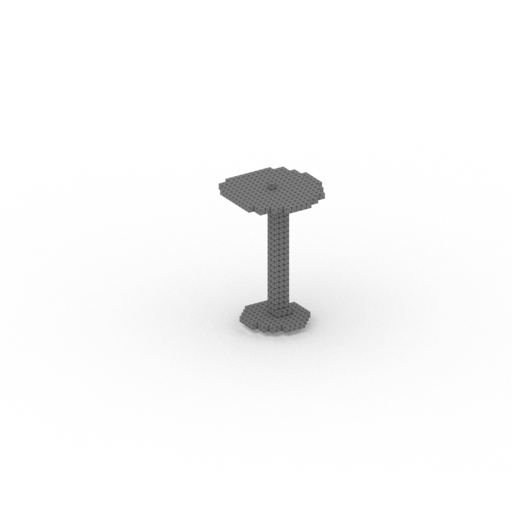} &
        \includegraphics[trim={3.5cm 4.5cm 3.5cm, 4.5cm}, clip,{width=.115\linewidth}]{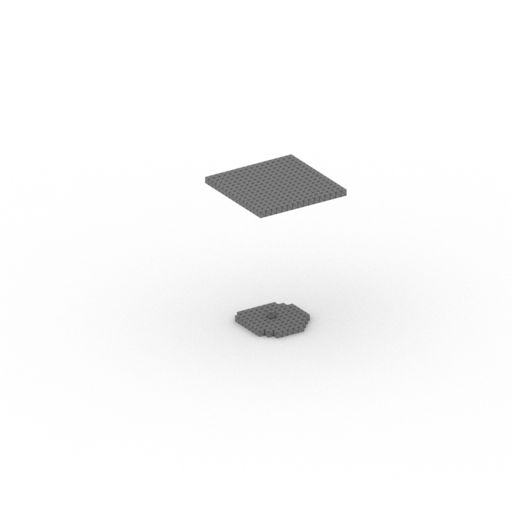} &
        \includegraphics[trim={3.5cm 4.5cm 3.5cm, 4.5cm}, clip,{width=.115\linewidth}]{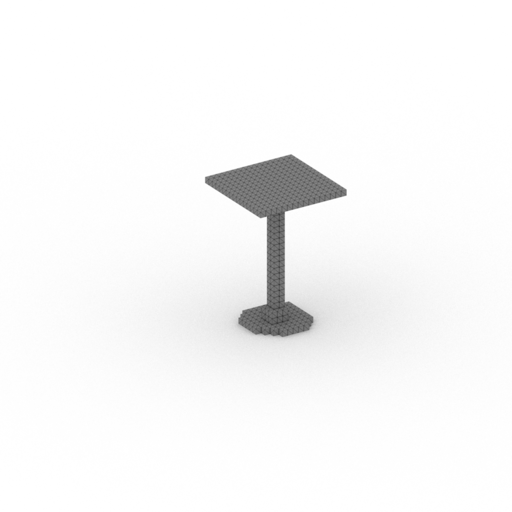} &
        \includegraphics[trim={3.5cm 4.5cm 3.5cm, 4.5cm}, clip,{width=.115\linewidth}]{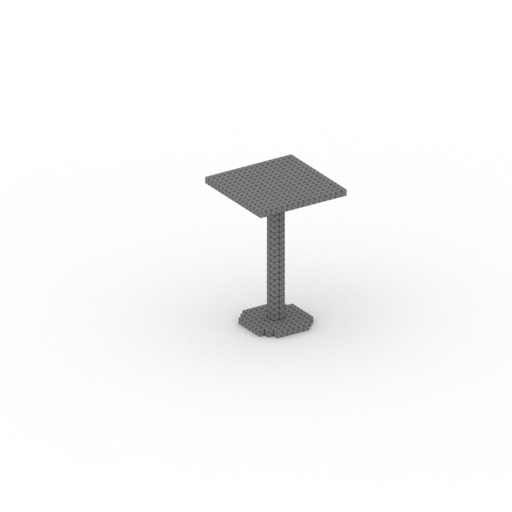} \\
        
        \includegraphics[trim={3.5cm 4.5cm 3.5cm, 4.5cm}, clip,{width=.115\linewidth}]{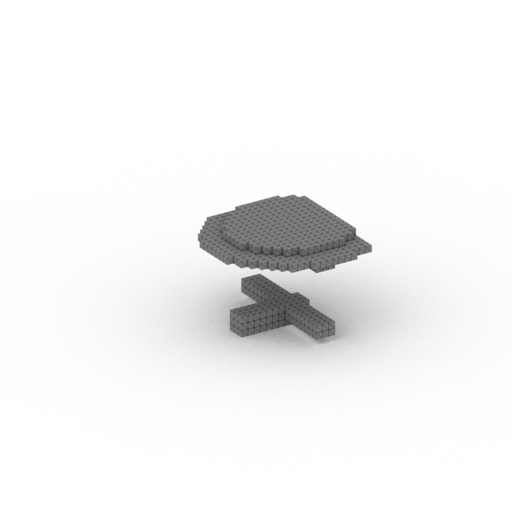} &
        \includegraphics[trim={3.5cm 4.5cm 3.5cm, 4.5cm}, clip,{width=.115\linewidth}]{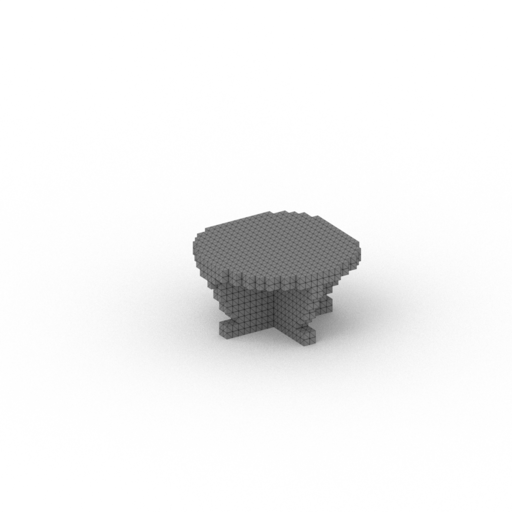} &
        \includegraphics[trim={3.5cm 4.5cm 3.5cm, 4.5cm}, clip,{width=.115\linewidth}]{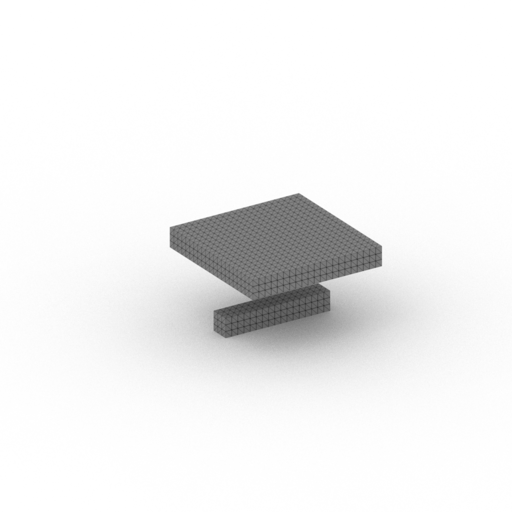} &
        \includegraphics[trim={3.5cm 4.5cm 3.5cm, 4.5cm}, clip,{width=.115\linewidth}]{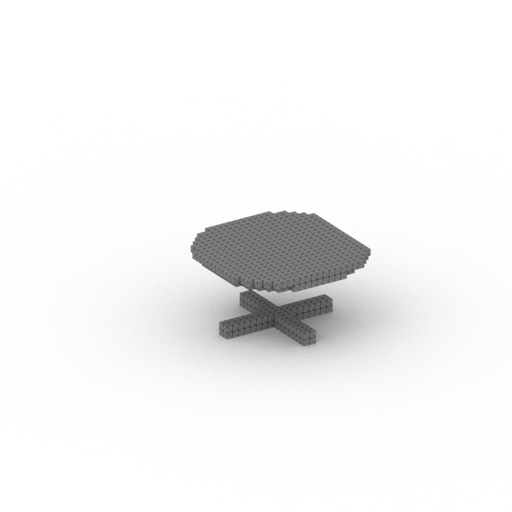} &
        \includegraphics[trim={3.5cm 4.5cm 3.5cm, 4.5cm}, clip,{width=.115\linewidth}]{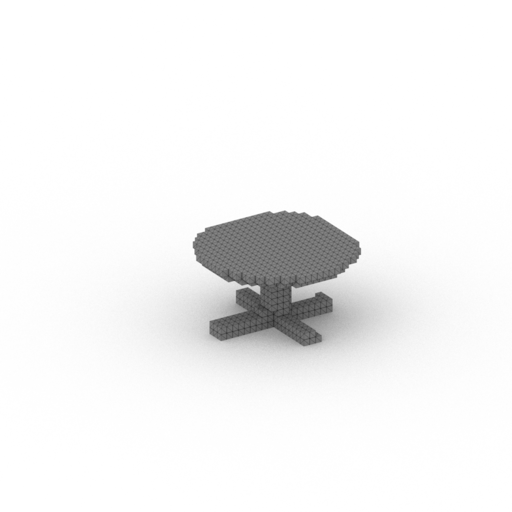} &
        \includegraphics[trim={3.5cm 4.5cm 3.5cm, 4.5cm}, clip,{width=.115\linewidth}]{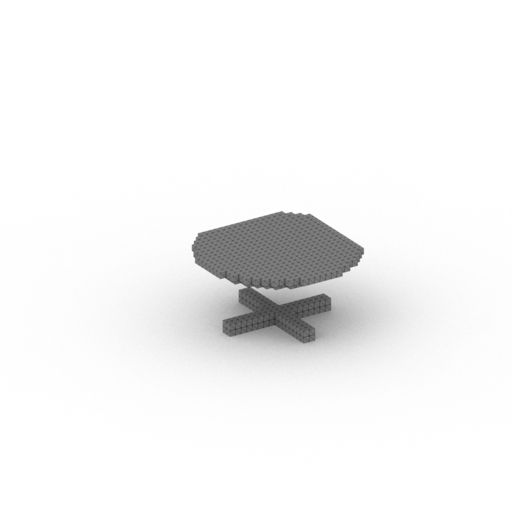} &
        \includegraphics[trim={3.5cm 4.5cm 3.5cm, 4.5cm}, clip,{width=.115\linewidth}]{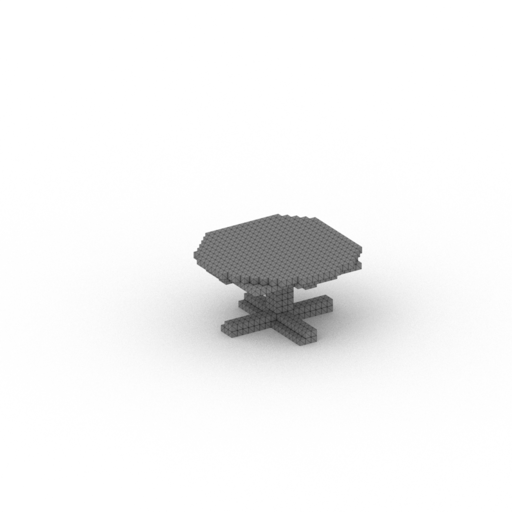} &
        \includegraphics[trim={3.5cm 4.5cm 3.5cm, 4.5cm}, clip,{width=.115\linewidth}]{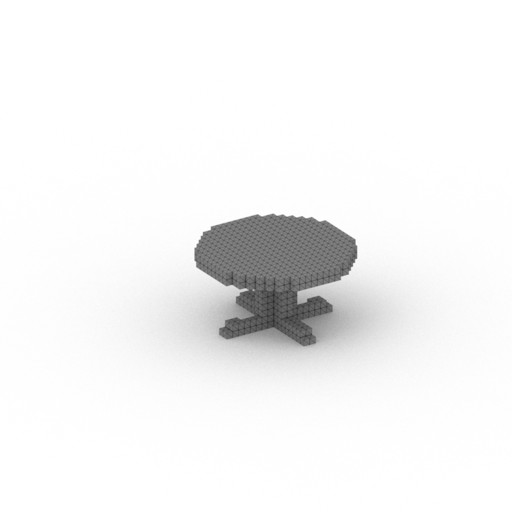} \\
        
        \includegraphics[trim={3.5cm 4.5cm 3.5cm, 4.5cm}, clip,{width=.115\linewidth}]{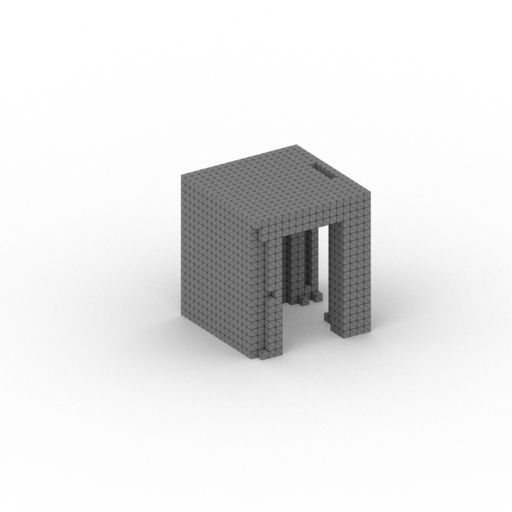} &
        \includegraphics[trim={3.5cm 4.5cm 3.5cm, 4.5cm}, clip,{width=.115\linewidth}]{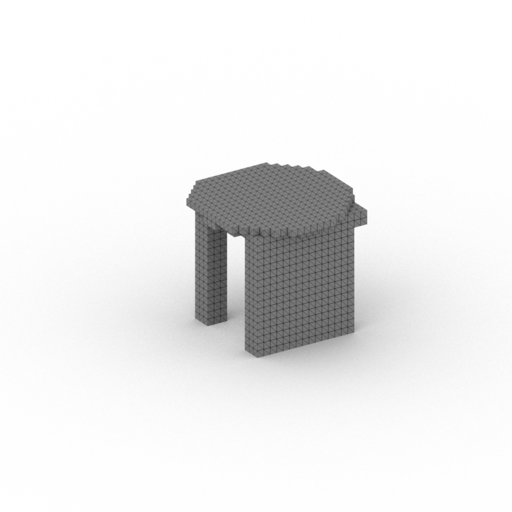} &
        \includegraphics[trim={3.5cm 4.5cm 3.5cm, 4.5cm}, clip,{width=.115\linewidth}]{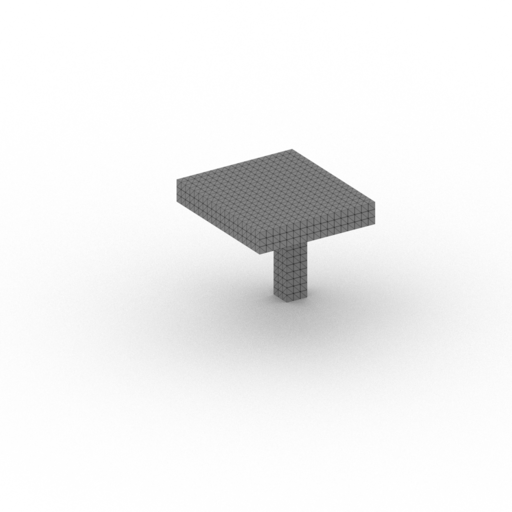} &
        \includegraphics[trim={3.5cm 4.5cm 3.5cm, 4.5cm}, clip,{width=.115\linewidth}]{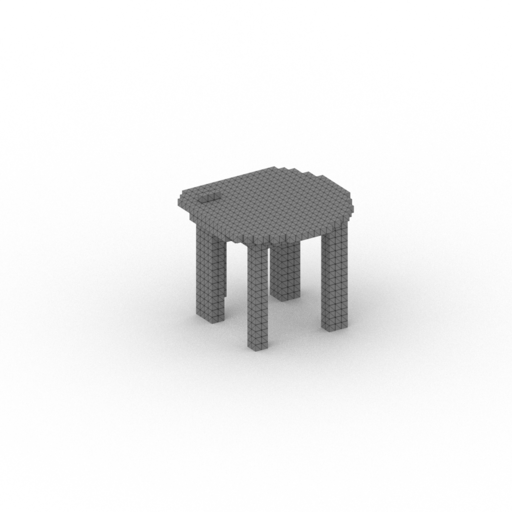} &
        \includegraphics[trim={3.5cm 4.5cm 3.5cm, 4.5cm}, clip,{width=.115\linewidth}]{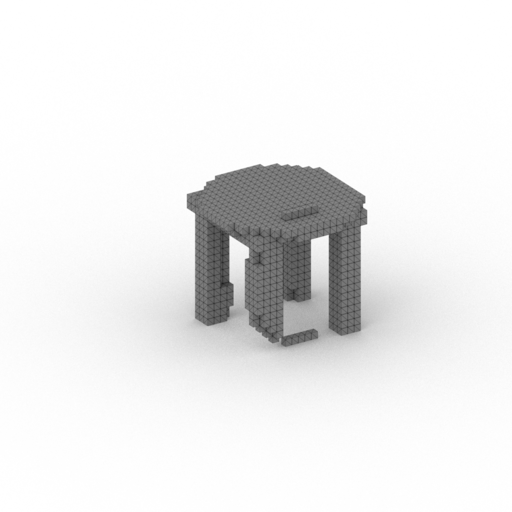} &
        \includegraphics[trim={3.5cm 4.5cm 3.5cm, 4.5cm}, clip,{width=.115\linewidth}]{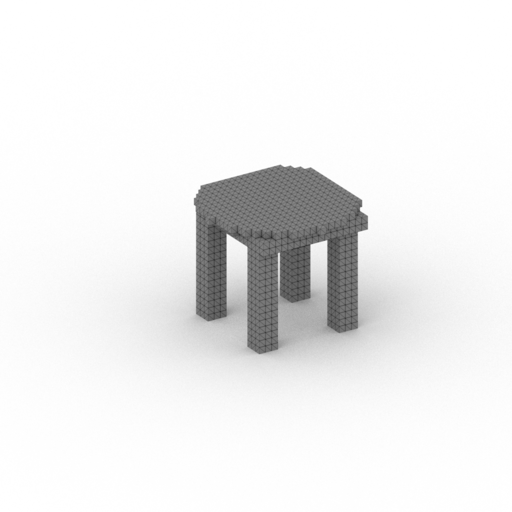} &
        \includegraphics[trim={3.5cm 4.5cm 3.5cm, 4.5cm}, clip,{width=.115\linewidth}]{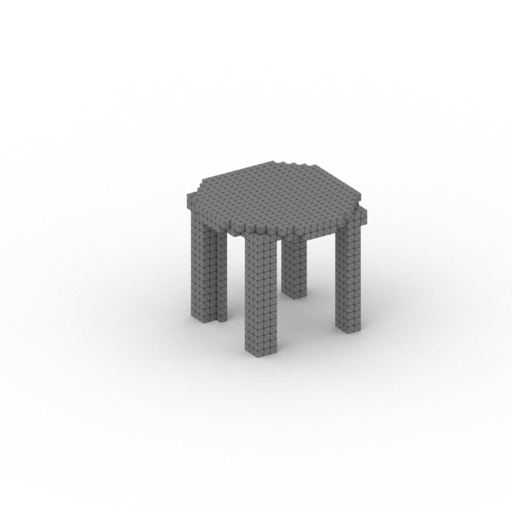} &
        \includegraphics[trim={3.5cm 4.5cm 3.5cm, 4.5cm}, clip,{width=.115\linewidth}]{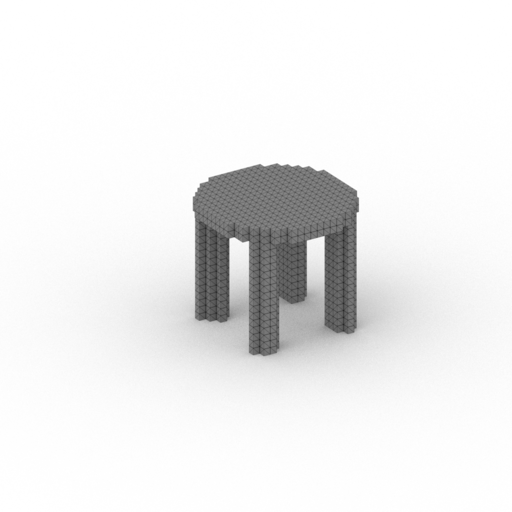} \\
        
        \includegraphics[trim={3.5cm 4.5cm 3.5cm, 4.5cm}, clip,{width=.115\linewidth}]{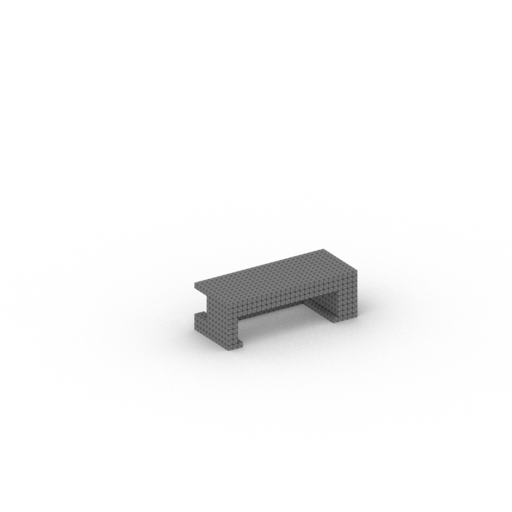} &
        \includegraphics[trim={3.5cm 4.5cm 3.5cm, 4.5cm}, clip,{width=.115\linewidth}]{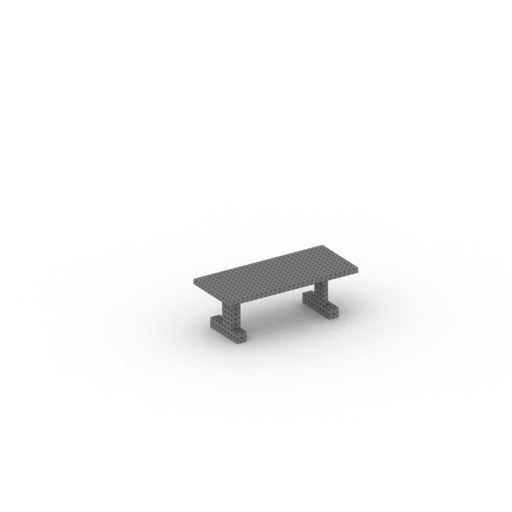} &
        \includegraphics[trim={3.5cm 4.5cm 3.5cm, 4.5cm}, clip,{width=.115\linewidth}]{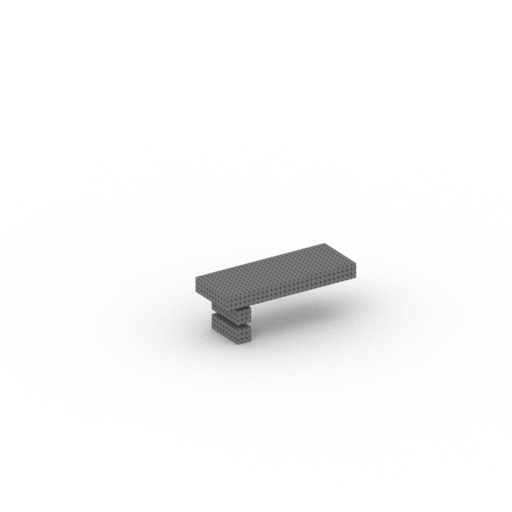} &
        \includegraphics[trim={3.5cm 4.5cm 3.5cm, 4.5cm}, clip,{width=.115\linewidth}]{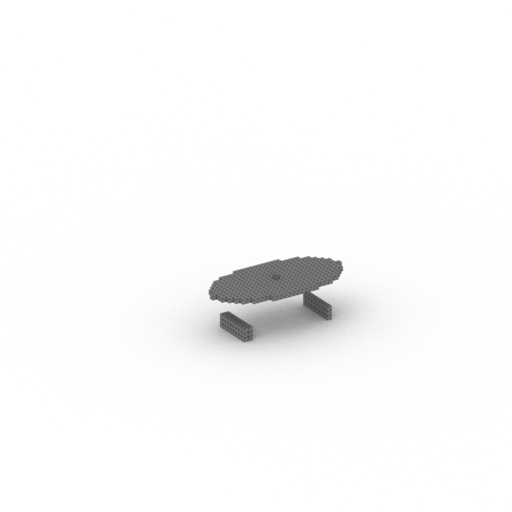} &
        \includegraphics[trim={3.5cm 4.5cm 3.5cm, 4.5cm}, clip,{width=.115\linewidth}]{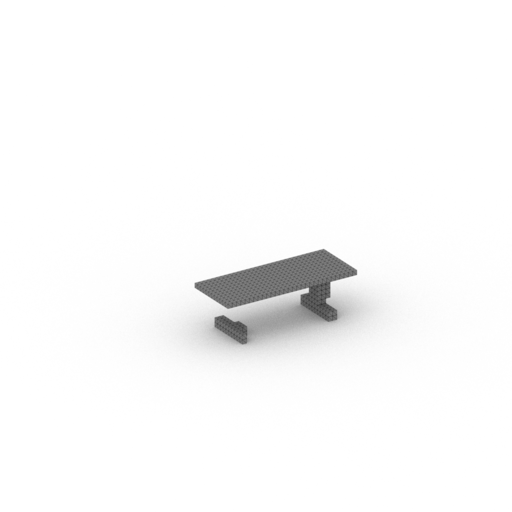} &
        \includegraphics[trim={3.5cm 4.5cm 3.5cm, 4.5cm}, clip,{width=.115\linewidth}]{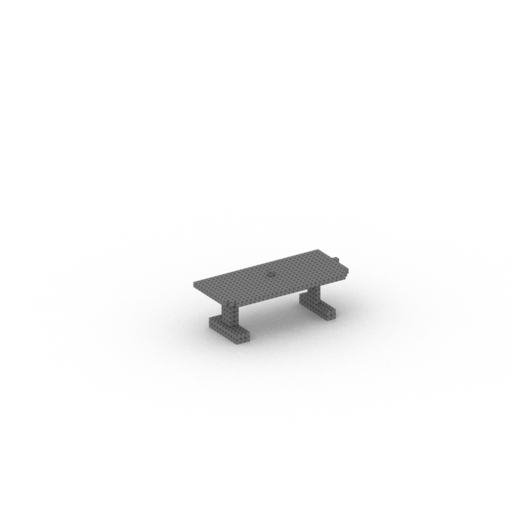} &
        \includegraphics[trim={3.5cm 4.5cm 3.5cm, 4.5cm}, clip,{width=.115\linewidth}]{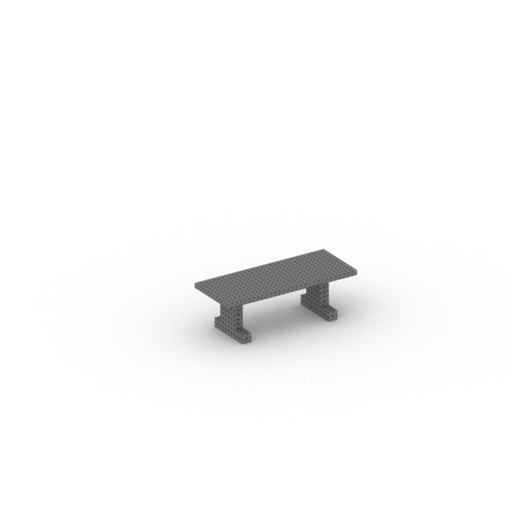} &
        \includegraphics[trim={3.5cm 4.5cm 3.5cm, 4.5cm}, clip,{width=.115\linewidth}]{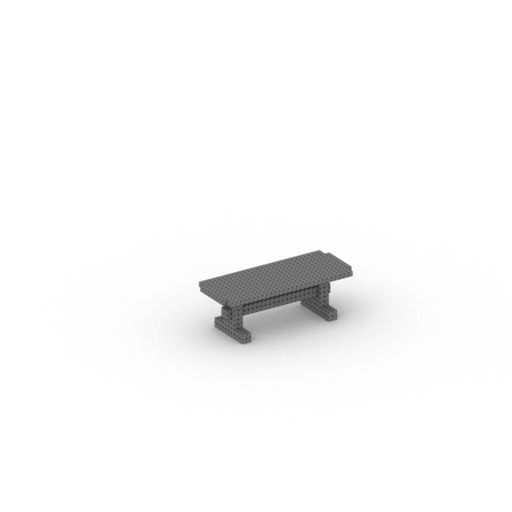} \\
        
        \includegraphics[trim={3.5cm 4.5cm 3.5cm, 4.5cm}, clip,{width=.115\linewidth}]{figs/qual/3DCSG/806_sp.png} &
        \includegraphics[trim={3.5cm 4.5cm 3.5cm, 4.5cm}, clip,{width=.115\linewidth}]{figs/qual/3DCSG/806_ws.png} &
        \includegraphics[trim={3.5cm 4.5cm 3.5cm, 4.5cm}, clip,{width=.115\linewidth}]{figs/qual/3DCSG/806_rl.png} &
        \includegraphics[trim={3.5cm 4.5cm 3.5cm, 4.5cm}, clip,{width=.115\linewidth}]{figs/qual/3DCSG/806_st.png} &
        \includegraphics[trim={3.5cm 4.5cm 3.5cm, 4.5cm}, clip,{width=.115\linewidth}]{figs/qual/3DCSG/806_lest.png} &
        \includegraphics[trim={3.5cm 4.5cm 3.5cm, 4.5cm}, clip,{width=.115\linewidth}]{figs/qual/3DCSG/806_lest_st.png} &
        \includegraphics[trim={3.5cm 4.5cm 3.5cm, 4.5cm}, clip,{width=.115\linewidth}]{figs/qual/3DCSG/806_lest_st_ws.png} &
        \includegraphics[trim={3.5cm 4.5cm 3.5cm, 4.5cm}, clip,{width=.115\linewidth}]{figs/qual/3DCSG/806_gt.png} \\
        
        \includegraphics[trim={3.5cm 4.5cm 3.5cm, 4.5cm}, clip,{width=.115\linewidth}]{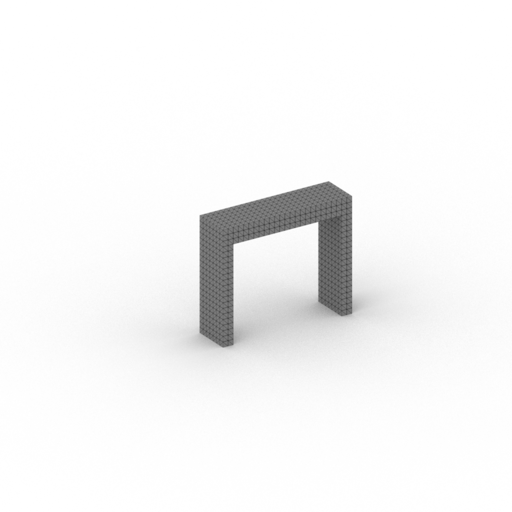} &
        \includegraphics[trim={3.5cm 4.5cm 3.5cm, 4.5cm}, clip,{width=.115\linewidth}]{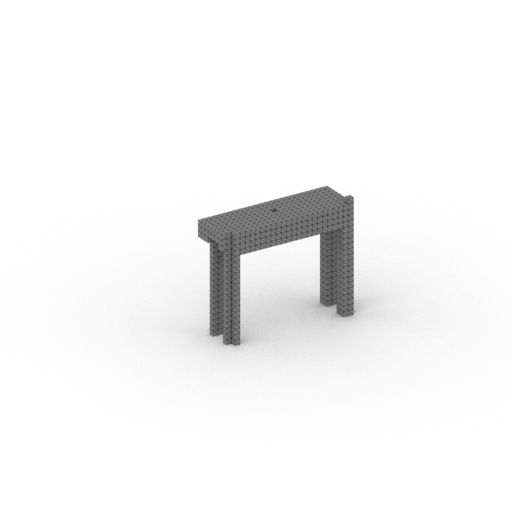} &
        \includegraphics[trim={3.5cm 4.5cm 3.5cm, 4.5cm}, clip,{width=.115\linewidth}]{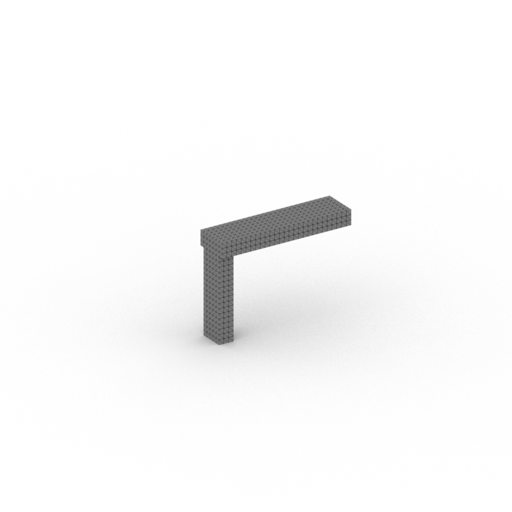} &
        \includegraphics[trim={3.5cm 4.5cm 3.5cm, 4.5cm}, clip,{width=.115\linewidth}]{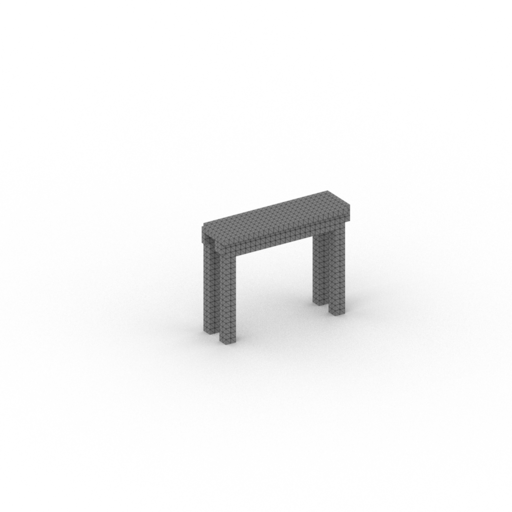} &
        \includegraphics[trim={3.5cm 4.5cm 3.5cm, 4.5cm}, clip,{width=.115\linewidth}]{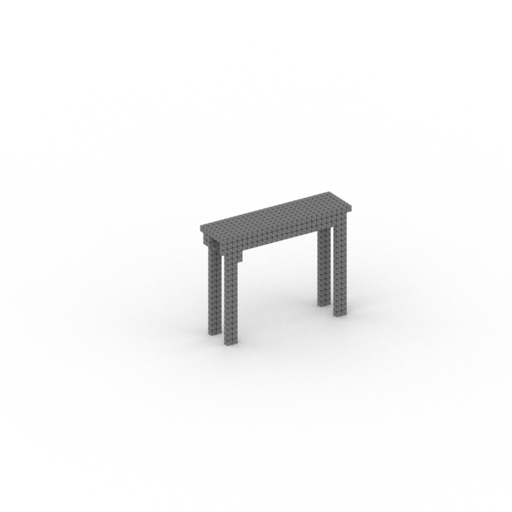} &
        \includegraphics[trim={3.5cm 4.5cm 3.5cm, 4.5cm}, clip,{width=.115\linewidth}]{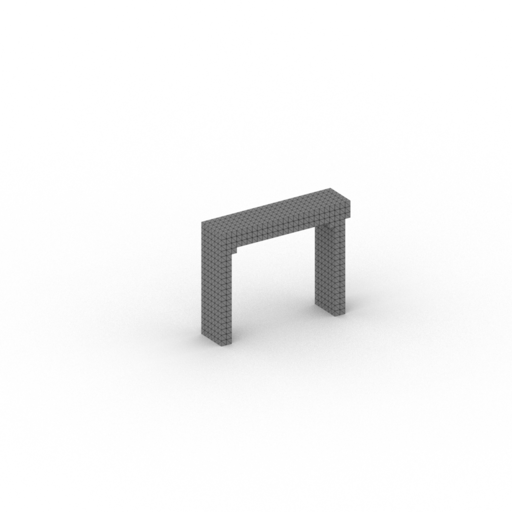} &
        \includegraphics[trim={3.5cm 4.5cm 3.5cm, 4.5cm}, clip,{width=.115\linewidth}]{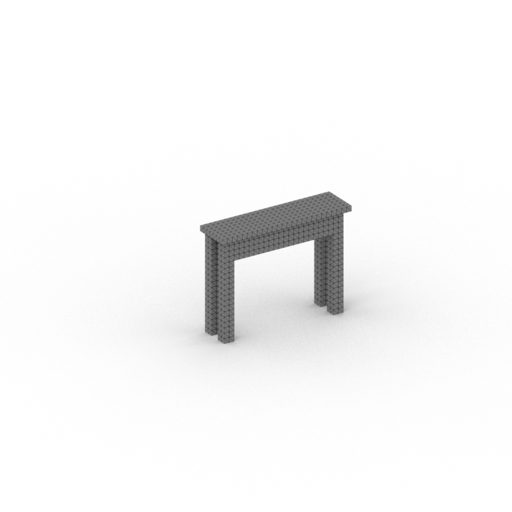} &
        \includegraphics[trim={3.5cm 4.5cm 3.5cm, 4.5cm}, clip,{width=.115\linewidth}]{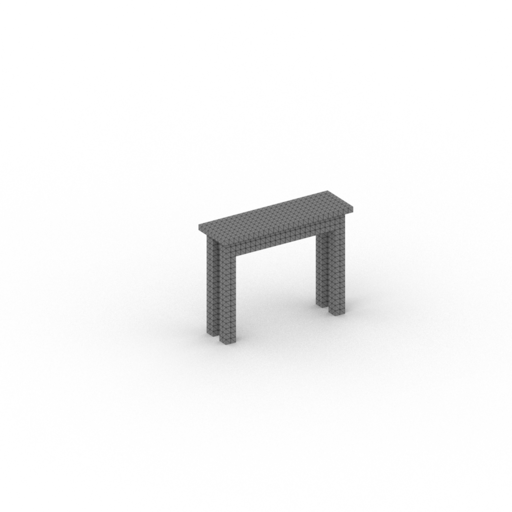} \\
        
        \includegraphics[trim={3.5cm 4.5cm 3.5cm, 4.5cm}, clip,{width=.115\linewidth}]{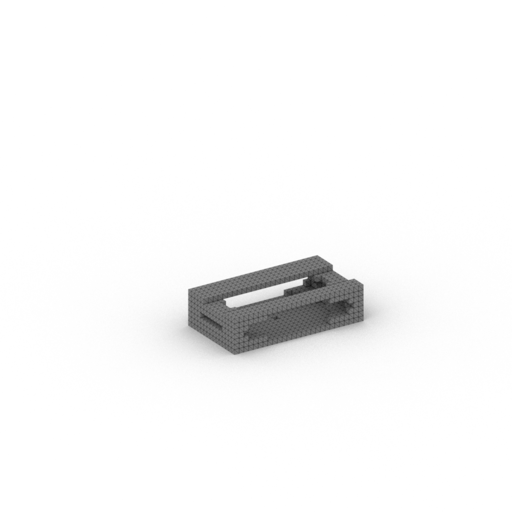} &
        \includegraphics[trim={3.5cm 4.5cm 3.5cm, 4.5cm}, clip,{width=.115\linewidth}]{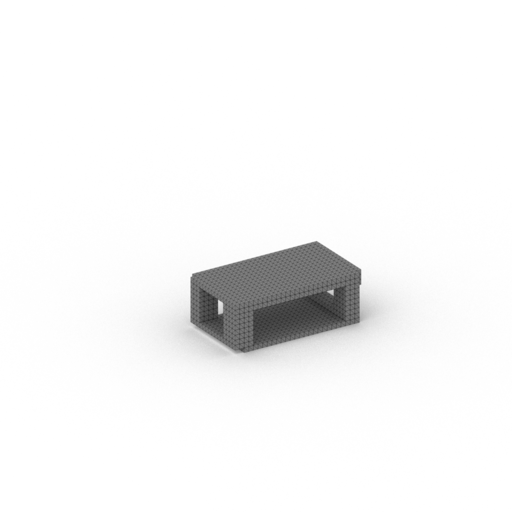} &
        \includegraphics[trim={3.5cm 4.5cm 3.5cm, 4.5cm}, clip,{width=.115\linewidth}]{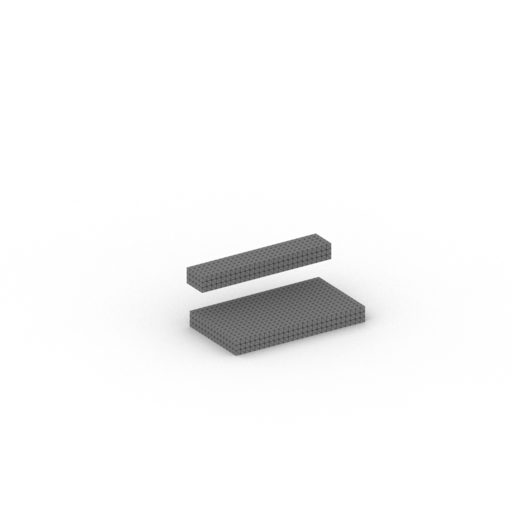} &
        \includegraphics[trim={3.5cm 4.5cm 3.5cm, 4.5cm}, clip,{width=.115\linewidth}]{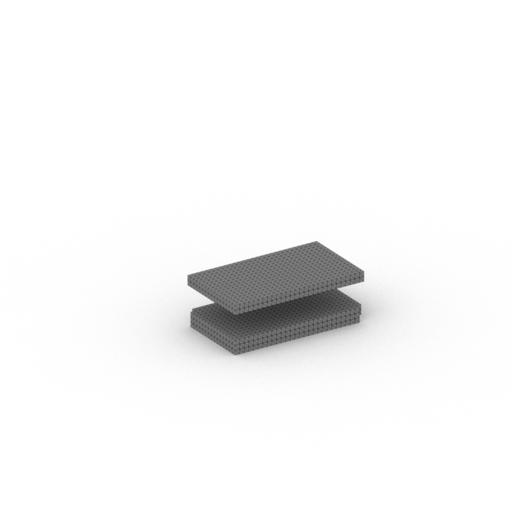} &
        \includegraphics[trim={3.5cm 4.5cm 3.5cm, 4.5cm}, clip,{width=.115\linewidth}]{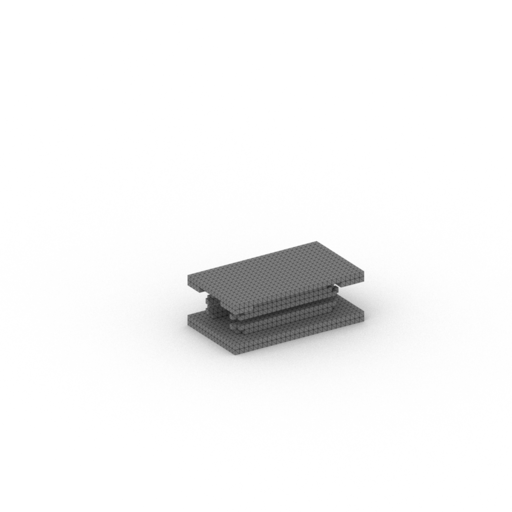} &
        \includegraphics[trim={3.5cm 4.5cm 3.5cm, 4.5cm}, clip,{width=.115\linewidth}]{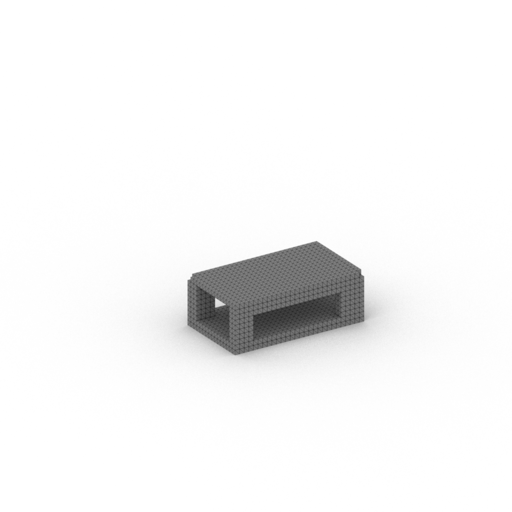} &
        \includegraphics[trim={3.5cm 4.5cm 3.5cm, 4.5cm}, clip,{width=.115\linewidth}]{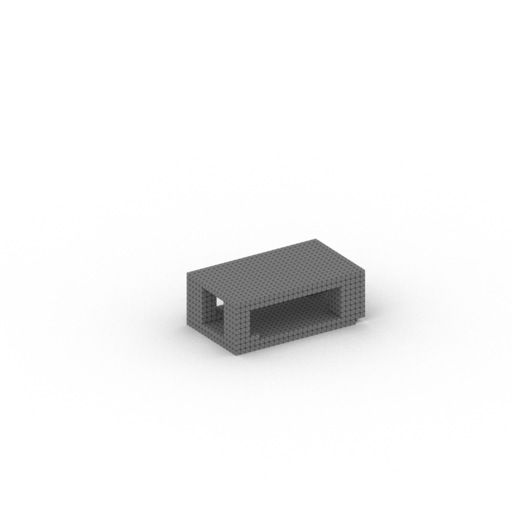} &
        \includegraphics[trim={3.5cm 4.5cm 3.5cm, 4.5cm}, clip,{width=.115\linewidth}]{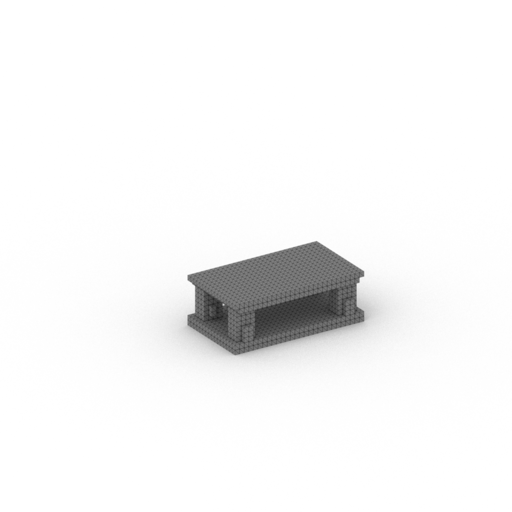} \\
        
        \includegraphics[trim={3.5cm 4.5cm 3.5cm, 4.5cm}, clip,{width=.115\linewidth}]{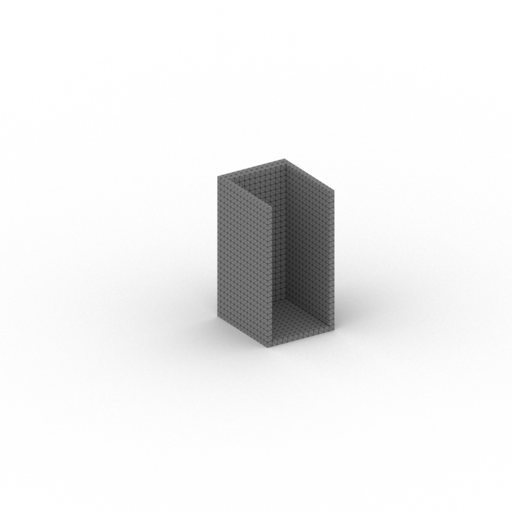} &
        \includegraphics[trim={3.5cm 4.5cm 3.5cm, 4.5cm}, clip,{width=.115\linewidth}]{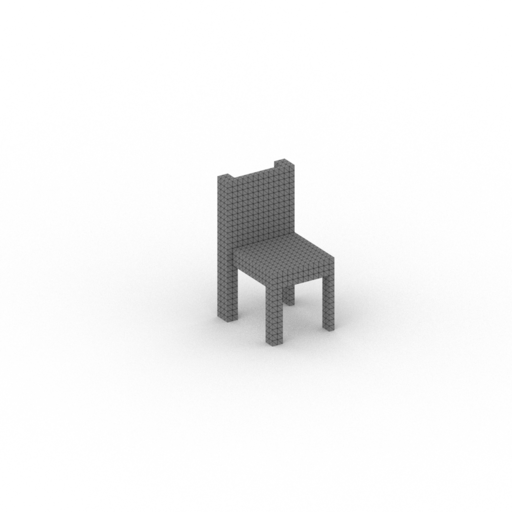} &
        \includegraphics[trim={3.5cm 4.5cm 3.5cm, 4.5cm}, clip,{width=.115\linewidth}]{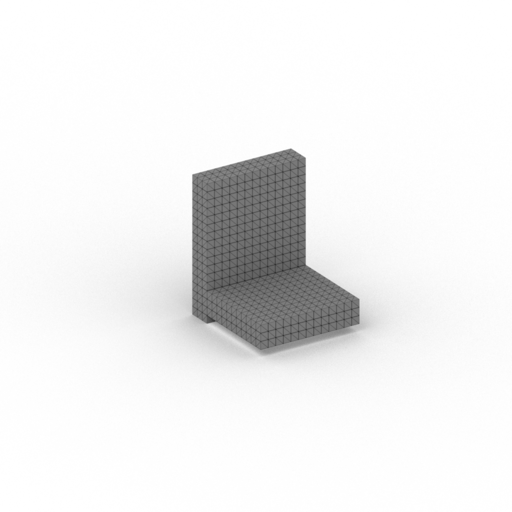} &
        \includegraphics[trim={3.5cm 4.5cm 3.5cm, 4.5cm}, clip,{width=.115\linewidth}]{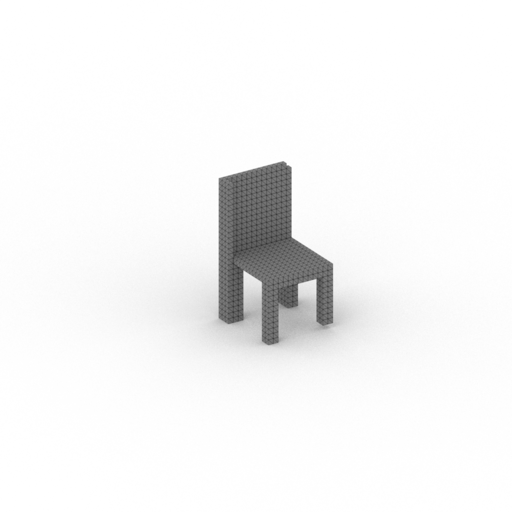} &
        \includegraphics[trim={3.5cm 4.5cm 3.5cm, 4.5cm}, clip,{width=.115\linewidth}]{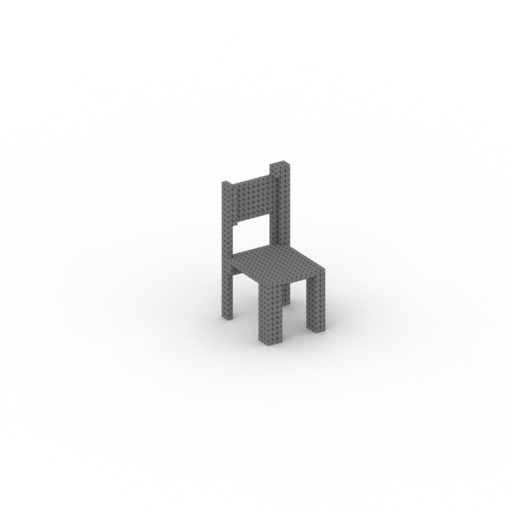} &
        \includegraphics[trim={3.5cm 4.5cm 3.5cm, 4.5cm}, clip,{width=.115\linewidth}]{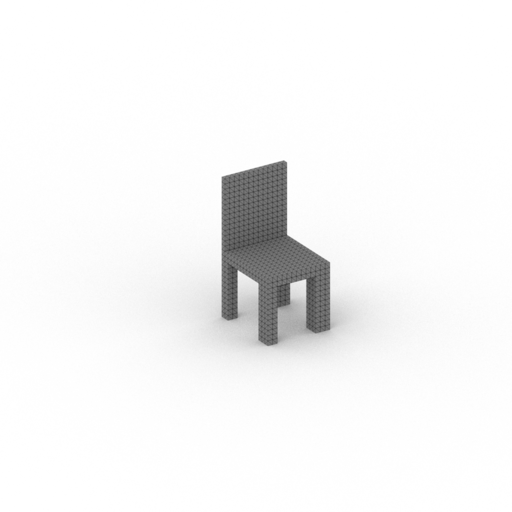} &
        \includegraphics[trim={3.5cm 4.5cm 3.5cm, 4.5cm}, clip,{width=.115\linewidth}]{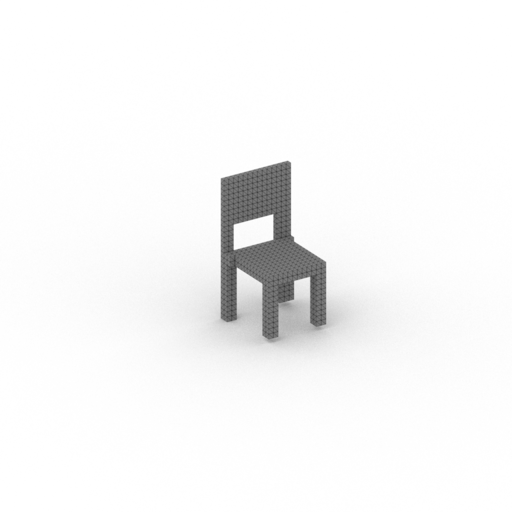} &
        \includegraphics[trim={3.5cm 4.5cm 3.5cm, 4.5cm}, clip,{width=.115\linewidth}]{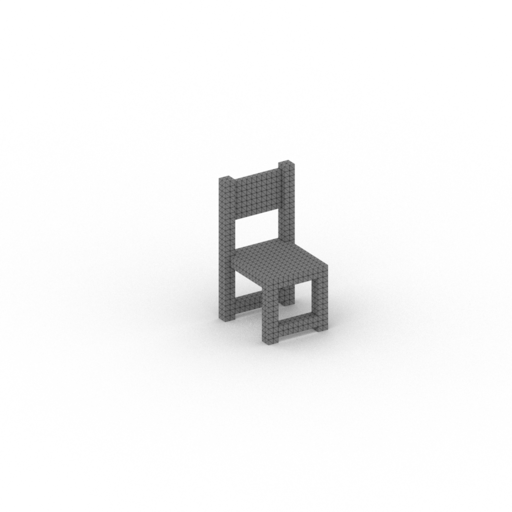} \\
        
        \includegraphics[trim={3.5cm 4.5cm 3.5cm, 4.5cm}, clip,{width=.115\linewidth}]{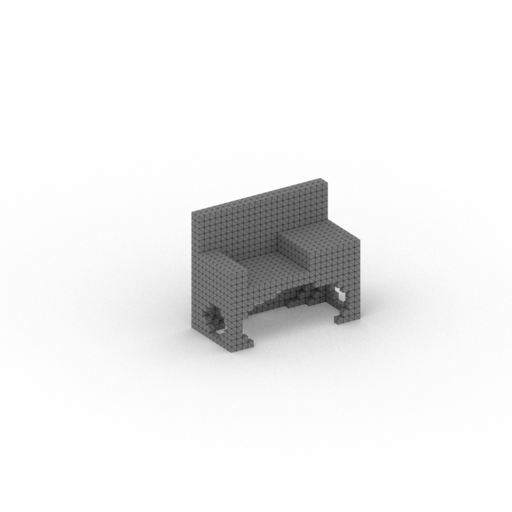} &
        \includegraphics[trim={3.5cm 4.5cm 3.5cm, 4.5cm}, clip,{width=.115\linewidth}]{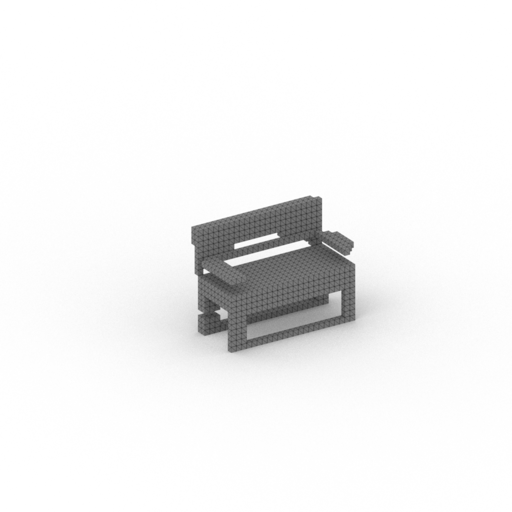} &
        \includegraphics[trim={3.5cm 4.5cm 3.5cm, 4.5cm}, clip,{width=.115\linewidth}]{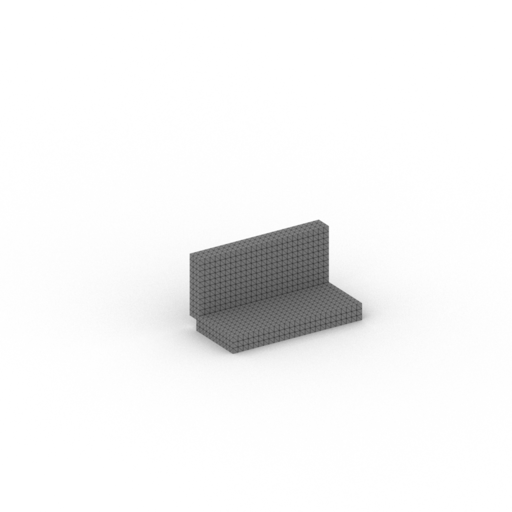} &
        \includegraphics[trim={3.5cm 4.5cm 3.5cm, 4.5cm}, clip,{width=.115\linewidth}]{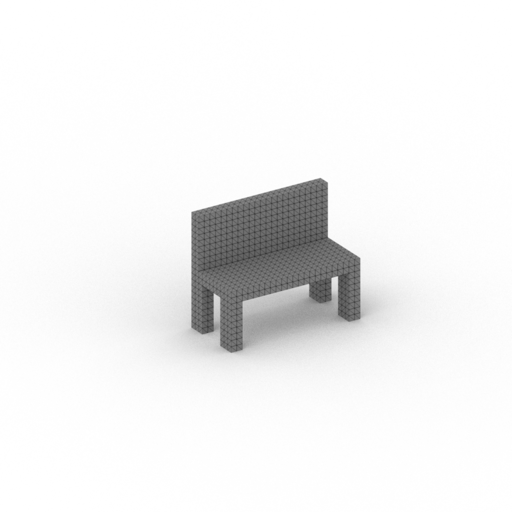} &
        \includegraphics[trim={3.5cm 4.5cm 3.5cm, 4.5cm}, clip,{width=.115\linewidth}]{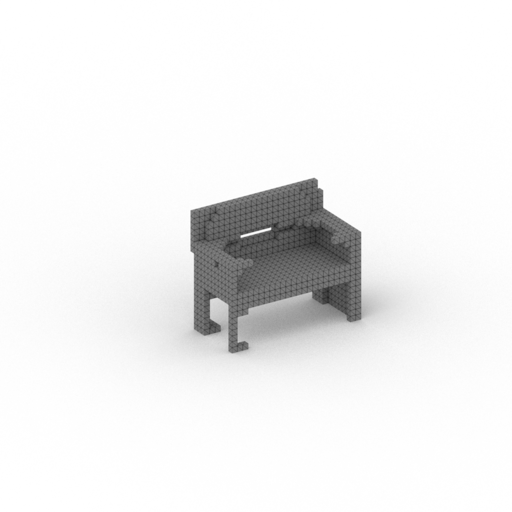} &
        \includegraphics[trim={3.5cm 4.5cm 3.5cm, 4.5cm}, clip,{width=.115\linewidth}]{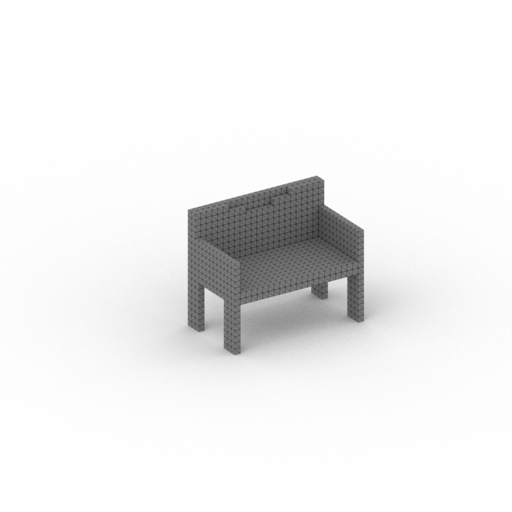} &
        \includegraphics[trim={3.5cm 4.5cm 3.5cm, 4.5cm}, clip,{width=.115\linewidth}]{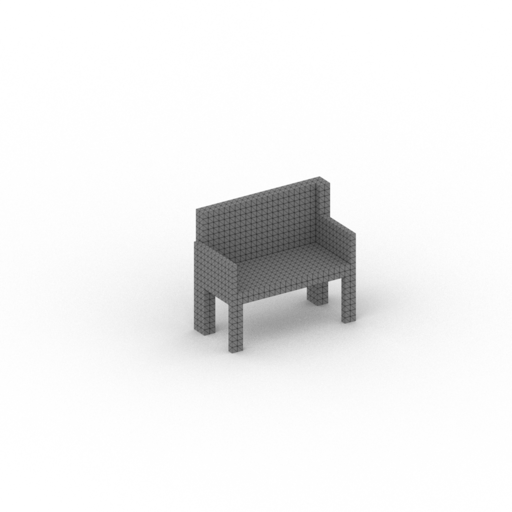} &
        \includegraphics[trim={3.5cm 4.5cm 3.5cm, 4.5cm}, clip,{width=.115\linewidth}]{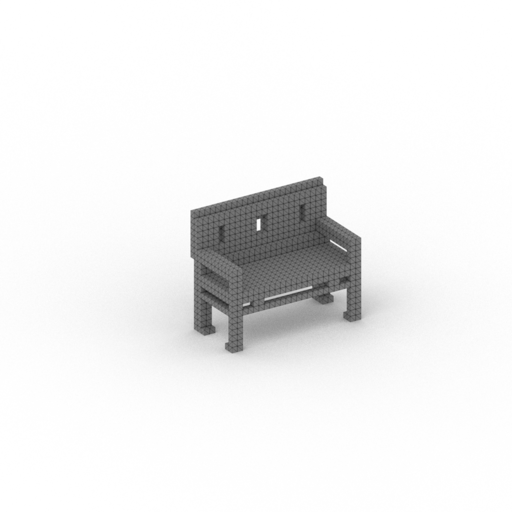} \\
        
    \end{tabular}
    \caption{3DCSG qualitative examples.} 
    \label{fig:qual_3d}
\end{figure*}

\begin{figure*}[t!]
    \centering
    \footnotesize
    \setlength{\tabcolsep}{1pt}
    \begin{tabular}{cccccccc}
        \textbf{SP} &  \textbf{WS} &  \textbf{RL} &  \textbf{ST} &  \textbf{LEST} & \textbf{LEST+ST} & \textbf{LEST+ST+WS} & \textbf{Target} \\
        
        \includegraphics[{width=.115\linewidth}]{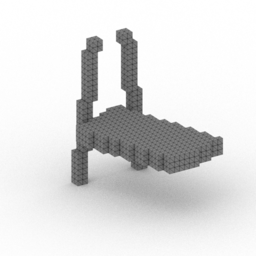} &
        \includegraphics[{width=.115\linewidth}]{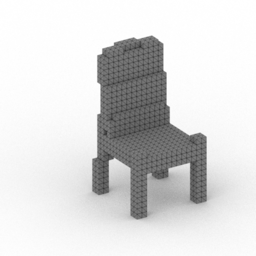} &
        \includegraphics[{width=.115\linewidth}]{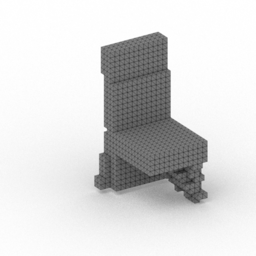} &
        \includegraphics[{width=.115\linewidth}]{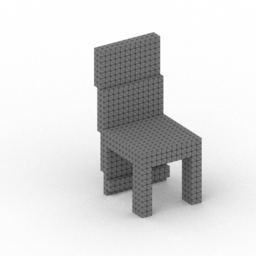} &
        \includegraphics[{width=.115\linewidth}]{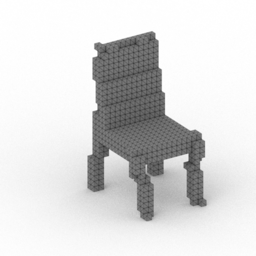} &
        \includegraphics[{width=.115\linewidth}]{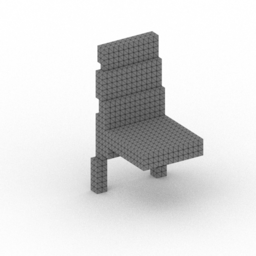} &
        \includegraphics[{width=.115\linewidth}]{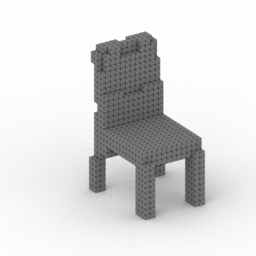} &
        \includegraphics[{width=.115\linewidth}]{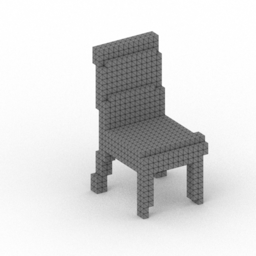} \\
        
        \includegraphics[{width=.115\linewidth}]{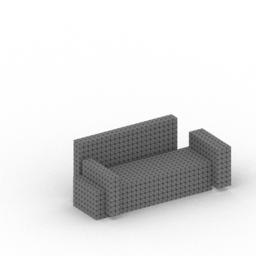} &
        \includegraphics[{width=.115\linewidth}]{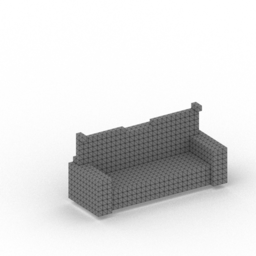} &
        \includegraphics[{width=.115\linewidth}]{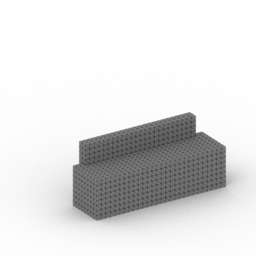} &
        \includegraphics[{width=.115\linewidth}]{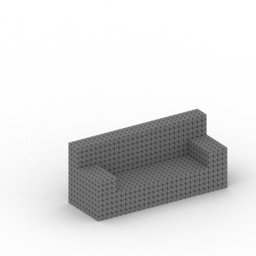} &
        \includegraphics[{width=.115\linewidth}]{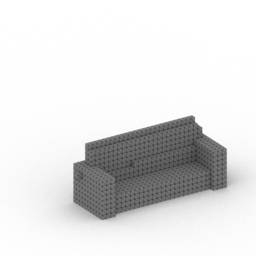} &
        \includegraphics[{width=.115\linewidth}]{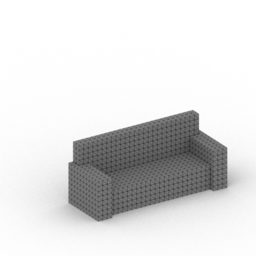} &
        \includegraphics[{width=.115\linewidth}]{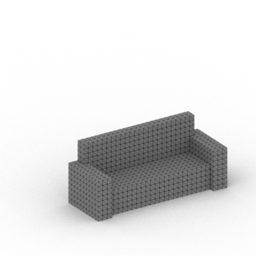} &
        \includegraphics[{width=.115\linewidth}]{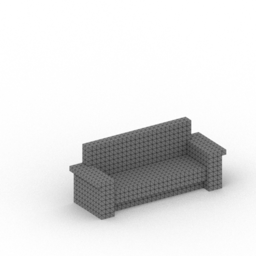} \\
        
        \includegraphics[{width=.115\linewidth}]{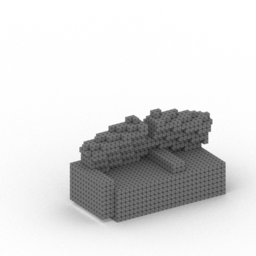} &
        \includegraphics[{width=.115\linewidth}]{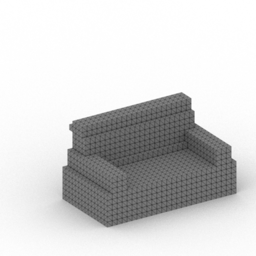} &
        \includegraphics[{width=.115\linewidth}]{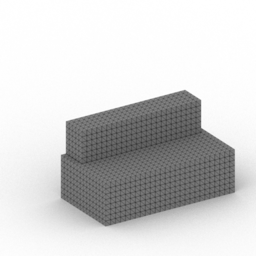} &
        \includegraphics[{width=.115\linewidth}]{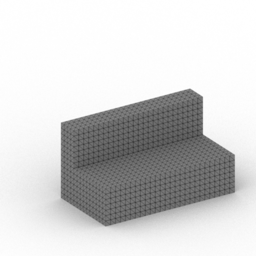} &
        \includegraphics[{width=.115\linewidth}]{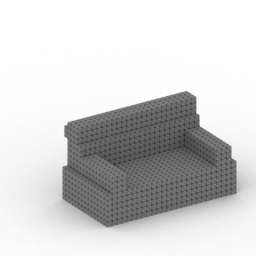} &
        \includegraphics[{width=.115\linewidth}]{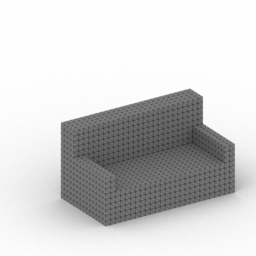} &
        \includegraphics[{width=.115\linewidth}]{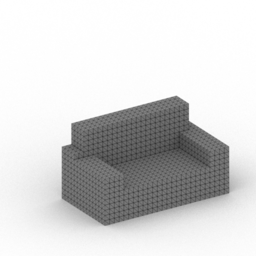} &
        \includegraphics[{width=.115\linewidth}]{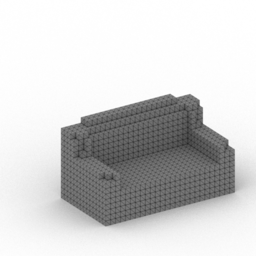} \\
        
        \includegraphics[{width=.115\linewidth}]{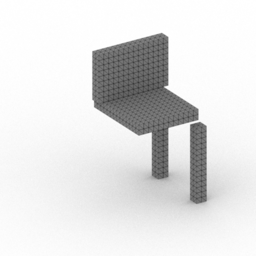} &
        \includegraphics[{width=.115\linewidth}]{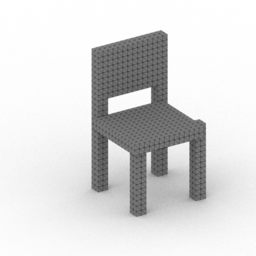} &
        \includegraphics[{width=.115\linewidth}]{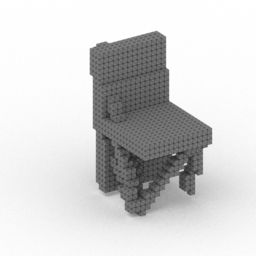} &
        \includegraphics[{width=.115\linewidth}]{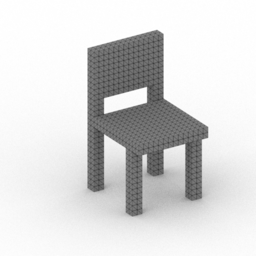} &
        \includegraphics[{width=.115\linewidth}]{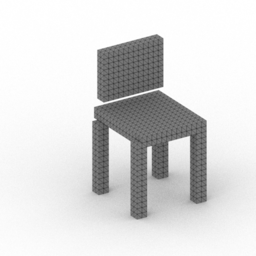} &
        \includegraphics[{width=.115\linewidth}]{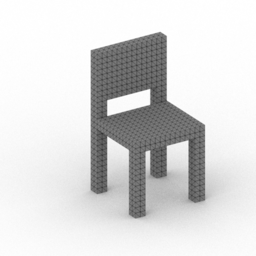} &
        \includegraphics[{width=.115\linewidth}]{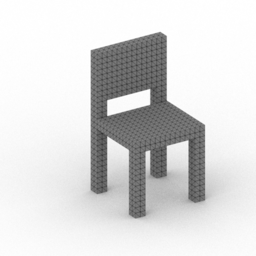} &
        \includegraphics[{width=.115\linewidth}]{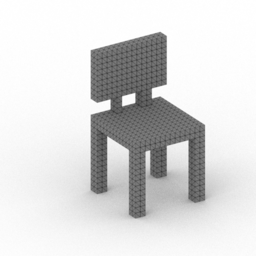} \\
        
        \includegraphics[{width=.115\linewidth}]{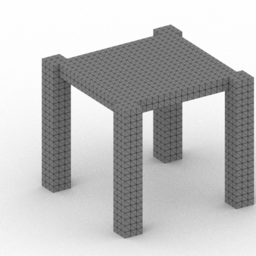} &
        \includegraphics[{width=.115\linewidth}]{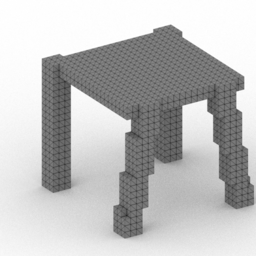} &
        \includegraphics[{width=.115\linewidth}]{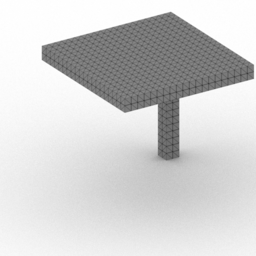} &
        \includegraphics[{width=.115\linewidth}]{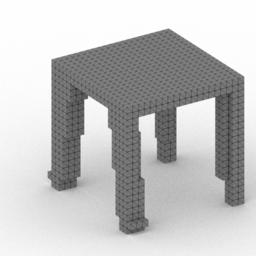} &
        \includegraphics[{width=.115\linewidth}]{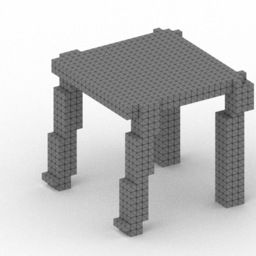} &
        \includegraphics[{width=.115\linewidth}]{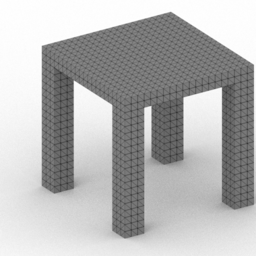} &
        \includegraphics[{width=.115\linewidth}]{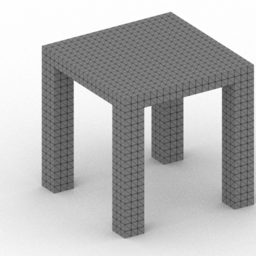} &
        \includegraphics[{width=.115\linewidth}]{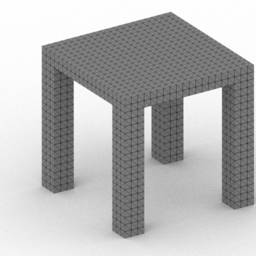} \\
        
        \includegraphics[{width=.115\linewidth}]{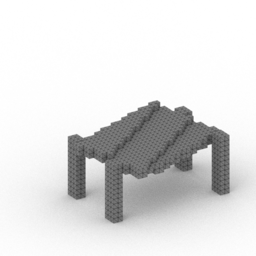} &
        \includegraphics[{width=.115\linewidth}]{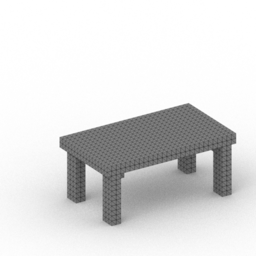} &
        \includegraphics[{width=.115\linewidth}]{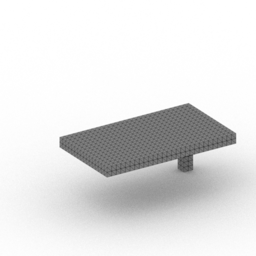} &
        \includegraphics[{width=.115\linewidth}]{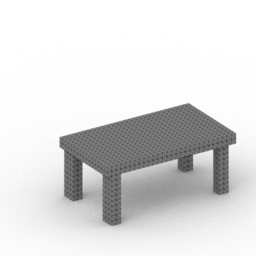} &
        \includegraphics[{width=.115\linewidth}]{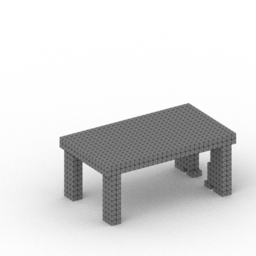} &
        \includegraphics[{width=.115\linewidth}]{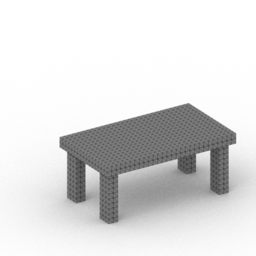} &
        \includegraphics[{width=.115\linewidth}]{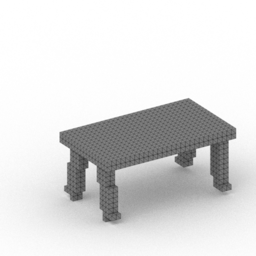} &
        \includegraphics[{width=.115\linewidth}]{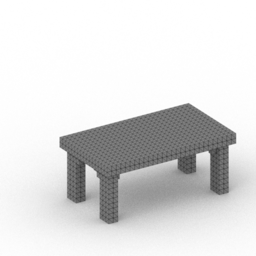} \\
        
        \includegraphics[{width=.115\linewidth}]{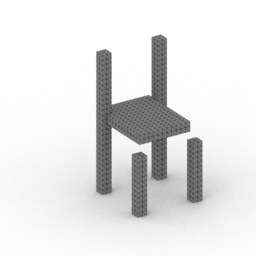} &
        \includegraphics[{width=.115\linewidth}]{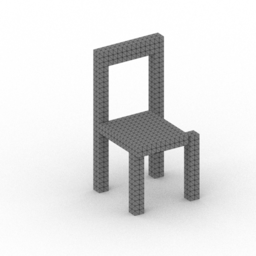} &
        \includegraphics[{width=.115\linewidth}]{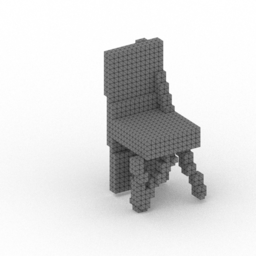} &
        \includegraphics[{width=.115\linewidth}]{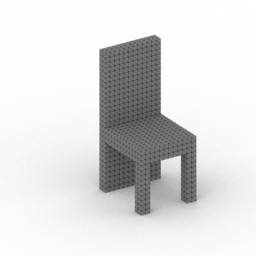} &
        \includegraphics[{width=.115\linewidth}]{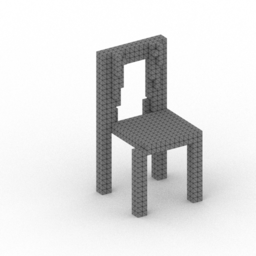} &
        \includegraphics[{width=.115\linewidth}]{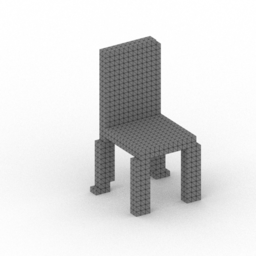} &
        \includegraphics[{width=.115\linewidth}]{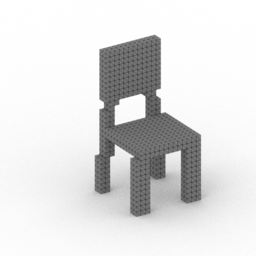} &
        \includegraphics[{width=.115\linewidth}]{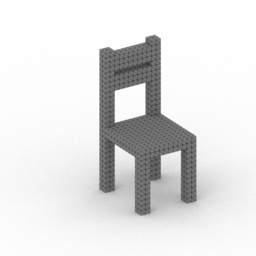} \\
        
        \includegraphics[{width=.115\linewidth}]{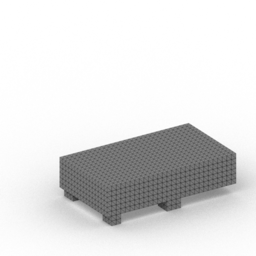} &
        \includegraphics[{width=.115\linewidth}]{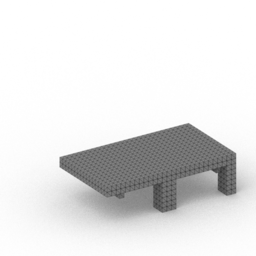} &
        \includegraphics[{width=.115\linewidth}]{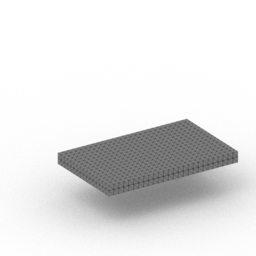} &
        \includegraphics[{width=.115\linewidth}]{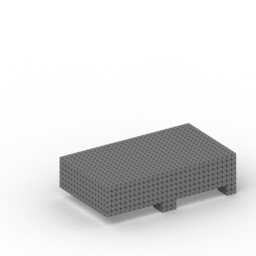} &
        \includegraphics[{width=.115\linewidth}]{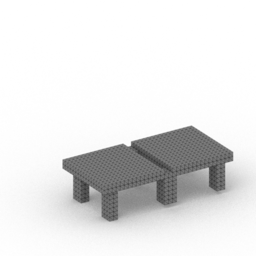} &
        \includegraphics[{width=.115\linewidth}]{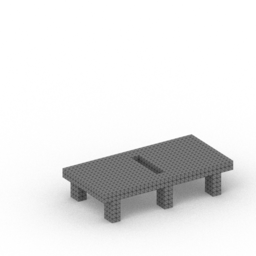} &
        \includegraphics[{width=.115\linewidth}]{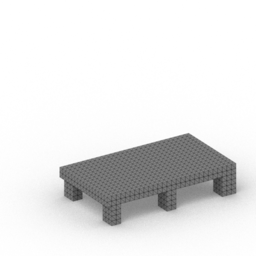} &
        \includegraphics[{width=.115\linewidth}]{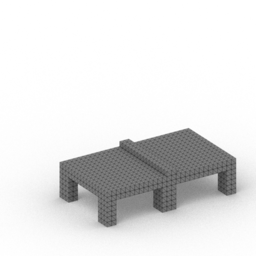} \\
        
        \includegraphics[{width=.115\linewidth}]{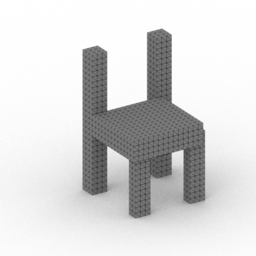} &
        \includegraphics[{width=.115\linewidth}]{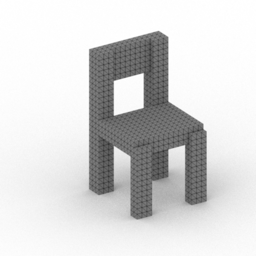} &
        \includegraphics[{width=.115\linewidth}]{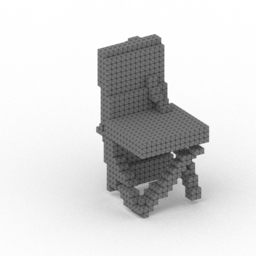} &
        \includegraphics[{width=.115\linewidth}]{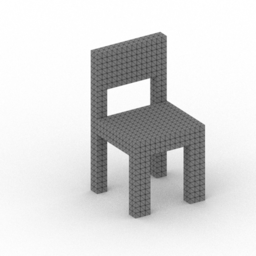} &
        \includegraphics[{width=.115\linewidth}]{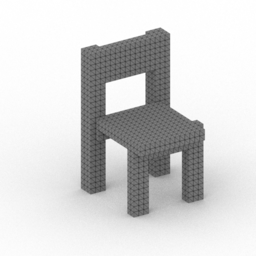} &
        \includegraphics[{width=.115\linewidth}]{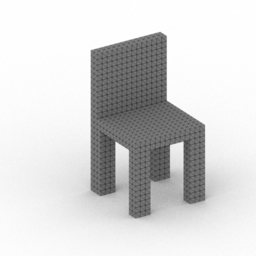} &
        \includegraphics[{width=.115\linewidth}]{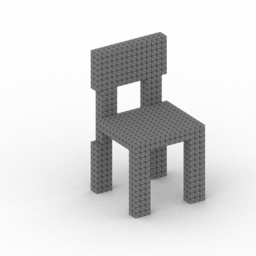} &
        \includegraphics[{width=.115\linewidth}]{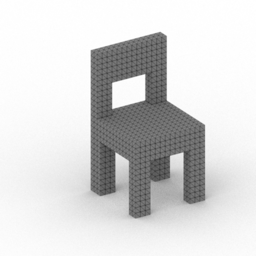} \\

    \end{tabular}
    \caption{ShapeAssembly qualitative examples.} 
    \label{fig:qual_sa}
\end{figure*}

We present additional qualitative results comparing various fine-tuning methods in Figure \ref{fig:qual_2d} (2D CSG), Figure \ref{fig:qual_3d} (3D CSG) and Figure \ref{fig:qual_sa} (ShapeAssembly).

